%% file: main.tex
\documentclass[12pt]{ociamthesis}  
\usepackage{amssymb}
\usepackage{graphicx}
\usepackage{amsmath}
\usepackage{amssymb}
\usepackage{booktabs}
\usepackage[hidelinks]{hyperref}
 
\usepackage{amsfonts}       
\usepackage{nicefrac}       
\usepackage{microtype}      
\usepackage{xcolor}         
\usepackage{microtype}      
\usepackage{graphicx}
\usepackage{amsmath,amssymb}
\usepackage{mathrsfs}
\usepackage{multirow}
\usepackage{float}
\usepackage{setspace}
\usepackage{mathrsfs} 
\usepackage{enumitem}

\title{\LARGE Exploring Probabilistic Models for Semi-supervised Learning }   

\author{Jianfeng Wang}             
\college{Linacre College}  

\degree{Doctor of Philosophy}     
\degreedate{Trinity 2023}         

\begin{document}

\baselineskip=18pt plus1pt

\setcounter{secnumdepth}{3}
\setcounter{tocdepth}{3}

\maketitle                  
\include{declarations}        
\include{acknowledgements}   
\include{abstract}          
\include{publications}          

\begin{romanpages}  
\tableofcontents 
\listoffigures              
\listoftables  
\end{romanpages}            


\chapter{Introduction} 
\section{Motivation}

Deep learning, an area of fervent research, has found numerous applications within the realm of vision processing \cite{krizhevsky2012imagenet, he2016deep, he2016identity, simonyan2014very, szegedy2015going}. Its triumphs in supervised learning scenarios, are primarily due to the employment of vast quantities of high-quality, labeled data. However, obtaining such labeled samples is often an arduous, costly, and time-intensive process. This process, typically demanding the expertise of professionals, stands as a significant hurdle in training an exceptional, fully-supervised deep neural network. For instance, in medical scenarios, measurements are taken using high-cost equipment, and the labels are derived from an exhaustive analysis by multiple expert humans. When only a handful of labeled samples are available, constructing a successful learning system becomes a formidable challenge. In contrast, unlabeled data is typically abundant and easily procured at a low cost. As such, the idea of harnessing a significant amount of unlabeled data to train deep models, given a small number of labeled samples, becomes enticing.

\begin{figure*}[t]
\centering
\includegraphics[width=0.8\linewidth]{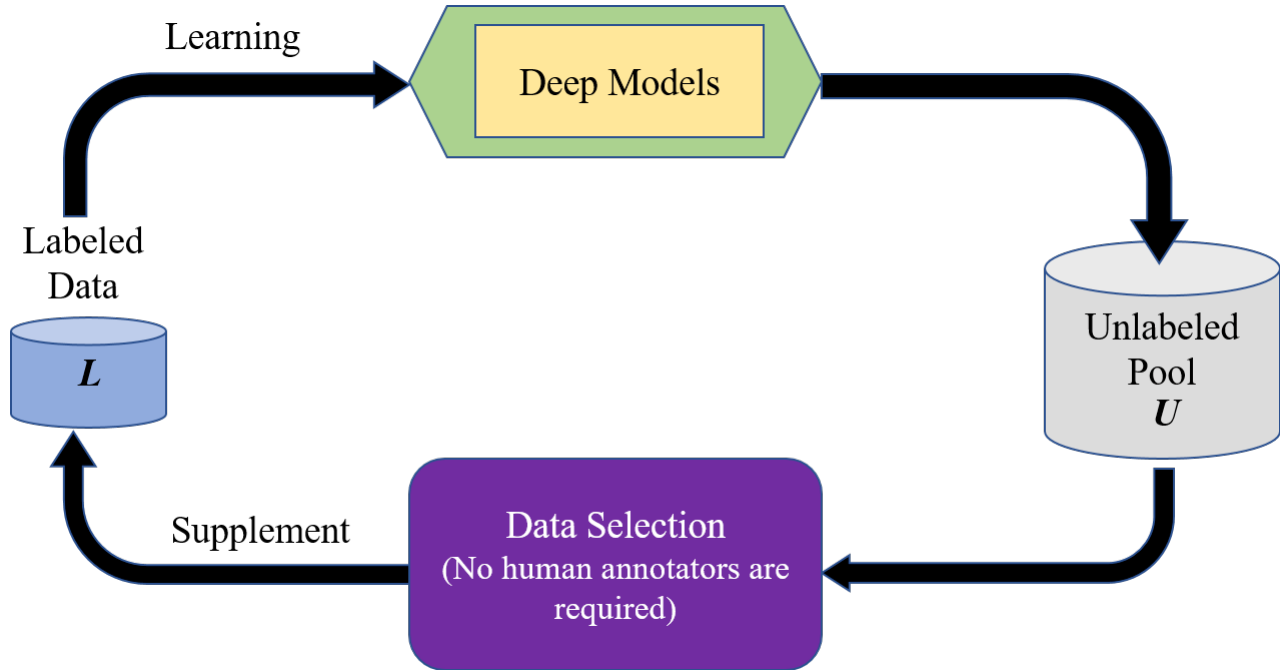}
\caption{
Overview of the SSL pipeline. Unlike active learning, SSL does not require user involvement in choosing data. The selected data are used to enlarge the original labeled dataset, and whether pseudo-labels are used depends on the training strategy employed (e.g., pseudo-labeling strategy or consistency strategy).}
\label{fig:overall_ssl} 
\end{figure*}

In recent years, two types of learning paradigms have drawn significant attention from researchers: active learning (AL) \cite{cacciarelli2023active} and semi-supervised learning (SSL) \cite{van2020survey}. Both paradigms harness unlabeled data to enhance the capabilities of deep neural networks. AL is a learning algorithm that interactively queries the user to select new and challenging samples and annotate their labels. Although it can be more efficient than manually labeling all data without the assistance of deep neural networks, it still requires continuous involvement from a human or another system to provide labels for queried data points, which can be time-consuming or impractical in some scenarios. By contrast, SSL is a learning paradigm that can more automatically utilize both labeled and unlabeled data to construct predictive models. 
A general learning pipeline of SSL, as shown in Figure~\ref{fig:overall_ssl}, first involves training deep models with labeled data. The labels are then applied to unlabeled data to assist in data selection. The selected unlabeled data are used to augment the original labeled dataset, which in turn is used to train deep models. 
In this approach, the involvement of users or human annotators is not required, significantly reducing the burden of manually labeling large volumes of data. Therefore, this thesis focuses on the learning paradigm, namely SSL, which can reduce reliance on human annotators to the greatest extent.

Most SSL methods have primarily employed two strategies: pseudo-labeling and consistency regularization. The term 'pseudo-labeling' refers to the process of assigning inferred labels to unlabeled samples by models. These labels are not the actual ground truth and could be incorrect, as they are generated entirely by deep learning models. 
``Consistency'' involves maintaining the same semantic meaning across different perturbations of an image. 
In particular, pseudo-labeling approaches \cite{lee2013pseudo, zhai2019s4l, wang2020enaet, pham2021meta} hinge on the high confidence of pseudo-labels, which can be integrated into the training dataset as labeled data. These methods typically train models on a minimal amount of labeled data initially, and then utilize the models' predictions on the unlabeled data as pseudo-labels. On the other hand, consistency regularization-based methods \cite{bachman2014learning, sajjadi2016regularization, laine2016temporal, berthelot2019mixmatch, xie2019unsupervised} operate by applying varying transformations (e.g., rotation or adding noise) to an input unlabeled image and incorporating a regularization term to ensure the consistency of their predictions. 
Very recent state-of-the-art methods \cite{sohn2020fixmatch, li2021comatch, rizve2021defense, zhang2021flexmatch} often combine the strengths of both these strategies. Given an unlabeled image, weak and strong data augmentations are performed, resulting in two versions of the image. Subsequently, a pseudo-label is generated based on its weakly-augmented version, which serves as the true label for its strongly augmented counterpart, thereby training the entire framework. Since SSL can reduce the labor involved in annotation and is easy to deploy in practice, it has become a prevalent technique in building AI systems. This is especially true for small companies or universities with limited computational resources and budgets.

Generally, it's a known issue that, due to the unknown distribution of unlabeled data, some of them presents challenging samples for a classifier trained on labeled data. Inevitably, this can lead to the generation of incorrect pseudo-labels during the training process, negatively affecting optimization and resulting in deep models that underperform. This raises a crucial question: Can decision-makers truly trust the predictions of these deep models, particularly in critical fields like medical image analysis and autonomous driving ? For a secure functioning AI system, every decision should be grounded in the model's certainty of its predictions, measurable through uncertainty estimates.

Recent methods in SSL can be largely classified as deterministic, as they aim to produce a point estimate that maximizes the likelihood. The investigation of using deterministic models to quantify uncertainty is still in its early stages. In contrast, the probabilistic approach has been heavily studied in recent years, and it strives to construct predictive distributions, which can naturally quantify uncertainty.
However, there has been a dearth of studies on probabilistic models in the realm of SSL. For example, Monte Carlo (MC) dropout \cite{gal2016dropout} has been the predominant option for implementing such models, which is a technique for approximating Bayesian inference in neural networks, allowing for uncertainty estimation in the model's predictions. 
Given that SSL is susceptible to the influence of incorrect pseudo-labels, potentially leading to suboptimal AI model performance, and acknowledging that uncertainty can enhance the safety of these models, it is essential to investigate and develop diverse and advanced probabilistic algorithms for SSL.


This dissertation not only explores the most popular tool for implementing Bayesian Neural Networks in the medical domain \cite{wang2022rethinking}, namely MC dropout, but also delves into novel probabilistic methodologies \cite{wang2022np, wang2023np, wang2023np1}. These new approaches are designed to advance the related research in SSL and also to address the limitations of the current primary tool for constructing probabilistic models in SSL.

\section{Contributions}

The work in this thesis can be grouped into the following parts.

\begin{itemize}[leftmargin=*, itemsep=0.5pt]
\item Building upon the widespread use of MC dropout as a means to approximate Bayesian Neural Networks (BNNs), we delve into its application in semi-supervised medical image segmentation. 
Previous works exhibit two notable limitations. First, these studies predominantly construct their models with only labeled data in the early training phase, which can lead to overfitting due to the limited volume of such data. This approach may result in the production of inaccurate pseudo labels, thereby negatively affecting the subsequent training process. Second, these models are merely partially grounded in Bayesian deep learning and lack comprehensive design within the Bayesian framework. This leads to a deficiency in providing a Bayesian formulation for their models and results in a lack of a sound theoretical basis. Consequently, some model modules are designed empirically, leaving the functionality of these components unclear. 
In response to these issues, we propose an innovative generative model, named the Generative Bayesian Deep Learning (GBDL) architecture \cite{wang2022rethinking}, along with its corresponding full Bayesian formulation. The GBDL does not solely depend on the minimal amount of labeled data in the initial training stage, but rather on both labeled and unlabeled data. This approach could alleviate the overfitting issue by considering the data distribution across the entire dataset. Additionally, the GBDL is fully constructed within the Bayesian framework. Its associated full Bayesian formulation lays a solid theoretical foundation, ensuring each part of GBDL has a corresponding probabilistic formulation and interpretation. 

\item 
Research in probabilistic models for SSL remains under-explored, and the conventional MC dropout method presents challenges for uncertainty quantification, such as low efficiency and subpar performance, restricting its real-world applicability. 
Motivated by these issues, I sought an alternative to MC dropout for SSL, leading to the adaptation of a renowned probabilistic approach, Neural Processes (NPs) \cite{garnelo2018neural}, for this purpose. 
NPs is a probabilistic model that establishes distributions over functions and is adept at quickly adapting to new observations while accurately estimating the uncertainty of each one, with relatively low time consumption. Due to these merits, I modified NPs, resulting in a new framework named NP-Match \cite{wang2022np}, designed specifically for large-scale semi-supervised image classification. 
Furthermore, to improve the robustness of NP-Match against the possible incorrect pseudo-labels, I replaced the Kullback-Leibler (KL) divergence in the evidence lower bound (ELBO) of the original NPs with a novel uncertainty-guided skew-geometric Jensen-Shannon (JS) divergence, which further enhances the performance of NP-Match. To verify the effectiveness of our method, we conducted experiments on several benchmarks with different settings.
As a result, NP-Match achieves competitive performance compared to the state-of-the-art results. Moreover, it estimates uncertainty more quickly than MC dropout-based probabilistic models, bolstering both training and testing efficiency.


\item In a bid to further investigate the application of NPs in SSL, I have devised a novel probabilistic model named NP-SemiSeg \cite{wang2023np} for semi-supervised semantic segmentation. The model has been  designed to address the inefficiencies prevalent in  MC-dropout-based segmentation models. When compared to NP-Match, I primarily made two modifications when designing NP-SemiSeg. Firstly, I re-engineered how the global latent variables are utilized based on the characteristic of the segmentation task. Secondly, I introduced attention mechanisms to enhance the model's capability to accurately fit the training data. Thorough experimental results demonstrate that NP-SemiSeg surpasses MC dropout not just in terms of accuracy but also in speed concerning uncertainty estimation. This indicates its potential to be an effective alternative as the probabilistic model for semi-supervised semantic segmentation.

\end{itemize}

\section{Thesis Statement}


This thesis studies advanced probabilistic models, including both their theoretical foundations and practical applications, for different SSL tasks. Our proposed probabilistic methods are able to improve the safety of AI systems by providing reliable uncertainty estimates, and at the same time, when compared to the performance of deterministic counterparts, achieve competitive performance. The promising results shown in this thesis indicate that our methods have the potential to revolutionize the way of estimating uncertainty for SSL, paving the way for the future discovery of highly effective and efficient probabilistic approaches.

\section{Outline}
 
The structure of this thesis is as follows:

{\bf Chapter 2} provides a comprehensive background that contains topics including Semi-supervised Learning, Monte Carlo Dropout for Bayesian Approximation, Gaussian Processes, and Neural Processes.

{\bf Chapter 3} delves into the Generative Bayesian Deep Learning (GBDL) architecture \cite{wang2022rethinking} that has been developed for semi-supervised medical image segmentation. This chapter provides a detailed overview of the framework design, its complete Bayesian formulation, as well as quantitative analyses and visualization results.

{\bf Chapter 4} details the modifications made to Neural Processes (NPs) for large-scale semi-supervised image classification \cite{wang2022np}. This includes a thorough explanation of the NP-Match construction, the use of the evidence lower bound (ELBO) for optimization, and our innovative proposal of the uncertainty-guided skew-geometric Jensen-Shannon (JS) divergence. In addition, this chapter presents extensive experimental outcomes and provides an analysis of the results on several benchmarks under different settings.


{\bf Chapter 5}  focuses on our approach to adapting Neural Processes (NPs) for semi-supervised semantic segmentation \cite{wang2023np}. It provides a meticulous description of the NP-SemiSeg design,  the evidence lower bound (ELBO) for optimization, and the attention mechanism. Comprehensive experimental results are also presented in this chapter to demonstrate the effectiveness of NP-SemiSeg. 

{\bf Chapter 6} concludes this thesis by summarizing its contributions and findings. It also outlines potential areas for future exploration and research.

\chapter{Background} 

\section{Semi-supervised Learning}

Semi-supervised learning (SSL) refers to the learning paradigm that trains deep models with limited labeled data and a large amount of unlabeled data. Its goal is to minimize the supervised loss on labeled data, and to explore the relationship between labeled and unlabeled data in the input or label space. SSL has been widely explored in different areas, such as computer vision \cite{krizhevsky2012imagenet, he2016deep}, data mining \cite{zhang2018survey}, and natural language processing \cite{devlin2018bert}. This thesis mainly studies SSL for computer vision tasks. To guarantee that unlabeled data can improve the performance of deep models, some basic assumptions related to the data distribution should be upheld, which can be summarized as follows:

{\bf  Smoothness assumption.}  
The smoothness assumption posits that if two input points, $x_1$ and $x_2$, within the input space are in close proximity, their corresponding labels, $y_1$ and $y_2$, should be identical. This principle, while frequently employed in supervised learning, provides additional benefits in a semi-supervised context by extending transitivity to unlabeled data. To illustrate, suppose we have a labeled data point, $x_1$, and two unlabeled data points, $x_2$ and $x_3$. If $x_1$ is close to $x_2$, and $x_2$ is close to $x_3$, but $x_1$ is not close to $x_3$, the smoothness assumption allows us to predict that $x_3$ will share the same label as $x_1$. This prediction operates on the basis that label-related proximity is transitively relayed through $x_2$.

{\bf Low-density assumption.}  The low-density assumption supposes that a classifier's decision boundary should ideally pass through low-density regions within the input space, thereby avoiding high-density areas. This assumption is predicated on $p(x)$, the true distribution of the input data. Given a finite sample from this distribution, the assumption suggests that the decision boundary should reside where data points are sparse. In this context, the low-density assumption closely parallels the smoothness assumption, and can even be viewed as its analogue in relation to the underlying data distribution. In particular, if the smoothness assumption is met, then any pair of closely located data points will share the same label. As a result, in any high-density area of the input space, we expect all data points to carry the same label. This allows us to construct a decision boundary that passes only through low-density areas in the input space, thus fulfilling the low-density assumption.

{\bf  Manifold assumption.} 
The manifold assumption is pivotal for SSL, especially for high-dimensional data like images or text. This assumption suggests that such data, while existing in a high-dimensional space, actually lie on a lower-dimensional manifold. The low-dimensional manifold refers to a space where local properties resemble those of a simpler Euclidean space. This implies that high-dimensional data points follow an underlying, simpler structure, which can be revealed in fewer dimensions. In the context of SSL, the manifold assumption presents that: (a) the input space consists of multiple lower-dimensional manifolds, all of which host the given data points, and (b) data points residing on the same manifold share the same label. As a result, if both the geometric structure of the data points and the manifolds they reside on are well identified during training, a model with good generalization and superior performance can be obtained.

Drawing on these foundational assumptions, modern semi-supervised learning algorithms designed for computer vision tasks employ three strategies. These strategies include consistency regularization,  pseudo-labeling, and an hybrid approach that combines both methodologies.  For the consistency regularization strategy, a consistency regularization term is applied to the loss function to express the constraints assumed by prior knowledge. Relying on either the manifold assumption or the smoothness assumption, consistency regularization encompasses a variety of methods. These methods maintain that realistic perturbations to data points should not change the model's output \cite{oliver2018realistic}, and such belief is usually implemented via designing loss functions over the feature representations or the predicted distributions of those data points. 
The pseudo-labeling strategy relies on the smoothness assumption, and it assigns pseudo-labels to unlabeled data based on the confidence of predictions, which can then be incorporated into the training dataset as labeled data.  
This method generally follows two predominant patterns. The first aims to enhance the overall framework's performance by leveraging unlabeled samples on which multiple networks trained on labeled data have divergent predictions, as these hard samples can provide useful information to improve performance \cite{qiao2018deep, dong2018tri}. 
For instance, tri-net \cite{dong2018tri} is composed of a shared encoder module and three decoder modules, each of which is trained by its corresponding training set. Then, given an unlabeled sample, the three modules predict its label. If their predictions are not consistent and, for example, the first two of them have the same prediction, such prediction will be used as its pseudo-label, and the unlabeled sample along with its pseudo-label will be added to the training set of the third module. 
The second pattern, self-training, aims at training a model on a small amount of labeled data, and using its predictions on the unlabeled data as pseudo-labels \cite{pham2021meta, wang2020enaet, zhai2019s4l}. Then, the unlabeled data and their pseudo-labels are used to expand the labeled dataset. 
The hybrid methods, such as FixMatch \cite{sohn2020fixmatch}, UPS \cite{rizve2021defense}, FlexMatch \cite{zhang2021flexmatch}, have achieved state-of-the-art results in recent years. They combine ideas from the above-mentioned methods, i.e., consistency regularization and pseudo-labeling, for performance improvement.

These assumptions can also serve as the foundation for exploring probabilistic models and uncertainty quantification under SSL. For instance, the smoothness assumption aligns with some classical probabilistic methods such as Gaussian Processes (GPs) or Neural Processes (NPs). The kernels of GPs (or implicit kernels in NPs) can compare data points when making predictions, in which case target points are more likely to be given labels similar to those of their close contextual data points. The low-density assumption can provide insights into the relationship between uncertainty  and out-of-distribution samples. In particular, if an unlabeled data point is out-of-distribution, it falls into a low-density region. When in this region, deep models tend to make predictions with high uncertainty, due to the high entropy value. The manifold assumption could also be useful when performing uncertainty quantification. When labeled data points and an unlabeled data point reside on the same manifold, the unlabeled data point can be assigned the same label as those labeled data points with low uncertainty.

\section{Monte Carlo Dropout for Bayesian Approximation}
\label{sec:2.2}

Dropout is a fundamental regularization technique used in neural network training to prevent overfitting \cite{srivastava2014dropout}. It works by randomly omitting a subset of nodes during each training epoch, which disrupts the co-dependence of neurons, compelling each to independently produce valuable features, and promoting the learning of more robust and generalizable features. This technique can also be seen as training a pseudo-ensemble of networks within a single model architecture. For the purpose of improving a model's generalization capability, dropout is typically not applied during the testing or inference phase. At this stage, all neurons are activated, effectively imitating an ensemble of models.

Besides, it has been proven that a neural network with arbitrary depth and non-linearities, with dropout applied before every weight layer, is mathematically equivalent to an approximation to the probabilistic deep Gaussian process \cite{gal2016dropout}. Specifically, 
let $\hat{y}$ be the output of a neural network (NN) model with $L$ layers and a loss
function $E(\cdot,\cdot)$ such as the softmax loss or the Euclidean loss (square loss). 
For each layer $i=1,2,...,L$, the NN's weight matrices of dimensions $K_i\times K_{i-1}$ are denoted by $W_{i}$, and the bias vectors of dimensions $K_i$ are denoted by $b_i$. 
We denote by $y_n$ the observed output corresponding to input $x_n$
for $1\leq n \leq N$ data points, and the input and output sets
as $X$, $Y$. 
During NN optimisation, a regularisation term is often added. We use $l_2$ regularisation is usually used weighted by some weight decay $\lambda$, resulting in a minimisation objective:
\begin{equation} 
\label{eq:2_2_1}
\mathcal{L}_{dropout}: = \frac{1}{N} \sum^{N}_{n=1} E(y_n, \hat{y_n}) + \lambda \sum^{L}_{i=1} (||W_i||_2^2 + ||b_i||_2^2).
\end{equation} 
With dropout, binary variables are sampled for every input point and for every network unit in each layer. Each binary variable takes value 1 with probability $\rho_i$ for layer $i$. A unit is dropped (i.e., its value is set
to zero) for a given input if its corresponding binary variable takes value 0. 

From the Bayesian perspective, the weights are sampled from distributions. In particular,  we write $\omega= \{W_i\}_{i=1}^L$ and let each row of $W_i$ distribute according to a prior distribution $p(w)$.  The predictive probability of the deep
GP model given some precision parameter $\tau > 0$ can be parameterized as:
\begin{equation} 
\label{eq:2_2_2}
\begin{aligned}
&P(y|x, X, Y) = \int p(y|x, \omega) p(\omega| X, Y)d\omega, \\ 
&p(y|x, \omega) =  \mathcal{N}(y;\hat{y}(x, \omega), \tau^{-1}I_D), \\
&\hat{y}(x, \omega=\{W_1,...,W_L\})=\sqrt{\frac{1}{K_L}}W_L\sigma(...\sqrt{\frac{1}{K_1}}W_2\sigma(W_1x+b_1)...).
\end{aligned}
\end{equation} 
The posterior distribution $p(\omega| X, Y)$  in Eqn.~(\ref{eq:2_2_2}) is intractable. To solve this issue, a variational distribution $q(\omega)$ over matrices is used to approximate the intractable posterior, whose columns are randomly set to zero. We define $q(\omega)$ as follows:
\begin{equation} 
\label{eq:2_2_3}
\begin{aligned}
&W_i = M_i \, diag([z_{i,j}]_{j=1}^{K_{i-1}})
\\ 
& z_{i,j} \sim Bernoulli(\rho_i)  \quad \text{for}\  i = 1, ..., L,\  j = 1, ..., K_{i-1}.
\end{aligned}
\end{equation} 
Some probabilities $\rho_i$ and matrices $M_i$ are variational parameters. 
$diag(\cdot)$ maps a vector to a diagonal matrix whose diagonal
is the elements of the vector. 
The binary variable $z_{i,j} = 0$ corresponds 
to unit $j$ in layer $i-1$ being dropped out as an input to layer $i$. 
To learn the variational parameters, the learning objective can be written as follows:
\begin{equation} 
\label{eq:2_2_4}
\begin{aligned}
& - \int q(\omega)\;\text{log}\,p(Y| X, \omega)d\omega + \text{KL}(q(\omega)||p(\omega)).
\end{aligned}
\end{equation} 
With every single sample pair ($x_n$, $y_n$), the first term $-\int q(\omega)\;\text{log}\,p(y_n| x_n, \omega)d\omega$ can be approximated by Monte Carlo integration with a single sample $\hat{\omega}_{n}\sim q(\omega)$, i.e., $-\text{log}\,p(y_n| x_n, \hat{\omega}_{n})$. Then, it has also been proven \cite{gal2016dropout} that the second term can be approximated by $\sum^{L}_{i=1} (\frac{\rho_il^2}{2}||W_i||_2^2 + \frac{l^2}{2}||b_i||_2^2$) with a prior length scale $l$.  Given model precision $\tau$ and constant $\frac{1}{\tau N}$, the objective can be written as:
\begin{equation} 
\label{eq:2_2_5}
\begin{aligned}
&  \frac{1}{N}\sum_{n=1}^{N}\frac{-\text{log}\,p(y_n| x_n, \hat{\omega}_{n})}{\tau} + \sum^{L}_{i=1} (\frac{\rho_il^2}{2\tau N}||W_i||_2^2 + \frac{l^2}{2\tau N}||b_i||_2^2).
\end{aligned}
\end{equation} 

\noindent The derived Eqn.~(\ref{eq:2_2_5}) is indeed equivalent to the objective shown in Eqn.~(\ref{eq:2_2_1}). This suggests  that optimizing a neural network with dropout before every weight layer can be understood as an approximation of a variational distribution to a deep Gaussian process.

MC dropout has been a primary tool for quantifying uncertainty in the field of computer vision and medical vision. In general, there are mainly two types of uncertainty named  epistemic (model) uncertainty and aleatoric 
(data) uncertainty \cite{kendall2017uncertainties}. The epistemic uncertainty accounts for uncertainty in the model parameters, which comes from the lack of knowledge. 
This type of uncertainty is potentially reducible through further research, data collection, or improved understanding. A common approach to estimate it is to perform several feedforward passes for a given input, to get their corresponding outputs, each of which is a prediction from a sampled model. The epistemic uncertainty can be estimated by calculating entropy or variance with those outputs \cite{kendall2017uncertainties}. The aleatoric uncertainty refers to the inherent randomness or variability in a system or observation that cannot be reduced, regardless of how much information is collected. In practice, this uncertainty is usually modeled as a fixed noise term (homoscedastic) or a network is used to predict it (heteroscedastic) \cite{kendall2017uncertainties}. 
As MC dropout is flexible and easy to use, it has become a widely-used tool for approximating BNNs in medical image segmentation frameworks, a topic also explored in this thesis. 
But MC dropout has its limitations for practical applications. Firstly, it's time-consuming as it necessitates several feedforward passes to obtain uncertainty at both the training and testing stages, especially when larger models are employed. Secondly, architectural decisions like the placement of dropout layers and the setting of the dropout rate are often empirically determined, potentially leading to suboptimal performance.

There are several variants, such as DropConnect \cite{wan2013regularization} and MC Batch Normalization \cite{teye2018bayesian}, that share similar ideas with MC dropout. However, MC dropout has been widely used in SSL tasks \cite{rizve2021defense, yu2019uncertainty, wang2021tripled, shi2021inconsistency, wang2020double}, and therefore, we mainly focus on MC dropout and treat it as an important baseline in this dissertation.

\section{Gaussian Processes}

Gaussian Processes (GPs) are a powerful and flexible non-parametric probabilistic model used for both regression and classification tasks \cite{gpml}. Even though they are not explored in this thesis, they lay a solid foundation for the proposed algorithms, and understanding this can help readers better comprehend our methods. 
GPs provide a way of defining prior probability distributions over functions, which can be updated given data to form a posterior distribution. Instead of just producing a point estimate, a GP provides a full probabilistic distribution that captures the uncertainty of predictions.

Formally, a Gaussian process is a collection of random variables, any finite number of which have a joint Gaussian distribution. In a Gaussian process, every point in some continuous input space is associated with a normally distributed random variable. Moreover, every finite collection of these random variables has a multivariate Gaussian distribution. A Gaussian process is completely specified by its mean function and covariance function. We define mean function $m(x)$ and the covariance function $k(x, x')$ of a real process $f(x)$ as:
\begin{equation} 
\label{eq:2_3_1}
\begin{aligned}
&  m(x) = \mathbb{E}[f(x)] \\
&  k(x, x') = \mathbb{E}[(f(x)-m(x))(f(x')-m(x'))], \\
\end{aligned}
\end{equation} 
and the Gaussian process as: 
\begin{equation} 
\label{eq:2_3_2}
\begin{aligned}
&  f(x) \sim \mathcal{GP}(m(x), k(x, x')).
\end{aligned}
\end{equation}
For instance, concerning a noise-free regression task, the linear regression model can be defined as $f(x)=\phi(x)w$ with prior $w \sim p(0, \Sigma)$. To use a GP model for this task, the mean and covariance can be defined as:
\begin{equation} 
\label{eq:2_3_3}
\begin{aligned}
&  \mathbb{E}[f(x)] = \phi(x)^{T}\mathbb{E}[w] = 0 \\
&  \mathbb{E}[f(x)f(x')] =  \phi(x)^{T}\mathbb{E}[ww^{T}]\phi(x')=\phi(x)^{T}\Sigma\,\phi(x').
\end{aligned}
\end{equation}
Therefore, $f(x)$ and $f(x')$ are jointly Gaussian with zero mean and covariance given by $\phi(x)^{T}\Sigma\,\phi(x')$. The goal of GP model is to obtain posterior from observations and then to leverage it to estimate predictive distributions for test samples. Specifically, given training points $X$ and test points $X'$, the joint prior
distribution of the training outputs $f$  and the test outputs $f'$ is: 
\begin{gather}
\label{eq:2_3_4}
\begin{bmatrix} f \\ f'\end{bmatrix} \sim \mathcal{N}(\mathbf{0}, \begin{bmatrix} k(X, X) & k(X, X') \\  k(X', X) & k(X, X) \end{bmatrix}).
\end{gather} 
In order to get the posterior distribution, the functions $f$ should be consistent with the observations. Based on the Bayesian theorem, the posterior is given by:
\begin{gather}
\label{eq:2_3_5}
f'|X',X, f \sim \mathcal{N}( k(X', X)k(X, X)^{-1}f,\;  k(X', X')-k(X', X)k(X, X)^{-1}k(X, X')).
\end{gather} 
Therefore, the function values $f'$, corresponding to test inputs $X'$,  can be sampled from the posterior, based on the relationship between $X'$ and $X$. Such relationship is measured by covariance functions (i.e., kernel functions), such as squared exponential covariance function and periodic covariance function \cite{gpml}. From another perspective, the training data can be regarded as context data points, which provide as ``cornerstones'' for a target test point for comparison. To learn parameters, such as the length-scale in the squared exponential covariance function, a GP model aims to maximize the likelihood function over the observations. For more details regarding the regression task with noise and the classification task, please refer to the reference \cite{gpml}.

Nevertheless, GPs have some well-known drawbacks, which hinder their applications in practice. Firstly, they are notorious for their computational complexity. GPs have a time and space complexity of $\mathcal{O}(n^3)$  and $\mathcal{O}(n^2)$ respectively, where $n$ is the number of training instances. This makes them infeasible for large datasets. Various approximation techniques exist for larger datasets, but these typically involve a trade-off between speed and accuracy \cite{hensman2015scalable, flaxman2015fast, snelson2005sparse}. Secondly, the effectiveness of a GP model relies heavily on the choice of the kernel. Choosing an appropriate kernel that captures the underlying data structure is not always straightforward, and might require domain expertise. Finally, the performance of GPs also relies on the hyperparameters. These typically need to be determined through optimization of the log marginal likelihood, which can be computationally expensive. If the hyperparameters are not well-specified or the optimization gets stuck in local optima, the GP may perform poorly.

\section{Neural Processes}
Neural processes (NPs) are a probabilistic model that leverages neural networks to approximate stochastic process via finite-dimensional marginal distributions. Their working mechanism is closely related to the classical GP model. 
Formally, a stochastic process can be  defined as $\{F(x, \omega) : x \in \mathcal{X} \}$ over a probability space $(\Omega, \Sigma, \Pi)$ and an index set $\mathcal{X}$, 
where $F(\cdot\ , \ \omega)$ is a sample function
mapping $\mathcal{X}$ to another space $\mathcal{Y}$ for any point $\omega \in \Omega$. Therefore, for any finite sequence $x_{1:n}$, a marginal joint distribution function can be defined on the function values $F(x_1, \ \omega), F(x_2, \ \omega), \ldots, F(x_n, \ \omega)$, which satisfies two conditions given by the Kolmogorov Extension Theorem \cite{oksendal2003stochastic}:

 \vspace{1ex}

\noindent{\bf Exchangeability}: \emph{This condition indicates that the marginal joint distribution should remain unaffected by any permutation of the sequence.}
 
\noindent{\bf Consistency}: \emph{ This condition requires that the marginal joint distribution should remain unaffected when a part of the sequence is marginalized out. }

 \vspace{1ex}
 
With the two conditions, a stochastic process can be described by the marginal joint distribution function, namely: 
\begin{equation}
\small
\label{eq:joint1}
p(y_{1:n}|x_{1:n}) = \int \pi(\omega) p(y_{1:n}|F(\cdot\ , \ \omega), x_{1:n}) d\mu(\omega),
\end{equation}
where $\pi$ denotes density, namely,  $d\Pi = \pi d\mu$. Here, the function $F(\cdot\ , \ \omega)$ is determined by the kernels, which measure how all variables interact with each other.

\begin{figure*}[t]
\centering
\includegraphics[width=\linewidth]{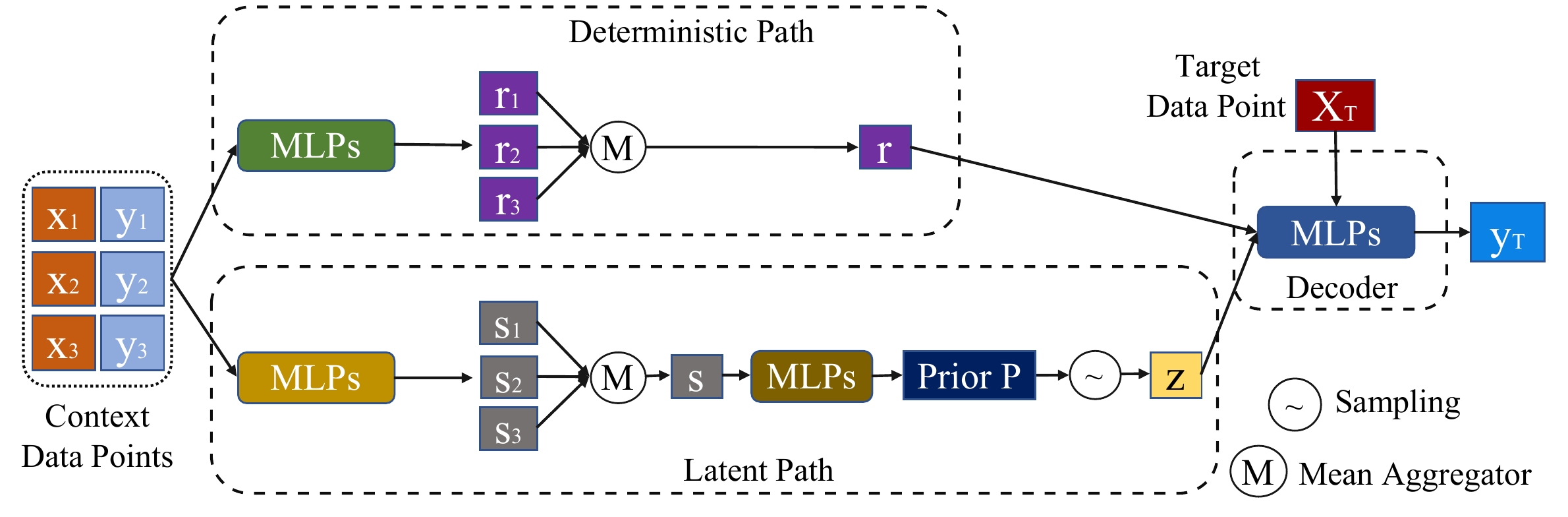}
\vspace{-4ex}
\caption{
Overview of the original NPs.}
\label{fig:np_ori} 
\end{figure*}

To approximate stochastic processes, NPs parameterize the function $F(\cdot\ , \ \omega)$ in the marginal joint distribution with neural networks and latent vectors.  Specifically, let $(\Omega, \Sigma)$ be $(\mathbb{R}^d, \mathcal{B}(\mathbb{R}^d))$,  where $\mathcal{B}(\mathbb{R}^d)$ denotes the {\it Borel} $\sigma${\it -algebra} of $\mathbb{R}^d$, and NPs use a latent vector $z \in \mathbb{R}^d$ sampled from a multivariate Gaussian distribution to govern the function $F(\cdot\ , \ \omega)$. Then, $F(x_i , \ \omega)$ can be replaced by $\phi(x_i, z)$, where $\phi(\cdot)$ denotes a neural network, and Eq.~(\ref{eq:joint1}) becomes:
\begin{equation}
\small
\label{eq:joint2}
p(y_{1:n}|x_{1:n}) = \int \pi(z)p(y_{1:n}|\phi(x_{1:n}, z), x_{1:n}) d\mu(z).
\end{equation} 
By doing this, NPs are capable of predicting and estimating uncertainty for each data point, circumventing the explicit access to kernel functions and comparisons of distances among distinct points. This capability renders them practical for application in real-world scenarios.

An instantiation of NPs is shown in Figure~\ref{fig:np_ori}. Specifically, in NPs, the training data are regarded as context points, while the test data are regarded as target points. NPs aim to make predictions for the target data points conditioned on context data points. There are two paths in neural processes, which are called the deterministic path and the latent path, respectively. From GPs, context points are needed to provide known information (``cornerstones'') for regression, and therefore, the deterministic path in Neural processes plays this role. In the deterministic path, data are first processed by Multi-Layer Perceptrons (MLPs), and the outputs are a set of feature representations. Then, a mean aggregator is used to fuse all features to get a single representation as context information. The goal of the latent path is to estimate a data-specific prior over the whole dataset, and therefore, it can be conditioned on both context and target points, since they all provide relevant information to the prior. In the original NPs \cite{garnelo2018neural}, only the context data points are used as the conditions for the latent path. Unlike MC dropout, the prior here is obtained based on the data, which is more helpful to get the final posterior than the zero-information prior. In the latent path, the input is first processed by MLPs, and then their feature representations are fused by a mean aggregator, whose output is further processed by another set of MLPs to get the prior. In practice, this is achieved by predicting mean and variance vectors, which is similar to the reparameterization trick in the variational autoencoder \cite{kingma2013auto}. There is also another decoder in the neural processes. This decoder, combined with $z$ sampled from the global prior and the global context information, mimics a sampled function, which is applied to a target point. The training objective of NPs is to maximize $p(y_{1:n}|x_{1:n})$, which can be implemented by maximizing its evidence lower-bound (ELBO).

The key difference between NPs and GPs is that the randomness of a sampled function in NPs is parameterized by the sampled $z$. Note that the mapping from a set of representations (before the mean aggregator) to the prior latent variables $z$ can be well captured after training,  which is guaranteed by the theorem proposed in Deep Sets \cite{zaheer2017deep}.

There are also many advanced NP models in recent years. For example, Kim {\it et al.} \cite{kim2019attentive} have observed that NPs tend to underfit the context set, which is caused by the mean aggregator giving equal weights to all the context points. To remedy this issue, they propose a new model, called attentive NP, which uses an attention mechanism to attend to relevant context points with respect to target predictions. Concerning that the application areas of NPs are time series or spatial data, the translation equivalence should be an important property, i.e., if the data are translated in time or space, the predictions should be translated correspondingly. This property was ignored in previous models, until Gordon {\it et al.} \cite{gordon2019convolutional} designed a new model called convolutional CNPs. Besides, concerning that the global latent variables are not flexible for encoding inductive biases, Louizos {\it et al.} \cite{louizos2019functional} employ local
latent variables along with a dependency structure among them instead, obtaining a new functional NP (FNP). Similarly, Lee {\it et al.} \cite{lee2020bootstrapping} also point out the limited flexibility of a single latent variable to model functional uncertainty, and they use a classic frequentist technique, namely,  bootstrapping, to model functional uncertainty, leading to a new NP variant, named Bootstrapping Neural Processes (BNPs). Bruinsma {\it et al.} \cite{bruinsma2021gaussian} propose a new NP variant called Gaussian NPs (GNPs), which not only involves translation
equivariance with Gaussian processes \cite{gpml}, but also provides universal approximation guarantees. 
Nguyen and Grover \cite{nguyen2022transformer} proposed a Transformer Neural Processes (TNPs) for uncertainty-aware meta-learning. They used a transformer architecture to parameterize a Neural Process by employing suitable modifications, such as removing positional embeddings and using a novel masking scheme. The learning objective of TNPs is to autoregressively maximize the conditional log-likelihood of the target points conditioned on the context points, allowing for an expressive parameterization of the predictive distribution.
Feng \emph{et al.}~\cite{feng2023latent} identify that the TNPs are computationally expensive in that they require quadratic computation in the number of context datapoints. To overcome this issue, they designed a Latent Bottlenecked Attentive Neural Processes (LBANP) that has querying computational complexity independent of the number of context data points, leading to constant computation per target data point. 
It should be noted that, in this thesis, the original NP model \cite{garnelo2018neural} is our focus. For more variants in terms of NPs,  please refer to the survey \cite{jha2022neural}.

The characteristics of NPs make them suitable for semi-supervised learning. For example, NPs possess the capability to make predictions for target points, conditioned upon the available context points. This particular feature is of great significance in SSL, where the learning process relies on a limited number of labeled samples to facilitate accurate predictions for unlabeled samples. Its approach to uncertainty quantification is also efficient, making it well-suited for large-scale datasets.

\chapter{Rethinking Bayesian Deep Learning Methods for Semi-Supervised Volumetric Medical Image Segmentation} 

In this chapter, I delve into a research field where uncertainty quantification is extensively applied, namely volumetric medical image segmentation. This is beneficial for understanding the current state of research in probabilistic models (e.g., BNNs) under semi-supervised learning (SSL), and such research brings real positive impact to human health. 
I pinpoint the existing problems in previous methods and propose a new framework along with its corresponding theoretical basis to solve these issues. This chapter is organized as follows: In Section~\ref{sec:3.1}, we introduce the background information and the contribution of this chapter. In Section~\ref{sec:3.2}, we review related methods. Section~\ref{sec:3.3} presents our proposed framework, GBDL, and its Bayesian formulation, followed by the experimental settings and results in Section~\ref{sec:3.4}. Sections~\ref{sec:3.5} and \ref{sec:3.6} provide a summary and discuss limitations, respectively.

\section{Introduction}
\label{sec:3.1}


In recent years, semi-supervised medical image segmentation has emerged as a prevalent research area, and several methods \cite{bai2017semi, zeng2021reciprocal, hang2020local, li2020shape,  huang20213d, bortsova2019semi, xie2020pairwise, luo2021semi, luo2021efficient, yu2019uncertainty, wang2020double, sedai2019uncertainty, shi2021inconsistency, wang2021tripled} have been proposed.\footnote{Note that we only consider previous methods that build their architectures with 3D CNNs, since our architecture is also based on 3D CNNs. Comparing our method with 2D-CNN-based methods is unfair to them.} 
Among these works, Bayesian deep-learning-based methods \cite{yu2019uncertainty, wang2020double, sedai2019uncertainty, shi2021inconsistency, wang2021tripled} are closely related to this chapter, as they are based on MC dropout \cite{gal2016dropout}, which is an approximation of BNNs. These methods are mainly built with the teacher-student architecture, which can be regarded as the discriminative model. Specifically, they initially build the model $P(Y|X)$ only from labeled data, and then apply the model to generate pseudo labels that are refined or rectified based on the epistemic uncertainty \cite{kendall2017uncertainties} provided by MC dropout for unlabeled ones. Then, the pseudo labels are further combined with unlabeled and labeled data to further train the overall architecture.

The works \cite{yu2019uncertainty, wang2020double, sedai2019uncertainty, shi2021inconsistency, wang2021tripled} mainly have two issues. Firstly, 
their models are built only with labeled data in the early stage of training and may overfit to them, since the quantity of labeled data is limited. Consequently, the new pseudo labels generated by the models might be inaccurate, which adversely impacts the subsequent training process. 
Secondly, their models are only partially based on Bayesian deep learning and are not designed under the Bayesian framework. Hence, they are
unable to give a Bayesian formulation with regard to their models, lacking a solid theoretical basis. 
As a result, some modules in their models are empirically designed, and the functions of those modules remain unclear.

We aim to fix these two problems by proposing a new generative model, named generative Bayesian deep learning (GBDL) architecture, as well as its corresponding full Bayesian formulation. 
Unlike  teacher-student-based architectures, GBDL estimates the joint probability distribution $P(X,Y)$ via both labeled and unlabeled data, based on which more reliable pseudo-labels can be generated, since the generative model is more faithful about the overall data distribution $P(X)$ and the intrinsic relationship between inputs and their corresponding labels (modeled by $P(X,Y)$), alleviating the overfitting problem.
Moreover, GBDL is entirely constructed under the Bayesian framework, and its related full Bayesian formulation lays a solid theoretical foundation, so that every part of GBDL has its corresponding probabilistic formulation and interpretation.

The main contributions are as follows:
\begin{itemize}[leftmargin=*, itemsep=0.5pt]
\item We propose a new generative Bayesian deep learning (GBDL) architecture. 
GBDL aims to capture the joint distribution of inputs and labels, which moderates the potential overfitting problem of the teacher-student architecture in previous works \cite{yu2019uncertainty, wang2020double, sedai2019uncertainty, shi2021inconsistency, wang2021tripled}. 

\item Compared with previous methods \cite{yu2019uncertainty, wang2020double, sedai2019uncertainty, shi2021inconsistency, wang2021tripled}, GBDL is completely designed under the Bayesian framework, and its full Bayesian formulation is also given here. Thus, it has a theoretical probabilistic foundation, and the choice of each module in GBDL and the construction of the loss functions are more mathematically grounded.

\item In extensive experiments,   GBDL outperforms previous methods with a healthy margin in terms of four evaluation indicators on three public medical datasets, illustrating that GBDL is a superior architecture for semi-supervised volumetric medical image segmentation.
\end{itemize}

\section{Related Work}
\label{sec:3.2}

In this section, we briefly review related Bayesian and other deep learning methods for semi-supervised volumetric medical image segmentation in the past few years.

\subsection{Bayesian Deep Learning Methods} 
Previous Bayesian deep learning methods simply use MC dropout as a tool to detect when and where deep models might make false predictions \cite{yu2019uncertainty, wang2020double, sedai2019uncertainty, shi2021inconsistency, wang2021tripled}. Specifically, an uncertainty-aware self-ensembling model \cite{yu2019uncertainty} was proposed  based on the teacher-student model and MC dropout. In particular, a teacher model and a student model were built, where the latter learnt from the former by minimizing the segmentation loss on the labeled data and the consistency loss on all the input data. 
MC dropout was leveraged to filter out the unreliable predictions and to preserve the reliable ones given by the teacher model, and the refined predictions can help the student model learn from unlabeled data. Based on this uncertainty-aware teacher-student model, a double-uncertainty weighted method \cite{wang2020double} was proposed, which takes both the prediction and the feature uncertainty into consideration for refining the predictions of the teacher model during training. Sedai {\it et al.}~\cite{sedai2019uncertainty} used MC dropout for training the teacher model as well, but they designed a novel loss function to guide the student model by adaptively weighting regions with unreliable soft labels to improve the final segmentation performance, rather than simply removing the unreliable predictions. Shi {\it et al.}~\cite{shi2021inconsistency} improved the way of uncertainty estimation by designing two additional models, which are called object conservation model and object-radical model, respectively. Based on these two models, certain region masks and uncertain region masks can be obtained to improve pseudo labels and to prevent the possible error propagation during training.  Since the multi-task learning can boost the performance of segmentation, and it has not been considered in previous works, Wang {\it et al.}~\cite{wang2021tripled} combined it into their architecture, and imposed the uncertainty estimation on all tasks to get a tripled-uncertainty that is further used to guide the training of their student model. 

\subsection{Other Deep Learning Methods} 
Concerning other deep learning methods proposed recently, they are mainly classified as three categories: new training and learning strategies \cite{bai2017semi, zeng2021reciprocal}, shape-aware or structure-aware based methods \cite{hang2020local, li2020shape, huang20213d}, and consistency regularization based methods \cite{bortsova2019semi, xie2020pairwise, luo2021semi, luo2021efficient}. 

The new training and learning strategies aim  to gradually improve the quality of pseudo labels during training. For example, Bai {\it et al.}~\cite{bai2017semi} proposed an iterative training strategy, in which the network is first trained on labeled data, and then it predicts pseudo labels for unlabeled data, which are further combined with the labeled data to train the network again. During the iterative training process, a conditional random field (CRF) \cite{krahenbuhl2011efficient} was used to refine the pseudo labels. 
Zeng {\it et al.} \cite{zeng2021reciprocal} proposed a reciprocal learning strategy for their teacher-student architecture. The strategy contains a feedback mechanism for the teacher network via observing how pseudo labels would affect the student, which is omitted in previous teacher-student based models. 

The shape-aware or structure-aware based methods encourage the network to explore complex geometric information in medical images. For instance, 
Hang {\it et al.}~\cite{hang2020local} proposed a local and global structure-aware entropy-regularized mean teacher (LG-ER-MT) architecture for semi-supervised segmentation. It extracts local spatial structural information by calculating inter-voxel similarities within small volumes and global geometric structure information by utilizing weighted self-information. Li {\it et al.}~\cite{li2020shape} introduced a shape-aware semi-supervised segmentation strategy that integrates a more flexible geometric representation into the network for boosting the performance. Huang {\it et al.} \cite{huang20213d} designed a 3D Graph Shape-aware Self-ensembling Network (3D Graph-$S^2$Net), which is composed by a multi-task learning network and a graph-based module. The former performs the semantic segmentation task and predicts the signed distance map that encodes
rich features of object shape and surface, while the latter explores co-occurrence relations and diffuse information between these two tasks.  

The consistency-regularization-based methods try to add different consistency constraints for unlabeled data. In particular, 
Bortsova {\it et al.}~\cite{bortsova2019semi} proposed a model that is implemented by a Siamese architecture with two identical branches to receive differently transformed versions of the same images. Its target is to learn the consistency under different transformations of the same input. 
Luo {\it et al.} \cite{luo2021semi} designed a novel dual-task-consistency
model, whose basic idea is to encourage consistent predictions of the same input under different tasks. Xie {\it et al.}~\cite{xie2020pairwise} proposed a pairwise relation-based semi-supervised model for gland segmentation on histology tissue images. In this model, a supervised segmentation network (S-Net) and an unsupervised pairwise relation network (PR-Net) are built. The PR-Net learns both semantic consistency and image representations from each pair of images in the feature space for improving the segmentation performance of S-Net. Luo {\it et al.} \cite{luo2021efficient} proposed a 
pyramid (i.e., multi-scale) architecture that encourages the predictions of an unlabeled input at multiple scales to be consistent, which serves as a regularization for unlabeled data.

\section{Methodology}
\label{sec:3.3}

In this section, we start from giving the Bayesian formulation of GBDL architecture. Thereafter, the architecture and related loss functions are introduced in detail.

\subsection{Bayesian Formulation}

\paragraph{Learning Procedure.} The objective of semi-supervised volumetric medical image segmentation is to learn a segmentation network, whose weights are denoted by $W$, with partially labeled data, and the Bayesian treatment of this target is to learn the posterior distribution, namely,  $P(W|X, Y_{L})$.  
Now, we use $X$ to denote the input volumes, which contains labeled input volumes ($X_L$) and unlabeled input volumes ($X_U$), i.e., $X=\{ X_L, X_U \}$, and we use $Y=\{ Y_L, Y_U \}$ to denote the ground-truth labels with respect to $X$. 
Each label in $Y$ is a voxel-wise segmentation label map that has the same shape as its corresponding input volume. Note that $Y_U$ represents the ground-truth labels of $X_U$, which are not observed in the training data. 

The whole learning procedure of $P(W|X, Y_{L})$ can be written as:
\begin{equation}
\begin{aligned}
\label{eqn:overall}
& P(W|X, Y_{L})  = \iiint P(W|X, Y)P(X, Y|Z)P(Z|X, Y_{L})dZdXdY,
\end{aligned}
\end{equation}
where $Z$ is the latent representation that governs the  joint distribution of $X$ and $Y$, denoted by $P(X,Y|Z)$. To estimate the joint distribution, one should learn $Z$ from $X$ and $Y_{L}$. Concerning that  Eq.~(\ref{eqn:overall}) is intractable, we take the MC approximation of it as follows:
\begin{equation}
\label{eqn:monte_appro}
\begin{aligned}
 P(W|X, Y_{L}) = \frac{1}{MN}\sum\nolimits_{i=0}^{N-1}\sum\nolimits_{j=0}^{M-1}P(W|X_{(i,j)}, Y_{(i,j)}).
\end{aligned}
\end{equation}
In this approximation, $M$ latent representations $Z$ are drawn from $P(Z|X, Y_{L})$. Then, $N$ pairs of input volumes and labels are obtained from $P(X, Y|Z)$, which are further used to get the posterior distribution. Thus, to generate $X$ and $Y$, one should obtain the joint distribution $P(X, Y)$, i.e., estimating the distribution of $Z$. Overall, the learning procedure contains two steps: 
\begin{itemize}[leftmargin=*, itemsep=0.5pt]
\item Learning the distribution of $Z$ that governs the joint distribution $P(X, Y)$. 

\item Learning the posterior distribution $P(W|X, Y)$ based on $X$ and $Y$ sampled from $P(X, Y)$.
\end{itemize}

\paragraph{Inference Procedure.} The inference procedure can be formulated as follows:
\begin{equation}
\begin{aligned}
\label{eqn:testing}
&P(Y_{pred}|X_{test}, X, Y_{L})  =  \int P(Y_{pred}|X_{test}, W)P(W|X, Y_{L}) dW,
\end{aligned}
\end{equation}
where $X_{test}$ and $Y_{pred}$ denote the test input and the predicted result, respectively. The posterior distribution $P(W|X, Y_{L})$ is learnt based on $X$ and $Y$ sampled from $P(X, Y)$, and thus the posterior can also be replaced by $P(W|X, Y)$. Since Eq.~(\ref{eqn:testing}) is intractable as well, its MC approximation can be written as:
\begin{equation}
\label{eqn:testing_monte}
\begin{aligned}
&P(Y_{pred}|X_{test}, X, Y_{L}) = \frac{1}{T}\sum\nolimits_{i=0}^{T-1}P(Y_{pred}|X_{test}, W_i),
\end{aligned}
\end{equation}
where $T$ models are drawn from $P(W|X, Y)$, which is usually implemented via MC dropout \cite{gal2016dropout} with $T$ times feedforward passes. For each model ($W_i$) sampled from $P(W|X, Y)$, a prediction result can be obtained, and the final prediction of $X_{test}$ can be calculated by averaging the $T$ results. In addition, 
we get the epistemic uncertainty by calculating the entropy of these predictions.

\begin{figure*}[t]
\centering
\includegraphics[width=\linewidth]{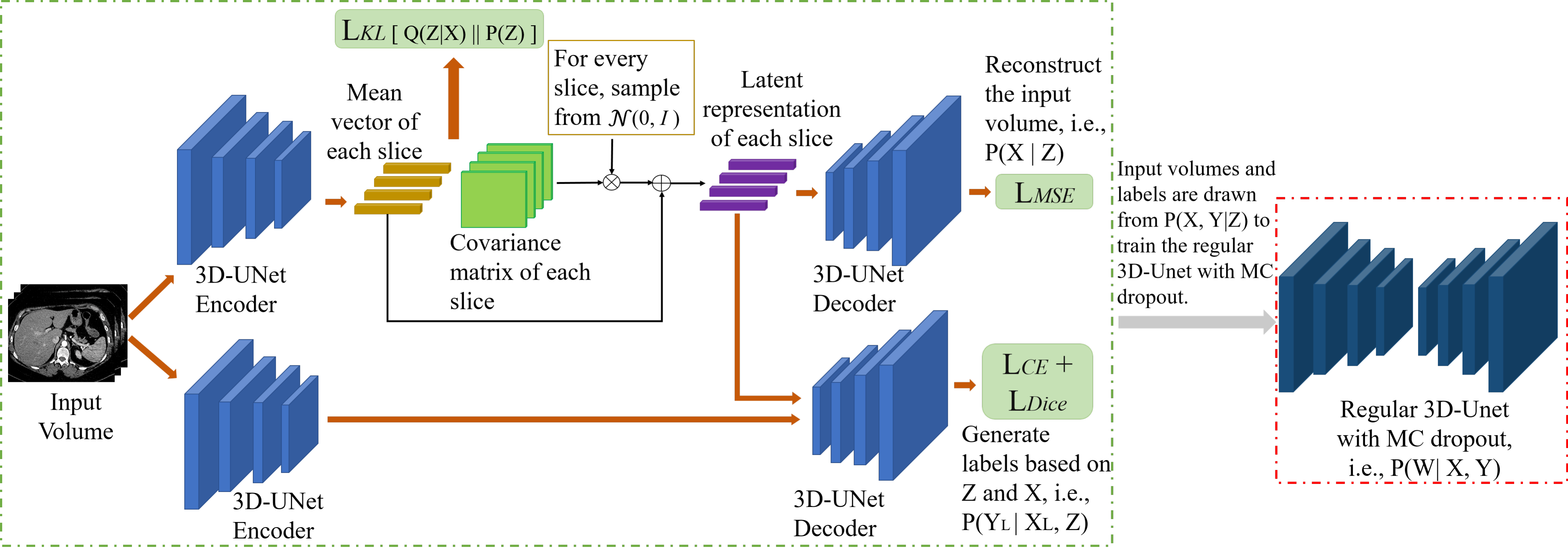} 
\caption{GBDL for semi-supervised volumetric medical image segmentation, including a latent representation learning (LRL) architecture (in the green dotted box) and a regular 3D-UNet with MC dropout (in the red dotted box). Only the regular 3D-UNet with MC dropout is used during testing. For simplicity, the shortcut connections between the paired 3D-UNet encoder and decoder are omitted.} 
\label{fig:framework}
\end{figure*} 

\subsection{GBDL and Loss Functions} 
\label{sec:gbdl} 

Under the guidance of the given Bayesian formulation, we start to introduce the details of GBDL and loss functions. 
Concerning the two steps mentioned in the learning procedure above, GBDL has a latent representation learning (LRL) architecture and a regular 3D-UNet with MC dropout, which are shown in the green dotted box and the red dotted box in Figure~\ref{fig:framework}, respectively. LRL is designed to learn the distribution of $Z$ and to capture the joint distribution $P(X, Y)$, while the regular 3D-UNet with MC dropout is the Bayesian deep model used for parameterizing  the posterior distribution $P(W|X, Y)$. 

As for LRL, we first assume a variational probability distribution $Q(Z)$ to represent the distribution of the latent representation $Z$. 
We follow the conditional variational auto-encoder (cVAE) \cite{sohn2015learning} to use the input to modulate $Z$, and thus, $Q(Z)$ can be rewritten as $Q(Z|X)$. 
Note that we focus on using 3D CNNs to process volumetric medical data, and therefore, 
each input volume contains several slices, $Q(Z|X)$ is actually the joint distribution of all these slices, i.e., $Q(Z|X) = q(z_0, z_1, z_2, ..., z_{n-1}|x_0, x_1, x_2, ..., x_{n-1})$, where $n$ is the number of slices in an input volume, $x_i$ and $z_i$ are a slice and its latent representation,  respectively. To obtain the latent representation of each slice $z_i$, a 3D-UNet encoder\footnote{All 3D upsampling and downsampling layers used in the 3D-UNet encoder and the 3D-UNet decoder only perform on the spatial size, and therefore the depth of the input volumes keeps unchanged, and we can obtain the latent representation of each slice. More details about the network configuration are given in the Appendix A.} is used in LRL. As the downsampling layers of the 3D-UNet encoder do not perform on the depth, and the 3D convolutional layers used in the encoder are of the kernel size $3$, the padding size $1$, and the stride $1$ (shown in the Appendix A), each $z_i$ is actually conditioned on $n_{rf}$ slices, where $n_{rf}$ is the total receptive field along the depth dimension, which is determined by the number of 3D convolutional layers in the encoder.  
For ease of calculation, we assume that the distribution of the latent representation of each slice follows a multivariate Gaussian distribution and that the latent representations of slices are independent from each other. 
It is worth noting that such independence assumption on the latent representations of slices is reasonable, as in this way, the latent representation of each slice just rests on $n_{rf}$ slices, which is consistent with the fact that each slice is closely relevant to its neighboring slices and distant slices may have no contribution to the slice. 
Therefore, based on the independence assumption, $Q(Z|X)$ can be decomposed into $\prod_{i=0}^{n-1} q(z_i| x_{(n_{rf})})$, where $x_{(n_{rf})}$ denotes the input slices that contribute to the $z_{i}$. 
After obtaining the latent representation $Z$, we can write the evidence lower bound (ELBO) as follows (with proof in the Appendix A): 
\begin{equation}
\begin{aligned}
\label{eqn:elbo}
 &logP(X,Y) \geq  \mathbb{E}_{Q}[logP(Y| X, Z) + logP(X|Z)] - \mathbb{E}_{Q}[log (\frac{Q(Z|X)} {P(Z)})],
\end{aligned}
\end{equation}
where $\mathbb{E}_{Q}$ denotes the expectation over $Q(Z|X)$.  Thus, the learning objective is to maximize the ELBO (Eq.~(\ref{eqn:elbo})), which is achieved by maximizing $ \mathbb{E}_{Q}[logP(Y| X, Z) + logP(X|Z)]$ and minimizing~$\mathbb{E}_{Q}[log({Q(Z|X)}/{P(Z)})]$.

Firstly, to maximize $P(X|Z)$, another 3D-UNet decoder is used to take $z_i$ as input and the mean square error ($L_{MSE}$) is used as the loss function, which aims to reconstruct each slice of the input volume. Secondly, to maximize the probability $P(Y| X, Z)$, another branch can be built, which contains a 3D-UNet encoder and a 3D-UNet decoder as well. Since only parts of data have labels, $P(Y| X, Z)$ can also be rewritten as $P(Y_L| X_L, Z)$. 
In this branch, the 3D-UNet encoder receives $X_L$ to extract features that are further combined with their corresponding $Z$ to be input to the 3D-UNet decoder. Then, $Y_L$ is leveraged to calculate the dice loss $L_{Dice}$ \cite{milletari2016v} and cross-entropy loss $L_{CE}$ with $X_L$ to maximize the probability $P(Y_L| X_L, Z)$. 
Thirdly, minimizing $\mathbb{E}_{Q}[log({Q(Z|X)}/{P(Z)})]$ is to use $Q(Z|X)$ to approximate the prior distribution of $Z$, and in most cases, assuming that $Q(Z|X)$ and $P(Z)$ follow the same probability distribution can make the computation convenient.
The following theorem (with proof in the Appendix A) is useful for constructing the distribution of $Q(Z|X)$.

\smallskip
{\bf Theorem 3.3.2.1.} \emph{The product of any number $n\ge 1$ of multivariate Gaussian probability density functions with precision $\Lambda_i$ and mean $\mu_i$ results in an unnormalized Gaussian curve with precision $\Lambda_{*}$ $=$ $\sum_{i=0}^{n-1}\Lambda_i$ and mean $\mu_{*}\,=$ $\Lambda_{*}^{-1} (\sum_{i=0}^{n-1}\Lambda_i\mu_i)$.}
\smallskip

Although an unnormalized Gaussian curve (denoted $f(x)$) is obtained by multiplying several multivariate Gaussian PDFs, a Gaussian distribution can still be obtained by simply normalizing $f(x)$ with $\int f(x)dx$ that does not contain $x$. Therefore, based on {Theorem 1}, we can suppose that $Q(Z|X)$ follows a multivariate Gaussian distribution. In addition, we also assume that the prior distribution $P(Z)$ follows a multivariate standard normal distribution. To minimize $\mathbb{E}_{Q}[log ({Q(Z|X)}/{P(Z)})]$, we propose a new loss function  as follows (with proof in the Appendix A). 

\smallskip
{\bf Corollary 3.3.2.1.} \emph{Minimizing $\mathbb{E}_{Q}[log({Q(Z|X)}/{P(Z)})]$ is equivalent to minimizing}
\begin{equation}
\begin{aligned} 
\label{eqn:kl1}
 L_{KL[Q(Z|X)||P(Z)]}  = \; & \frac{1}{2} \cdot (
-log\ det [(\sum\nolimits_{i=0}^{n-1}\Lambda_i)^{-1}] +  tr[(\sum\nolimits_{i=0}^{n-1}\Lambda_i)^{-1}] \ +  \\
 & (\sum\nolimits_{i=0}^{n-1}\Lambda_i\mu_i)^T (\sum\nolimits_{i=0}^{n-1}\Lambda_i)^{-2} (\sum\nolimits_{i=0}^{n-1}\Lambda_i\mu_i)
- D), 
\end{aligned}
\end{equation}
\emph{where $n$ denotes the number of slices, and $D$ denotes the number of dimensions of mean vectors.}
\smallskip

To summarize, maximizing ELBO (Eq.~(\ref{eqn:elbo})) is equivalent to optimizing LRL and minimizing the loss function:
\begin{equation}
\begin{aligned}
\label{eqn:ELBO_loss}
& L_{ELBO} =   \lambda_1 L_{CE} + \lambda_2 L_{Dice} + \lambda_3  L_{MSE} + \lambda_4  L_{KL[Q(Z|X)||P(Z)]},
\end{aligned}
\end{equation}
where $\lambda_1$, $\lambda_2$, $\lambda_3$, and $\lambda_4$  are coefficients. 
{Significantly}, cVAE \cite{sohn2015learning} has some similar properties when compared with our LRL architecture, and it is important to emphasize the differences between cVAE-based models and our LRL, which mainly contain three aspects:

\begin{itemize}[leftmargin=*, itemsep=0.5pt]
\item Different ELBO: cVAE is mainly used for generation tasks, in which the data distribution $P(X)$ is considered, and labels $Y$ are not involved into the ELBO of cVAE. In contrast, our LRL concerns  the joint distribution of $X$ and $Y$, leading to a different ELBO (Eq.~(\ref{eqn:elbo})).

\item Independent Components Assumption: 
cVAE assumes that the components in the latent representation of an input image are independent from each other. As a result, the encoder of cVAE only gives a mean vector and a variance vector for a latent representation\footnote{In practice, the encoder of cVAE gives log-variance vectors instead of variance vectors.}. However, our LRL does not follow this assumption for each slice, and the encoder gives a mean vector and a covariance matrix for the latent representation of each slice\footnote{Although the independent components assumption is widely used, it could be too strong and may degrade the performance.}.

\item Joint Variational Distribution: cVAE is mainly used for 2D inputs. In contrast, our LRL is specifically designed for 3D-CNN-based architectures for processing volumetric inputs. Therefore, since each input volume contains several slices, the variational distribution becomes a joint distribution over their latent representations, resulting in a different formulation of Kullback–Leibler (KL) divergence between the joint variational distribution and the prior distribution (Eq.~(\ref{eqn:kl1})).
\end{itemize}

Once LRL is trained well, the joint distribution of $X$ and $Y$ will be captured, and 
the second step is to learn the posterior distribution $P(W|X, Y)$. Therefore, input volumes and corresponding labels are generated from the fixed LRL to train the regular 3D-UNet with MC dropout with two loss functions that are widely used in medical image segmentation, namely,  the dice loss $L_{Dice}$ and the cross-entropy loss $L_{CE}$. Here, we use $L_{Seg}$ to denote their summation: 
\begin{equation}
\label{eqn:segment_loss}
L_{Seg} =  \beta_1 L_{CE} + \beta_2 L_{Dice},
\end{equation}
where $\beta_1$ and $\beta_2$ are the coefficient of the two loss terms in $L_{Seg}$. In the real implementation, the new generated pseudo labels will be combined with the unlabeled data and labeled data to train the regular 3D-UNet with MC dropout. According to  Eq.~(\ref{eqn:testing_monte}), only the the regular 3D-UNet with MC dropout is used in the test phase. 
In addition, as we do a full Bayesian inference during testing, it is convenient to obtain the voxel-wise epistemic uncertainty for each predicted result by calculating the entropy ($-\sum_{c=0}^{C-1}p_clog_2p_c$) voxel-by-voxel, where $C$ is the number of classes.

\section{Experiments}
\label{sec:3.4}
We now report on experiments of the proposed GBDL on three public medical benchmarks. To save space, implementation details are shown in the Appendix A.  

\subsection{Datasets}
\label{sec:dataset}

The \textit{KiTS19} dataset 
is a kidney tumor segmentation dataset, which has 210 labeled 3D computed tomography (CT) scans for training and validation. We followed the settings in previous works \cite{wang2020double, fang2020dmnet} to use 160 CT scans for training and 50 CT scans for testing. The 3D scans centering at the kidney region were used, and we used the soft tissue CT window range of [-100, 250] HU for the scans.

The \textit{Atrial Segmentation Challenge} dataset \cite{xiong2020global} includes 100 3D gadolinium-enhanced magnetic resonance imaging scans (GE-MRIs) with labels. We followed previous works \cite{wang2020double, xiong2020global, luo2021semi, huang20213d, li2020shape} to use 80 samples for training and the other 20 samples for testing.

The \textit{Liver Segmentation} dataset has been released by the MICCAI 2018 medical segmentation decathlon challenge. 
The dataset has 131 training and 70 testing points, and the ground-truth labels of the testing data are not released. Therefore, we used those 131 CTs in our experiments, in which 100 CT scans and 31 CT scans were used for training and testing respectively. The 3D scans centering at liver regions were utilized, and we used the soft tissue CT window range of [-100, 250] HU for the scans.

\subsection{Evaluation Metrics}
Four evaluation metrics are used as our evaluation indicators, namely, \textit{Dice Score}, \textit{Jaccard Score}, \textit{95$\%$ Hausdorff Distance (95HD)}, and \textit{Average SurfaceDistance~(ASD)}. Dice Score and Jaccard Score mainly compute the percentage of overlap between two object regions. ASD computes the average distance between the boundaries of two object regions, while 95HD measures the closest point distance between two object regions.

\subsection{Ablation Studies}
\label{sec:ablation study}

We conducted  ablation studies on the KiTS19 nd the Atrial Segmentation Challenge dataset.

Firstly, we evaluated the baseline model, i.e., the regular 3D-UNet with MC dropout that is trained with a limited number of labeled data, and we also evaluated the upper bound of the performance by utilizing all data to train the model. As Table~\ref{tab:abla_kits_LA_1} shows, when the number of labeled data is decreased, the performance of the regular 3D-UNet drops drastically.

\begin{table*}
\centering
\resizebox{\textwidth}{!}{
\begin{tabular}{@{}c|c|cccc|c|cccc@{}}
 \toprule[1pt]
 \multirow{3}{*}{} &  \multicolumn{5}{c|}{KiTS19} & \multicolumn{5}{c}{Atrial Segmentation Challenge} \\
  \cline{2-11}
  &  Scans used  &\multicolumn{4}{c|}{Metrics} &  Scans used &\multicolumn{4}{c}{Metrics}\\
 \cline{2-11}
 & L vs. U  & Dice $\uparrow$ & Jaccard  $\uparrow$ & 95HD $\downarrow$ & ASD $\downarrow$ & L  vs. U  & Dice $\uparrow$ & Jaccard  $\uparrow$ & 95HD $\downarrow$ & ASD $\downarrow$ \\
 \hline
\multirow{3}{*}{\shortstack{3D-UNet w/ \\   MC dropout}} & 160 vs. 0 &  0.940 {\scriptsize($\pm$0.004)}  & 0.888 {\scriptsize($\pm$0.005)}   & 3.66 {\scriptsize($\pm$0.26)}  &  0.78 {\scriptsize($\pm$0.11)} & 80 vs.  0  & 0.915 {\scriptsize($\pm$0.006)}  & 0.832 {\scriptsize($\pm$0.007)}  & 3.89 {\scriptsize($\pm$0.17)} &  1.25 {\scriptsize($\pm$0.14)}   \\
  & 16 vs.  0 & 0.838 {\scriptsize($\pm$0.044)}  & 0.745 {\scriptsize($\pm$0.051)}  & 11.36 {\scriptsize($\pm$1.89)} & 3.30 {\scriptsize($\pm$0.78)} & 16 vs.  0  & 0.834 {\scriptsize($\pm$0.044)}  & 0.720 {\scriptsize($\pm$0.049)} & 8.97 {\scriptsize($\pm$1.27)} & 2.48 {\scriptsize($\pm$0.64)} \\
 & 4 vs. 0 & 0.710 {\scriptsize($\pm$0.051)} & 0.578 {\scriptsize($\pm$0.065)} & 19.56 {\scriptsize($\pm$1.95)} & 6.11 {\scriptsize($\pm$0.92)}  & 8  vs.   0  & 0.801  {\scriptsize($\pm$0.057)}  & 0.676 {\scriptsize($\pm$0.051)} & 11.40 {\scriptsize($\pm$1.53)} & 3.27 {\scriptsize($\pm$0.71)} \\ 
\hline
Volume-based LRL    &  16  vs.  144  & 0.892 {\scriptsize($\pm$0.016)} & 0.823 {\scriptsize($\pm$0.021)} & 7.47 {\scriptsize($\pm$0.39)} & 1.88 {\scriptsize($\pm$0.27)} &  16  vs. 64  &  0.871 {\scriptsize($\pm$0.010)} & 0.784 {\scriptsize($\pm$0.017)} & 5.28 {\scriptsize($\pm$0.37)} & 1.89 {\scriptsize($\pm$0.16)} \\
\cline{1-1}
IC-LRL    &  16  vs.  144  & 0.900 {\scriptsize($\pm$0.011)} & 0.828 {\scriptsize($\pm$0.012)} & 6.99 {\scriptsize($\pm$0.26)} & 1.75 {\scriptsize($\pm$0.22)} &  16   vs. 64  &  0.882 {\scriptsize($\pm$0.014)} & 0.796 {\scriptsize($\pm$0.026)} & 4.69 {\scriptsize($\pm$0.22)} & 1.66 {\scriptsize($\pm$0.12)} \\
\cline{1-1}
 GBDL (LRL) &   16  vs.  144  &  {\bf 0.911} {\scriptsize($\pm$0.010)}     &  {\bf 0.840} {\scriptsize($\pm$0.013)} & {\bf 6.38} {\scriptsize($\pm$0.37)} & {\bf 1.51} {\scriptsize($\pm$0.22)} &  16  vs.  64  &    {\bf 0.894} {\scriptsize($\pm$0.008)}  & {\bf 0.822} {\scriptsize($\pm$0.011)} & {\bf 4.03} {\scriptsize($\pm$0.41)} & {\bf 1.48} {\scriptsize($\pm$0.14)} \\
\hline 
Volume-based LRL    &  4  vs.  156  &   0.883 {\scriptsize($\pm$0.011)}    &  0.810 {\scriptsize($\pm$0.014)}  & 8.32 {\scriptsize($\pm$0.42)} &  1.99 {\scriptsize($\pm$0.24)} &  8  vs. 72 & 0.865 {\scriptsize($\pm$0.017)} & 0.773 {\scriptsize($\pm$0.013)}& 6.81 {\scriptsize($\pm$0.33)} & 2.49 {\scriptsize($\pm$0.13)}  \\
\cline{1-1}
IC-LRL    &  4  vs. 156  &   0.889 {\scriptsize($\pm$0.014)}    &  0.814 {\scriptsize($\pm$0.014)}  & 8.01 {\scriptsize($\pm$0.31)} &  2.01 {\scriptsize($\pm$0.12)} &  8  vs. 72 & 0.871 {\scriptsize($\pm$0.009)} & 0.779 {\scriptsize($\pm$0.024)}& 5.96 {\scriptsize($\pm$0.22)} & 1.97 {\scriptsize($\pm$0.17)}  \\
\cline{1-1}
 GBDL (LRL)   &  4  vs. 156  &  {\bf 0.898}   {\scriptsize($\pm$0.008)}   &   {\bf 0.821} {\scriptsize($\pm$0.011)} & {\bf 6.85} {\scriptsize($\pm$0.44)} & {\bf 1.78} {\scriptsize($\pm$0.23)} &    8  vs.  72  &   {\bf 0.884} {\scriptsize($\pm$0.007)} & {\bf 0.792} {\scriptsize($\pm$0.012)} & {\bf 5.89} {\scriptsize($\pm$0.31)} & {\bf 1.60} {\scriptsize($\pm$0.15)}  \\
  \bottomrule[1pt]
  \end{tabular}}
 \vspace{-1.0ex}
 \caption{Ablation studies of different LRL variants on the KiTS19 dataset and the Atrial Segmentation Challenge dataset. For each of our results, the ``mean {\scriptsize($\pm$std)}" is reported. ``L vs. U'' denotes the number of labeled data versus that of unlabeled data.
 The first row shows the upper bound performance, i.e., all training data with their labels are used.} 
 \label{tab:abla_kits_LA_1}
 \end{table*}

Second, we conducted ablation studies on our VAE-like structure. In our experiments, we found that if the number of dimensions in the latent vectors is too small, the GBDL might fail to capture all essential features, missing some important anatomical structures when performing reconstruction. Therefore, we set the dimensions to 256 in our GBDL, and we explored different VAE bottleneck structures in this paragraph. 
As for an input volume, LRL assumes that the latent representation of each slice follows a multivariate Gaussian distribution, and therefore, the latent representation of the whole volume is sampled from a joint Gaussian distribution that is obtained by multiplying the Gaussian distribution PDFs of those slices. 
However, LRL can also be designed to learn the latent representation of the whole volume, i.e., estimating the joint Gaussian distribution directly, and we denote this variant as ``volume-based LRL''.  Compared to the original LRL, this variant is much closer to a 3D-CNN-based cVAE, since each input volume is treated as a whole, and the volume-based LRL represents the whole volume with a latent representation, similar to the original cVAE that represents the whole input image with a latent representation. 
In this case, 
$L_{KL[Q(Z|X)||P(Z)]}$ degenerates to $1/2\cdot(-log\, det [\Lambda_v^{-1}] +  tr[\Lambda_v^{-1}] +   \mu_v ^T \mu_v - {D})$, where $\mu_v$ and $\Lambda_v$ are the mean vector and the precision matrix of the joint Gaussian distribution, respectively. 
We evaluated such volume-based LRL, and the results are displayed in Table~\ref{tab:abla_kits_LA_1}. For a fair comparison, the $\mu_v$ and $\Lambda_v$ are of the same number of dimensions as the $\mu_i$ and $\Lambda_i$ in the original LRL. The results show that the volume-based LRL performs worse than the original LRL that is presented in Section~\ref{sec:gbdl}, demonstrating that the original LRL can better model the joint distribution. Based on this observation, we hypothesize that keeping a difference among slices is important, since slices and their corresponding segmentation masks are different from each other. The original LRL adds perturbation (random noise) to each slice, which is beneficial to keeping such a difference. However, the volume-based LRL only adds random noise to the latent representation of the whole volume, which goes against to keeping such a difference. 
\begin{table*}
\centering
\resizebox{\textwidth}{!}{
\begin{tabular}{@{}c|c|cccc|c|cccc@{}}
 \toprule[1pt]
 \multirow{3}{*}{} &  \multicolumn{5}{c|}{KiTS19} & \multicolumn{5}{c}{Atrial Segmentation Challenge} \\
  \cline{2-11}
  &  Scans used  &\multicolumn{4}{c|}{Metrics} &  Scans used &\multicolumn{4}{c}{Metrics}\\
 \cline{2-11}
 & L vs. U  & Dice $\uparrow$ & Jaccard  $\uparrow$ & 95HD $\downarrow$ & ASD $\downarrow$ &  L vs. U  & Dice $\uparrow$ & Jaccard  $\uparrow$ & 95HD $\downarrow$ & ASD $\downarrow$ \\
 \hline
w/o $L_{MSE}$     &  16  vs. 144 &  0.877 {\scriptsize($\pm$0.014)}  & 0.789 {\scriptsize($\pm$0.019)} & 8.89 {\scriptsize($\pm$0.33)} &  1.96 {\scriptsize($\pm$0.19)}   &  16  vs.  64 &  0.873 {\scriptsize($\pm$0.012)}   & 0.781 {\scriptsize($\pm$0.011)}   & 4.76 {\scriptsize($\pm$0.46)}  & 1.66 {\scriptsize($\pm$0.17)}    \\
\cline{1-1}
w/o $L_{KL}$    &  16   vs.  144 &  0.868  {\scriptsize($\pm$0.013)}   & 0.776 {\scriptsize($\pm$0.009)}   & 8.99 {\scriptsize($\pm$0.29)}  & 2.03 {\scriptsize($\pm$0.17)}  &  16  vs.  64  &  0.876 {\scriptsize($\pm$0.014)}   & 0.775 {\scriptsize($\pm$0.016)}  & 5.12 {\scriptsize($\pm$0.55)}  & 1.87 {\scriptsize($\pm$0.15)}  \\
\cline{1-1}
Shared Encoder &   16   vs.  144  &  0.854 {\scriptsize($\pm$0.019)}   & 0.742 {\scriptsize($\pm$0.021)}   & 10.04 {\scriptsize($\pm$0.67)}  & 2.56 {\scriptsize($\pm$0.27)}  &  16 vs.  64  &  0.862 {\scriptsize($\pm$0.021)}    & 0.749 {\scriptsize($\pm$0.025)}  & 6.33 {\scriptsize($\pm$0.73)}  & 2.21 {\scriptsize($\pm$0.31)}  \\
\cline{1-1}
GBDL   &  16  vs.  144  &  {\bf 0.911} {\scriptsize($\pm$0.010)}     &  {\bf 0.840} {\scriptsize($\pm$0.013)} & {\bf 6.38} {\scriptsize($\pm$0.37)} & {\bf 1.51} {\scriptsize($\pm$0.22)} &  16   vs.  64  &    {\bf 0.894} {\scriptsize($\pm$0.008)}  & {\bf 0.822} {\scriptsize($\pm$0.011)} & {\bf 4.03} {\scriptsize($\pm$0.41)} & {\bf 1.48} {\scriptsize($\pm$0.14)} \\
\hline
w/o $L_{MSE}$     &  4   vs.  156  &  0.864 {\scriptsize($\pm$0.016)}   & 0.769 {\scriptsize($\pm$0.014)}   & 9.03 {\scriptsize($\pm$0.39)}  & 2.05 {\scriptsize($\pm$0.18)}  &  8   vs.  72 & 0.867 {\scriptsize($\pm$0.011)}   & 0.771 {\scriptsize($\pm$0.017)}  & 4.96 {\scriptsize($\pm$0.34)} & 1.85 {\scriptsize($\pm$0.20)} \\
\cline{1-1}
w/o $L_{KL}$    &  4   vs.  156  &   0.860 {\scriptsize($\pm$0.017)}   & 0.769 {\scriptsize($\pm$0.019)}   & 9.32 {\scriptsize($\pm$0.46)}  & 2.23  {\scriptsize($\pm$0.31)}   &  8   vs.  72  &  0.854 {\scriptsize($\pm$0.015)}  & 0.762 {\scriptsize($\pm$0.018)}  & 5.33 {\scriptsize($\pm$0.42)}  & 2.12 {\scriptsize($\pm$0.33)}  \\
\cline{1-1}
Shared Encoder &  4   vs.  156   &  0.843 {\scriptsize($\pm$0.021)}   & 0.734 {\scriptsize($\pm$0.024)}  & 11.08 {\scriptsize($\pm$0.67)}  & 3.46 {\scriptsize($\pm$0.44)}   &  8  vs.  72  &  0.831 {\scriptsize($\pm$0.019)}   & 0.724 {\scriptsize($\pm$0.012)}  & 7.19 {\scriptsize($\pm$0.68)} & 2.86 {\scriptsize($\pm$0.41)} \\
\cline{1-1}
GBDL   &  4   vs.  156  &  {\bf 0.898}   {\scriptsize($\pm$0.008)}   &   {\bf 0.821} {\scriptsize($\pm$0.011)} & {\bf 6.85} {\scriptsize($\pm$0.44)} & {\bf 1.78} {\scriptsize($\pm$0.23)} &  8   vs.  72  &   {\bf 0.884} {\scriptsize($\pm$0.007)} & {\bf 0.792} {\scriptsize($\pm$0.012)} & {\bf 5.89} {\scriptsize($\pm$0.31)} & {\bf 1.60} {\scriptsize($\pm$0.15)}  \\
  \bottomrule[1pt]
  \end{tabular}}
 \vspace{-1.0ex}
 \caption{Ablation studies of different parts in GBDL on the KiTS19 dataset and the Atrial Segmentation Challenge dataset. ``L vs. U'' denotes the number of labeled data versus that of unlabeled data. 
 For each of results, the ``mean {\scriptsize($\pm$std)}" is reported. }  
 \label{tab:abla_kits_LA_2}
 \end{table*}
\begin{table}
\centering 
\resizebox{0.75\textwidth}{!}{
\begin{tabular}{@{}ccccccc@{}}
 \toprule[1pt]
 \multirow{2}{*}{}    & \multicolumn{2}{c}{Scans used} &\multicolumn{4}{c}{Metrics} \\
 \cline{2-7}
 & Labeled & Unlabeled & Dice $\uparrow$ & Jaccard  $\uparrow$ & 95HD $\downarrow$ & ASD $\downarrow$ \\ 
 \hline  
 UA-MT \cite{yu2019uncertainty}     &  16  &  144 &     0.883 & 0.802 & 9.46 & 2.89  \\ 
 SASSNet \cite{li2020shape}  &  16   & 144 &  0.891 & 0.822 & 7.54 & 2.41 \\
 Double-UA  \cite{wang2020double}    &  16   &  144  &   0.895    & 0.828 & 7.42 & 2.16  \\
 Tripled-UA  \cite{wang2021tripled} &  16  &  144 &   0.887 &  0.815  & 7.55  &  2.12  \\
 CoraNet  \cite{shi2021inconsistency} & 16 &  144 &   0.898  & 0.820 & 7.23 &  1.89 \\  
 UA-MT* \cite{yu2019uncertainty}     &  16  &  144 &     0.878   & 0.797  & 8.93 & 2.76  \\
 Double-UA*  \cite{wang2020double}    &  16   &  144  &   0.899   & 0.823   & 7.55   & 2.21   \\
 Tripled-UA* \cite{wang2021tripled} &  16  &  144 &   0.886   &     0.819  & 7.43   &  2.26  \\
 CoraNet* \cite{shi2021inconsistency} & 16 &  144 &   0.894  &     0.820   & 7.37&  1.92 \\ 
 GBDL  &  16   &  144  &  {\bf 0.911}  &  {\bf 0.840} & {\bf 6.38} & {\bf 1.51} \\
 \hline  
 UA-MT \cite{yu2019uncertainty}    &  4   &  156  &    0.871      &   0.787  & 11.74 & 3.56 \\
 SASSNet \cite{li2020shape}  &  4   &  156 &  0.888 & 0.816 & 8.32 & 2.44 \\
 Double-UA  \cite{wang2020double}   &  4   &  156  &   0.887  &   0.817   & 8.04   &  2.34 \\
 Tripled-UA \cite{wang2021tripled} &  4   &  156  &   0.878 &  0.813  & 7.94  &  2.42 \\
 CoraNet \cite{shi2021inconsistency} & 4 &  156 &   0.882  &     0.814  & 8.21   &  2.44  \\  
 Double-UA*  \cite{wang2020double}   &  4   &  156  &   0.890  &     0.819   & 7.93    &  2.33  \\
 UA-MT* \cite{yu2019uncertainty}     &  4   &  156  &   0.874   &   0.790   & 11.33 & 3.21 \\
 Tripled-UA* \cite{wang2021tripled} &  4   &  156  &   0.882   &     0.810   & 7.81    &  2.47 \\
 CoraNet* \cite{shi2021inconsistency} & 4 &  156 &   0.886  &     0.812   & 8.43  &  2.39 \\
  GBDL     &  4   &  156  &  {\bf 0.898}   &   {\bf 0.821}  & {\bf 6.85}& {\bf 1.78} \\
  \bottomrule[1pt]
  \end{tabular}}
 \caption{Comparison with state-of-the-art semi-supervised segmentation methods on the KiTS19 dataset. ``*'' denotes previous methods based on the regular 3D-UNet with MC dropout.} 
 \label{tab:sota_kits} 
 \end{table}
Considering that the independent components assumption is widely
used in cVAE, we also evaluated LRL variant with this assumption, and we denote this variant as ``IC-LRL''.  
In this case,  $L_{KL[Q(Z|X)||P(Z)]}$ degenerates to $1/2 \cdot \sum_{D}[-log\frac{1}{\sum_{i=0}^{n-1}\frac{1}{\sigma_i^2}}
+ \frac{1}{\sum_{i=0}^{n-1}\frac{1}{\sigma_i^2}} + (\frac{1}{\sum_{i=0}^{n-1}\frac{1}{\sigma_i^2}}\sum_{i=0}^{n-1}\frac{\mu_i}{\sigma_i^2})^2  - 1 ]$, where $\mu_i$ and $\sigma_i$ are the mean vector and the variance vector of the Gaussian distribution for each slice in the input volume, respectively. The results shown in Table~\ref{tab:abla_kits_LA_1} indicate that the IC-LRL performs worse than our original LRL, verifying that the independent components assumption harms the performance of LRL.
Note that the VAE bottleneck might have some adverse impacts on reconstruction results and segmentation performance since it essentially compresses input data. The VAE bottleneck results in a low-dimensional representation, and some information is inevitably lost, even though there is another path in GBDL to compensate for it. How to address the loss in the VAE bottleneck can be further investigated in future studies. 

Third, we analyzed the two important parts of the proposed LRL, i.e., the reconstruction loss $L_{MSE}$ and the term $L_{KL[Q(Z|X)||P(Z))]}$. We removed these two loss terms,\footnote{Removing $L_{MSE}$ is equivalent to removing the reconstruction path, i.e., when we did an ablation study about removing the reconstruction loss $L_{MSE}$, we also removed the upper 3D-UNet decoder of LRL in Figure~\ref{eqn:overall}.} respectively, to study their impact on GBDL. As Table~\ref{tab:abla_kits_LA_2} shows, when $L_{MSE}$ or $L_{KL[Q(Z|X)||P(Z))]}$ is removed, a performance decline can be observed, demonstrating the importance of these two terms and the correctness of our Bayesian formulation.

\begin{table}[t]
\centering 
\resizebox{0.8\textwidth}{!}{
\begin{tabular}{@{}ccccccc@{}}
 \toprule[1pt]
 \multirow{2}{*}{}    & \multicolumn{2}{c}{Scans used} &\multicolumn{4}{c}{Metrics} \\
 \cline{2-7}
 & Labeled & Unlabeled & Dice $\uparrow$ & Jaccard  $\uparrow$ & 95HD $\downarrow$ & ASD $\downarrow$ \\
 \hline  
 UA-MT \cite{yu2019uncertainty}     &  16   &  64 &    0.889      &   0.802 & 7.32 &  2.26  \\
 SASSNet \cite{li2020shape}  &  16   &  64 &  0.895 & 0.812 & 8.24 & 2.20 \\
 Double-UA  \cite{wang2020double}    &  16   &  64  &   0.897    &   0.814 &  7.04 & 2.03 \\ 
 Tripled-UA \cite{wang2021tripled} & 16 &  64 &   0.893 &     0.810  & 7.42  &  2.21  \\
 CoraNet \cite{shi2021inconsistency} & 16 &  64 &   0.887 &     0.811   & 7.55   &  2.45 \\
 Reciprocal Learning \cite{zeng2021reciprocal} & 16 & 64 & {\bf 0.901} & 0.820 & 6.70 & 2.13 \\
 DTC \cite{luo2021semi} & 16 &  64 &   0.894 &     0.810   & 7.32   &  2.10 \\ 
 3D Graph-S$^2$ Net \cite{huang20213d} & 16  & 64  & 0.898 & 0.817  & 6.68 & 2.12 \\
 LG-ER-MT \cite{hang2020local} & 16 & 64 & 0.896 & 0.813 & 7.16 & 2.06 \\
 Double-UA*  \cite{wang2020double}    &  16   &  64  &  0.894   &   0.809  &  6.16 & 2.28 \\
 UA-MT* \cite{yu2019uncertainty}     &  16   &  64 &  0.891   &   0.793   & 6.44  &  2.39 \\
 Tripled-UA* \cite{wang2021tripled} & 16 &  64 &   0.889 &     0.809  & 6.88  &  2.48  \\
 CoraNet* \cite{shi2021inconsistency} & 16 &  64 &   0.883 &     0.805  & 6.73   &  2.67 \\
 GBDL &  16   &  64  &    0.894  & {\bf 0.822} & {\bf 4.03} & {\bf 1.48}   \\
 \hline 
 UA-MT \cite{yu2019uncertainty}    &  8   &  72  &    0.843      &   0.735 &  13.83 & 3.36    \\
 SASSNet \cite{li2020shape}  &  8   &  72 &  0.873 & 0.777 & 9.62 & 2.55 \\ 
 Double-UA  \cite{wang2020double}   &  8   &  72  &   0.859      &   0.758 & 12.67 &  3.31  \\
 Tripled-UA \cite{wang2021tripled} &  8  &  72 &   0.868  &     0.768   & 10.42   &  2.98  \\
 CoraNet \cite{shi2021inconsistency} & 8 &  72 &   0.866  &     0.781   & 12.11  &  2.40   \\
 Reciprocal Learning \cite{zeng2021reciprocal} & 8 & 72 & 0.862 & 0.760 & 11.23 & 2.66 \\
 DTC \cite{luo2021semi} & 8 &  72 &   0.875 &     0.782   & 8.23   &  2.36 \\ 
 LG-ER-MT \cite{hang2020local} & 8 & 72 & 0.855 & 0.751 & 13.29 & 3.77 \\
 3D Graph-S$^2$ Net \cite{huang20213d} & 8  & 72  & 0.879 & 0.789  & 8.99 & 2.32\\
 Double-UA*  \cite{wang2020double}   &  8   &  72  &   0.864  &   0.767  & 10.99 &  3.02  \\
 UA-MT* \cite{yu2019uncertainty}      &  8   &  72 &    0.847    &   0.744   & 12.32  &  3.20 \\
 Tripled-UA* \cite{wang2021tripled} &  8  &  72 &   0.868 &     0.760  & 9.73  &  3.31 \\
 CoraNet* \cite{shi2021inconsistency} & 8 &  72 &   0.861 &     0.770 & 11.32    &  2.46  \\
 GBDL  &  8   &  72  &   {\bf 0.884} & {\bf 0.792}& {\bf 5.89}  & {\bf 1.60}  \\
  \bottomrule[1pt]
  \end{tabular}}
 \caption{Comparison with state-of-the-art semi-supervised segmentation methods on the Atrial Segmentation Challenge. ``*'' denotes previous methods based on the regular 3D-UNet with MC dropout.} 
 \label{tab:sota_LA}
 \end{table}

 \begin{table} 
\centering
\resizebox{0.8\textwidth}{!}{
\begin{tabular}{@{}ccccccc@{}}
 \toprule[1pt]
 \multirow{2}{*}{}    & \multicolumn{2}{c}{Scans used} &\multicolumn{4}{c}{Metrics} \\
 \cline{2-7}
 & Labeled & Unlabeled & Dice $\uparrow$ & Jaccard  $\uparrow$  & 95HD $\downarrow$ & ASD $\downarrow$  \\
 \hline 
 3D-UNet w/ MC dropout &   100   &  0    & 0.950   & 0.899   & 6.04   & 1.53  \\ 
 \hline 
 UA-MT \cite{yu2019uncertainty}     &   5   &  95   &    0.920   & 0.867   & 13.21    & 4.54    \\ 
 Double-UA  \cite{wang2020double}    &   5   &  95   &   0.927   &   0.878   & 12.11   & 4.19   \\
 Tripled-UA  \cite{wang2021tripled}   &   5   &  95   &   0.921  &   0.869    & 11.77   & 3.62   \\
 CoraNet \cite{shi2021inconsistency} & 5 &  95 &  0.923  &  0.877   & 10.84  & 4.28 \\ 
 GBDL    &   5   &  95   &   {\bf 0.935}  &  {\bf 0.884} & {\bf 7.89}  & {\bf 2.42}  \\
 \bottomrule[1pt]
\end{tabular}}
\vspace{-1ex}
 \caption{Comparison with state-of-the-art semi-supervised segmentation methods on the Liver Segmentation dataset. Previous methods in this table are also based on the regular 3D-UNet with MC dropout.}  
\label{tab:sota_liver}

\end{table}

Besides, as the two 3D-UNet encoders in  LRL are of the same architecture (see the Appendix A), and both aim at extracting features from input volumes, we evaluated a case where the two encoders share their weights. Nevertheless, sharing the weights degrades the performance, as shown in Table~\ref{tab:abla_kits_LA_2}. Similarly, although some layers of the two 3D-UNet decoders in the LRL also have the same configuration, sharing their weights  would  clearly lead to a lower performance, as the respective objectives of them are different, leading to different gradient descent directions.

Finally, when compared with the baseline model, the proposed GBDL can improve the performance with a huge margin with respect to the four assessment metrics for both datasets. So, the proposed method is a good solution for  semi-supervised volumetric medical imaging segmentation.

 \subsection{Comparison with State-of-the-Art Methods}
The proposed GBDL is compared with previous state-of-the-art methods, and since previous works provide no standard deviation, we also only report the mean values of the four evaluation metrics in Tables~\ref{tab:sota_kits}, \ref{tab:sota_LA}, and \ref{tab:sota_liver}. 
Since most previous methods are based on VNet, for fair comparison, our regular 3D-UNet with MC dropout is designed to have similar numbers of parameters. Furthermore, we also re-implement previous Bayesian deep learning methods based on the regular 3D-UNet with MC dropout, and according to the results in Tables~\ref{tab:sota_kits} and \ref{tab:sota_LA},  their performance are similar to their VNet counterparts, meaning that the comparison between our method and previous methods is relatively fair.

\begin{table}
\centering 
\resizebox{0.65\textwidth}{!}{
\begin{tabular}{@{}c|c|c|c|c@{}}
 \toprule[1pt]
 \multirow{2}{*}{Method} & \multicolumn{2}{c|}{KiTS19}  & \multicolumn{2}{c }{Atrial} \\
 \cline{2-5}
       & 16 labeleld  & 4 labeleld & 16 labeleld  & 8 labeleld \\
 \hline  
 UA-MT \cite{yu2019uncertainty}  &  0.868  & 0.841 &  0.851  & 0.836 \\
 Double-UA \cite{wang2020double}  &  0.881  &  0.864 &  0.861  & 0.842 \\
 Triplet-UA  \cite{wang2021tripled} &  0.874  &  0.852 &  0.864  & 0.847  \\
 CoraNet  \cite{shi2021inconsistency} &  0.882  &  0.866 &  0.867  & 0.855 \\
 GBDL  &  {\bf 0.897}  &  {\bf 0.877} &  {\bf 0.882}  & {\bf 0.864} \\ 
  \bottomrule[1pt]
  \end{tabular}} 
 \caption{  PAvPU of different Bayesian deep learning methods.} 
 \label{tab:pavpu} 
 \end{table}

\begin{figure*}[t] 
\centering
\includegraphics[width=\linewidth]{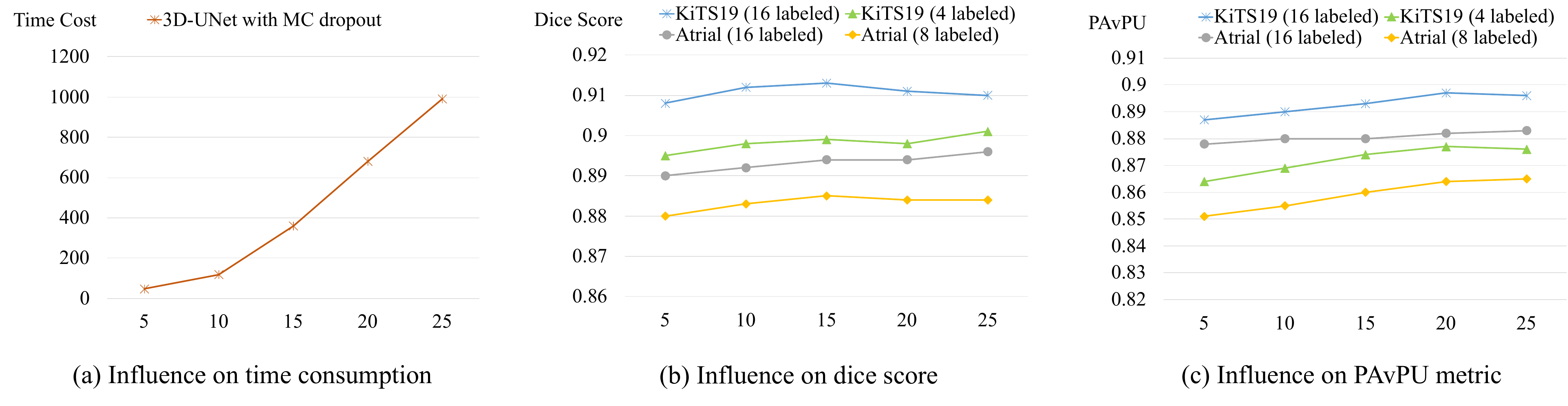} 
 
\caption{The influence of the number of feedforward passes on: (a) time costs (ms) for processing a $128\times128\times32$ volume input on a single GeForce GTX 1080 Ti; (b) Dice score; and (c) PAvPU metric.  The horizontal axis refers to the number of feedforward passes, namely, $T$.}
\vspace{-1ex}
\label{fig:runtime}
\end{figure*}

By  Tables~\ref{tab:sota_kits},  \ref{tab:sota_LA}, and \ref{tab:sota_liver}, the performance of GBDL is higher than previous state-of-the-art results on the three public medical datasets. When the number of labeled data decreases, the performance gap between our GBDL and previous state-of-the-art methods becomes larger, which indicates that our method is more robust when fewer labeled data were used for training. More importantly, GBDL performs better than all previous Bayesian deep-learning-based methods \cite{yu2019uncertainty, wang2020double, shi2021inconsistency, wang2021tripled}, which verifies that GBDL is better than the teacher-student architecture used in them and that the generative learning paradigm is more suitable for pseudo-label generation and solving the semi-supervised segmentation task than the discriminative counterpart.

Furthermore, we compare the patch accuracy vs. patch uncertainty (PAvPU) metric \cite{mukhoti2018evaluating} for GBDL
and previous Bayesian-based methods to evaluate their uncertainty estimation. By   Table~\ref{tab:pavpu}, GBDL can output more reliable and well-calibrated uncertainty estimates than other methods, so that clinicians can do a post-processing and refine the results in practice based on the uncertainty given by GBDL.

Finally, since full Bayesian inference is time-consuming, we show the relationship among the number of feedforward passes $T$, the time consumption, and the performance in Figure~\ref{fig:runtime}. 
By Figure~\ref{fig:runtime}(a), 
the time costs rise with increasing~$T$, but by Figure~\ref{fig:runtime}(b) and (c), the Dice score and the PAvPU metric are not greatly impacted by $T$. Thus, $T=5$ can be chosen for saving time to get the uncertainty in practice, with only a minor performance degradation.

\section{Visualization Results}

\begin{figure}
 \centering
 \includegraphics[width=\linewidth]{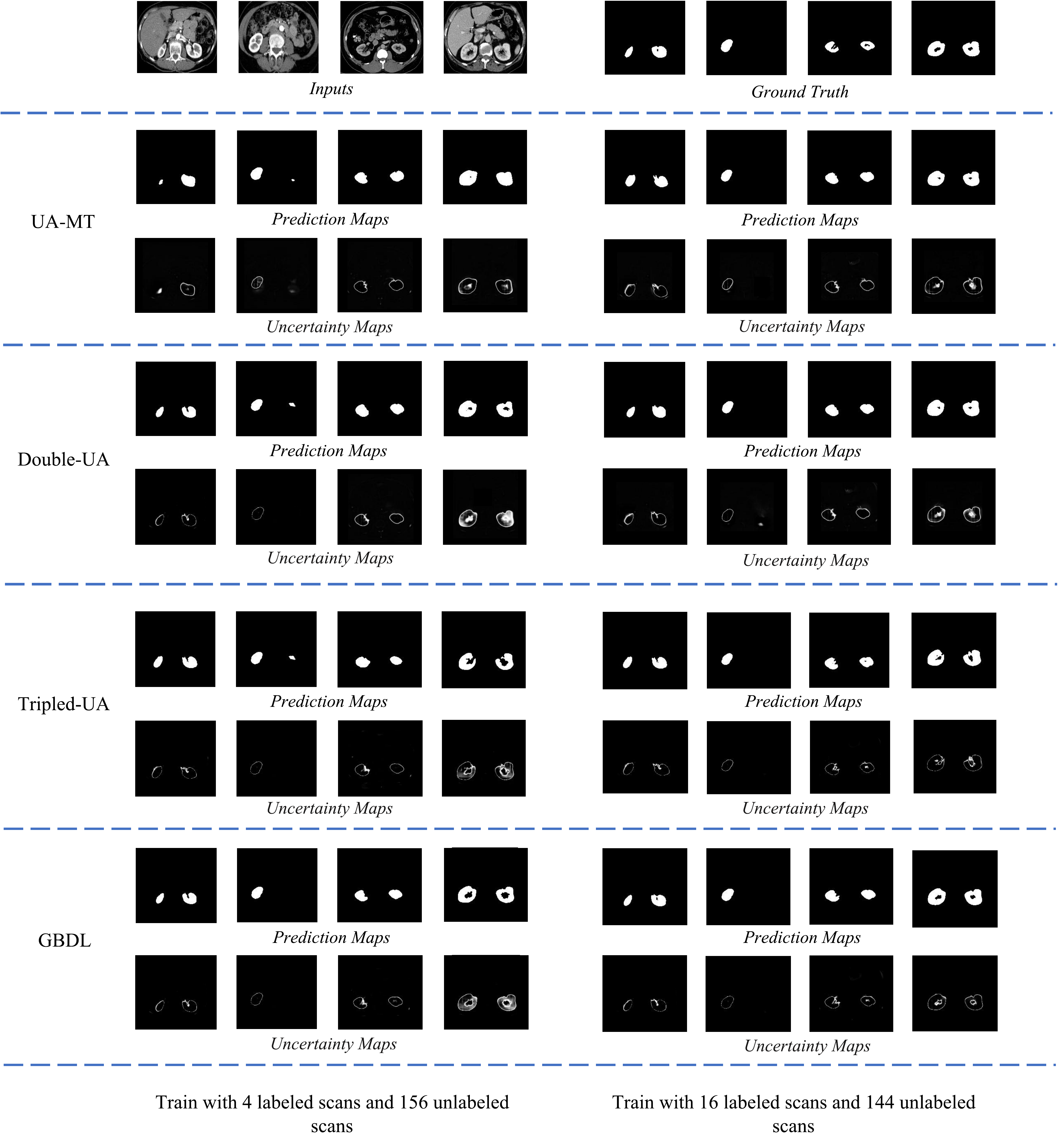}
 \vspace*{-3ex}
 \caption{The visualization of some predicted slices from the KiTS19 dataset for different Bayesian deep learning based methods  \cite{yu2019uncertainty, wang2020double, wang2021tripled}.}
 \label{fig:vis_kits} 
\end{figure}

\begin{figure}
 \centering
 \includegraphics[width=\linewidth]{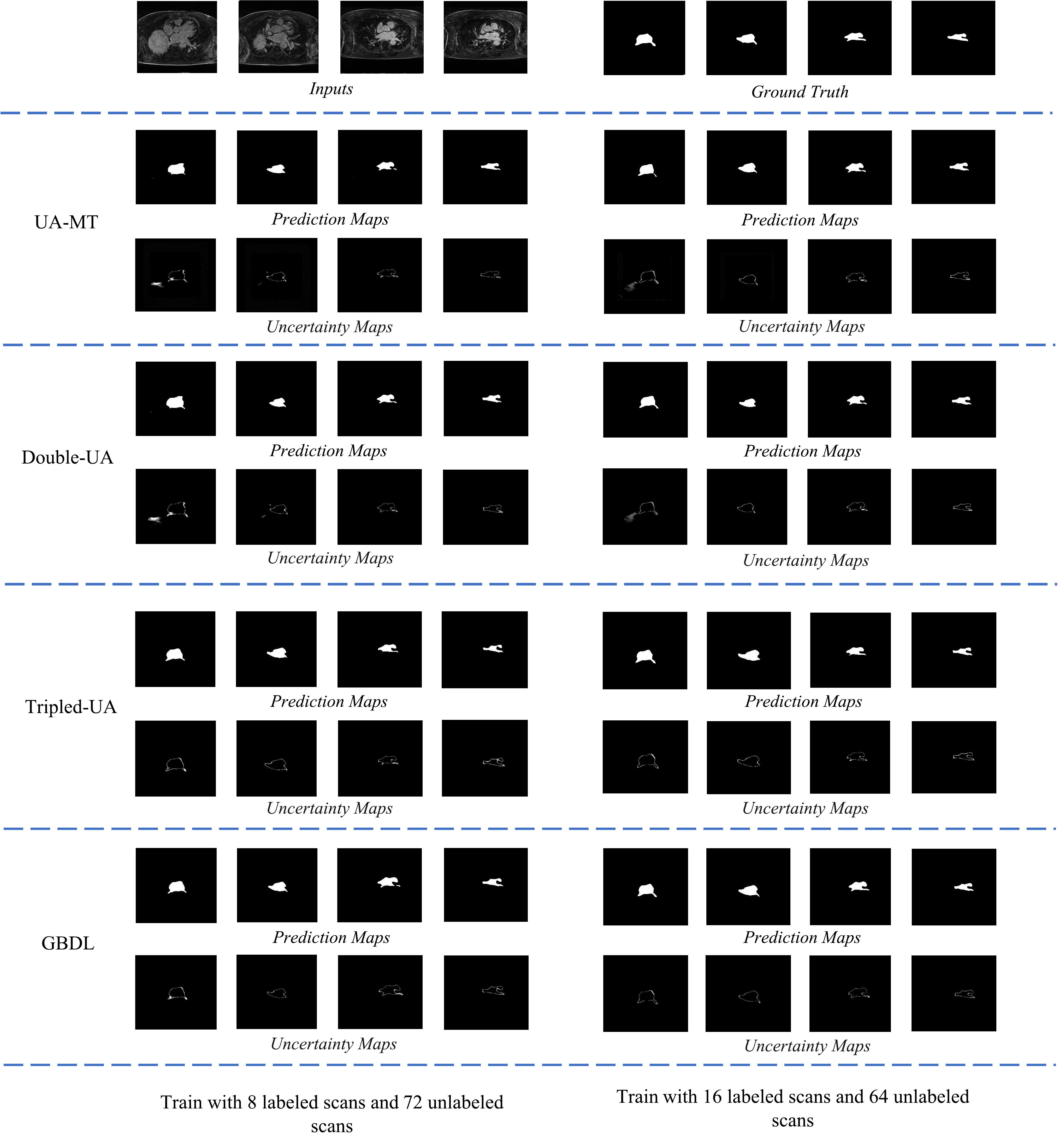}
 \vspace*{-3ex}
 \caption{The visualization of some predicted slices from the Atrial Segmentation Challenge dataset for different Bayesian deep learning based methods \cite{yu2019uncertainty, wang2020double, wang2021tripled}.}
 \label{fig:vis_atrial}  
\end{figure}

We visualize some predicted results from the KiTS19 dataset and the Atrial Segmentation Challenge dataset in Figure~\ref{fig:vis_kits} and Figure~\ref{fig:vis_atrial} for different Bayesian deep learning based methods, including UA-MT \cite{yu2019uncertainty}, Double-UA \cite{wang2020double}, Tripled-UA \cite{wang2021tripled}, and GBDL. Each result contains a prediction map and a corresponding uncertainty map. The uncertainty maps for the other three methods come from their teacher models \cite{yu2019uncertainty, wang2020double, wang2021tripled}, and 
the brightness of each pixel in an uncertainty map is negatively associated with the model's confidence for predicting that pixel. 
According to the visualization results, when only a limited number of training data are annotated, GBDL can still find and segment the foreground. Although there are some misclassified regions, uncertainty maps are able to give high uncertainties to them, so that clinicians can further improve the results in a real-world scenario.

\section{Summary} 
\label{sec:3.5}

In this chapter, we rethink the issues in previous Bayesian deep-learning-based methods for  semi-supervised volumetric medical image segmentation, and have designed a new generative Bayesian deep learning (GBDL) architecture to solve them. The proposed method outperforms previous state-of-the-art methods on three public medical benchmarks, showing its effectiveness for handling the data with only a very limited number of annotations. 
It is worth noting that the proposed method can be applied to traditional computer vision data formats with 3D dimensions, such as videos, in principle, even though it is evaluated on medical imaging datasets in this chapter.

\section{Limitations}
\label{sec:3.6}

Based on the discussions in Section~\ref{sec:2.2} and some experimental results in this chapter (e.g., Figure~\ref{fig:runtime}), certain limitations can be observed. A critical issue is the time consumption for uncertainty estimation. With MC dropout, several feed-forward passes are required to quantify uncertainty, which restricts its practical use in dealing with large-scale datasets. Moreover, choosing the hyperparameters for MC dropout and integrating them into deep neural networks can be quite tricky. Hence, finding an alternative to break the monopoly of MC dropout in SSL is necessary.

\chapter{NP-Match: When Neural Processes meet Semi-Supervised Learning}

In this chapter, I develop a new probabilistic model for large-scale semi-supervised image classification, aiming to solve the shortcomings of MC dropout. I first adapt NPs to the image classification task and integrate them into a recent SSL framework. Then, to tackle the challenges caused by noisy labels, I also propose a new divergence term to aid in optimization. This chapter is organized as follows. In Section~\ref{4.1}, an introduction is given to show the motivation for exploring new probabilistic methods. In Section~\ref{4.2}, related methods are presented. Section~\ref{4.3} presents NP-Match and the uncertainty-guided skew-geometric JS divergence ($JS^{G_{\alpha_u}}$), followed by the experimental settings and results in Section~\ref{4.4}. Finally, a summary and limitations are discussed in Sections~\ref{4.5} and~\ref{4.6}, respectively.

\section{Introduction}
\label{4.1}

Most approaches to SSL for image classification 
can be further classified into two categories, namely,  deterministic \cite{sohn2020fixmatch, li2021comatch, zhang2021flexmatch, nassar2021all, pham2021meta, hu2021simple} and probabilistic ones \cite{rizve2021defense}. 
A deterministic approach aims at directly making predictions, while a probabilistic approach tries to additionally model the predictive distribution. 
Current SOTA methods are deterministic, including FixMatch \cite{sohn2020fixmatch}, CoMatch \cite{li2021comatch}, and FlexMatch \cite{zhang2021flexmatch}, which have achieved promising results on public benchmarks. 
In contrast, progress on probabilistic approaches 
lags behind, which is mainly shown by the fact that there are only few studies on this task and MC dropout becomes the main option for implementing the probabilistic model \cite{rizve2021defense}. 
Meanwhile, MC dropout also dominates uncertainty-based approaches in other SSL tasks \cite{sedai2019uncertainty, shi2021inconsistency, wang2021tripled, yu2019uncertainty, zhu2020grasping}. However, it has some drawbacks when used in practice, which have been discussed above, encouraging us to explore more possible alternatives in the realm of SSL.


Considering that MC dropout is an approximation to the Gaussian process (GP) model \cite{gal2016dropout}, we turn to another approximation model called
neural processes (NPs) \cite{garnelo2018neural}, which can be regarded as an NN-based formulation that approximates GPs.
Similarly to a GP, a neural process  is also a probabilistic model that defines distributions over functions.
Thus, an NP is able to rapidly adapt to new observations, with the advantage of estimating the uncertainty of each observation.
There are two main aspects that motivate us to investigate NPs in SSL. Firstly, GPs have been preliminarily explored for some SSL tasks \cite{sindhwani2007semi, jean2018semi, yasarla2020syn2real}, because of the property that their kernels are able to compare labeled data with unlabeled data when making predictions. 
NPs share this property, since it has been proved that NPs can learn non-trivial implicit kernels from data \cite{garnelo2018neural}. As a result, NPs are able to make predictions for target points conditioned on context points. This feature is highly relevant to SSL, which must learn from limited labeled samples in order to make predictions for unlabeled data, similarly to how NPs are able to impute unknown pixel values (i.e., target points) when given only a small number of known pixels (namely, context points) \cite{garnelo2018neural}. 
Due to the learned implicit kernels in NPs \cite{garnelo2018neural} and the successful application of GPs to different SSL tasks \cite{sindhwani2007semi, jean2018semi, yasarla2020syn2real}, NPs could be a suitable probabilistic model for SSL, as the kernels can compare labeled data with unlabeled data in order to improve the quality of pseudo-labels for the unlabeled data at the training stage. Secondly, previous GP-based works for SSL do not explore the semi-supervised large-scale image classification task, since GPs are computationally expensive, which usually incur a $\mathcal{O}(n^3)$ runtime for $n$ training points. But, unlike GPs, NPs are more efficient than GPs, providing the possibility of applying NPs to this task. NPs are also computationally significantly more efficient than current MC-dropout-based approaches to SSL, since, given an input image, they only need to perform one feedforward pass to obtain the prediction with an uncertainty estimate.

We take the first step to explore NPs in large-scale semi-supervised image classification, and propose a new probabilistic method called NP-Match. NP-Match still rests on the combination of consistency regularization and pseudo-labeling, but it incorporates NPs to the top of deep neural networks, and therefore it is a probabilistic approach. Compared to the previous probabilistic method for semi-supervised image classification, i.e., the MC-dropout-based method \cite{rizve2021defense}, NP-Match not only can make predictions and estimate uncertainty more efficiently,  inheriting the advantages of NPs, but also can achieve a better performance on public benchmarks. 

Summarizing, the main contributions are:\vspace*{-1ex}
\begin{itemize}[leftmargin=*, itemsep=0.5pt]
\item We propose NP-Match, which adjusts NPs to SSL, and explore its use in semi-supervised large-scale image classification. To our knowledge, this is the first such work. In addition, NP-Match has the potential to break the monopoly of MC dropout as the probabilistic model in SSL. 

\item We experimentally show that the Kullback-Leibler (KL) divergence in the evidence lower bound (ELBO) of NPs \cite{garnelo2018neural} is not a good choice in the context of SSL, which may negatively impact the learning of global latent variables. To tackle this problem, we propose a new uncertainty-guided skew-geometric Jensen-Shannon (JS) divergence ($JS^{G_{\alpha_u}}$) for NP-Match. 

\item We show that NP-Match outperforms SOTA results or achieves competitive results on public benchmarks, demonstrating its effectiveness for SSL. We also  show that NP-Match estimates uncertainty faster than the MC-dropout-based probabilistic model, which can improve the training and the test efficiency.  
\end{itemize}

\section{Related Work}
\label{4.2}

We now briefly review related works, including {\it semi-super\-vised learning (SSL) for image classification},
{\it Gaussian processes (GPs) for SSL},  
{\it imbalanced semi-supervised image classification}, and {\it automatic chest X-ray analysis}.

{\bf SSL for image classification.}  
Most methods for semi-supervised image classification in the past few years are based on pseudo-labeling and consistency regularization. Pseudo-labeling approaches
rely on the high confidence of pseudo-labels, which can be added to the training data set as labeled data, and those approaches can be classified into two classes, namely, disagreement-based models and self-training models. The former models aim to train multiple learners and exploit the disagreement during the learning process \cite{qiao2018deep, dong2018tri}, while the latter models aim at training the model on a small amount of labeled data, and then using its predictions on the unlabeled data as pseudo-labels \cite{lee2013pseudo, zhai2019s4l, wang2020enaet, pham2021meta}. 
Consistency-regularization-based approaches work by performing different transformations on an input image and adding a regularization term to make their predictions consistent \cite{bachman2014learning, sajjadi2016regularization, laine2016temporal, berthelot2019mixmatch, xie2019unsupervised}. 
Based on these two approaches, FixMatch \cite{sohn2020fixmatch} is proposed, which achieves new state-of-the-art (SOTA) results on the most commonly-studied SSL benchmarks.
FixMatch \cite{sohn2020fixmatch} combines the merits of these two approaches: given an unlabeled image, weak data augmentation and strong data augmentation are performed on the image, leading to two versions of the image, and then FixMatch produces a pseudo-label based on its weakly-augmented version and a preset confidence threshold, which is used as the true label for its strongly augmented version to train the whole framework.  
The success of FixMatch inspired several subsequent methods \cite{li2021comatch, rizve2021defense, zhang2021flexmatch, nassar2021all, pham2021meta, hu2021simple}. For instance, Li {\it et al.} \cite{li2021comatch} additionally design the classification head and the projection head for generating a class probability and a low-dimensional embedding, respectively. The projection head and the classification head are jointly optimized during training. 
Specifically, the former is learnt with contrastive learning on pseudo-label graphs to encourage the embeddings of samples with similar pseudo-labels to be close, and the latter is trained with pseudo-labels that are smoothed by aggregating information from nearby samples in the embedding space. Zhang {\it et al.} \cite{zhang2021flexmatch} propose to use dynamic confidence thresholds that are automatically adjusted according to the model’s learning status of 
each~class. Rizve {\it et al.} \cite{rizve2021defense} propose an uncertainty-aware
pseudo-label selection (UPS) framework for semi-supervised image classification. The UPS framework introduces MC dropout to obtain uncertainty estimates, which are then leveraged as a tool for selecting pseudo-labels. This is the first work using MC dropout for semi-supervised image classification. 

{\bf GPs for SSL.}  Since NPs are also closely related to GPs, we review the application of GPs to different SSL tasks in this part.
GPs, which are non-parametric models, have been preliminarily investigated in different semi-supervised learning tasks. For example, Sindhwani {\it et al.} \cite{sindhwani2007semi} introduce a semi-supervised GP classifier, which incorporates the information of relationships among labeled and unlabeled data  into the kernel. Their approach, however, has high  computational costs and is thus only evaluated for a simple binary classification task on small datasets.  Deep kernel learning  \cite{wilson2016deep} also lies on the spectrum between NNs and GPs, and has been integrated into a new framework for the semi-supervised regression task, named semi-supervised deep kernel learning  \cite{jean2018semi}, which  aims to minimize the predictive variance for unlabeled data,  encouraging  unlabeled embeddings to be near labeled embeddings. Semi-supervised deep kernel learning, however, has not been applied to SSL image classification, and (similarly to semi-supervised GPs) also comes with a high (cubic) computational complexity. 
Recently, Yasarala {\it et al.} \cite{yasarla2020syn2real} proposed to combine GPs with UNet \cite{ronneberger2015u} for SSL image deraining. Here, GPs are used to get pseudo-labels for unlabeled samples based on the feature representations of labeled and unlabeled images. GPs have also been combined with graph convolutional networks for  semi-supervised learning on graphs \cite{ng2018bayesian, walker2019graph, liu2020uncertainty}. 
Although many previous works explore GPs in different semi-supervised learning tasks, none of them investigates the application of GPs to  semi-supervised large-scale image classification.

{\bf Imbalanced semi-supervised image classification.} 
Since we conducted experiments under the imbalanced semi-supervised image classification setting, and in order to facilitate readers' access to literature related to imbalanced learning, we present some related works in this section. Training deep neural networks requires large-scale datasets, and one fundamental characteristic of these datasets in practice is that the data distribution over different categories is imbalanced (e.g., long-tailed), as it is common that the images of some categories are difficult to be collected. Training deep models on imbalanced datasets makes the models overfit to majority classes, and several methods have been proposed to ameliorate this issue \cite{wang2021rsg, cui2021parametric, liu2019large, li2021self, cao2019learning, hong2021disentangling, feng2021exploring, cai2021ace, samuel2021distributional, desai2021learning, hou2022batchformer, li2022nested, zhong2021improving, menon2021long}. 
According to Yang and Xu~\cite{yang2020rethinking}, 
semi-supervised learning benefits  imbalanced learning, as using extra data during training can reduce label bias, which greatly improves the final classifier. Therefore, imbalanced semi-supervised image classification has been drawing extensive attention in recent years.  
Specifically, Hyun et al.~\cite{hyun2020class} analyze how class imbalance affects SSL by looking into the topography of the decision boundary, and then they propose a suppressed consistency loss (SCL) to suppress the influence of consistency loss on infrequent classes. 
Kim et al.~\cite{kim2020distribution} design an iterative procedure, called distribution aligning
refinery of pseudo-labels (DARP), to solve a convex optimization problem that aims at refining biased pseudo-labels so that their distribution can match the true class distribution of unlabeled data.  Wei
et al.~\cite{wei2021crest} introduce a class-rebalancing selftraining scheme (CReST) to re-sample pseudo-labeled data and to add them into the labeled set, for refining the whole model. Inspired by the concept of decoupled learning \cite{kang2020decoupling}, namely, training a feature extractor and a classifier separately,  He et al.~\cite{he2021rethinking} propose a bi-sampling method, which integrates two data samplers with various sampling strategies for decoupled learning, therefore benefiting both the feature extractor and the classifier.  Lee et al.~\cite{lee2021abc} design an auxiliary balanced classifier (ABC) that is trained  by
using a mask that re-balances the class distribution within every mini-batch. Besides, Oh et al.~\cite{oh2022daso} solve the imbalanced semi-supervised image classification problem by proposing a new framework called distribution-aware semantics-oriented (DASO) pseudo-labels. DASO focuses on debiasing pseudo-labels through blending two complementary types of pseudo-labels, which enables the framework to deal with the imbalanced semi-supervised image classification task, even when the class distribution of unlabeled data is inconsistent with that of labeled data. Here, we evaluate our method for imbalanced semi-supervised image classification based on the DASO framework.

{\bf Automatic chest X-ray analysis.}  
Considering that we evaluated NP-Match under the multi-label semi-supervised image classification setting with a chest X-ray dataset, we summarize some related works with respect to automatic chest X-ray analysis. 
Wang {\it et al.} \cite{wang2017chestx} and Rajpurkar {\it et al.} \cite{rajpurkar2017chexnet} proposed a traditional convolutional neural network for localizing disease using a class activation map. Taghanaki {\it et al.} \cite{taghanaki2019infomask} applied a variational online mask on a minuscule region within the image to predict disease based on the unmasked region. Guan {\it et al.} \cite{guan2020multi} introduced a class-specific attention method, while Ma {\it et al.} \cite{ma2019multi} implemented cross-attention with two traditional convolutional neural networks. Hermoza {\it et al.} \cite{hermoza2020region} utilized a feature pyramid network and an extra detection module to diagnose disease. Li {\it et al.} \cite{li2018thoracic} proposed a framework that simultaneously identifies and localizes disease, leveraging a limited amount of additional supervision. Liu {\it et al.} \cite{liu2019align}, also using extra supervision, suggested a method to align chest X-ray images and learn discriminative features by contrasting positive and negative samples. Due to the prevalent of vision transformers (ViTs) nowadays, Xiao {\it et al.} \cite{xiao2023delving} have pre-trained ViTs on a large number of chest X-rays using Masked Autoencoders (MAE), which achieve competitive results for multi-label thorax disease classification. While numerous researchers have explored disease identification via chest radiography, they do not thoroughly study the case when labels are limited, which is more common in practice.  Recently, Liu {\it et al.} \cite{liu2022acpl} propose a anti-curriculum pseudo-labelling (ACPL) framework to address multi-class and multi-label SSL problems on chest X-rays. ACPL boosts its performance with unlabeled data,  by selecting highly informative samples and using an ensemble of classifiers to produce pseudo labels for them.

\section{NP-Match}
\label{4.3}
\begin{figure*}[t]
\centering
\includegraphics[width=\linewidth]{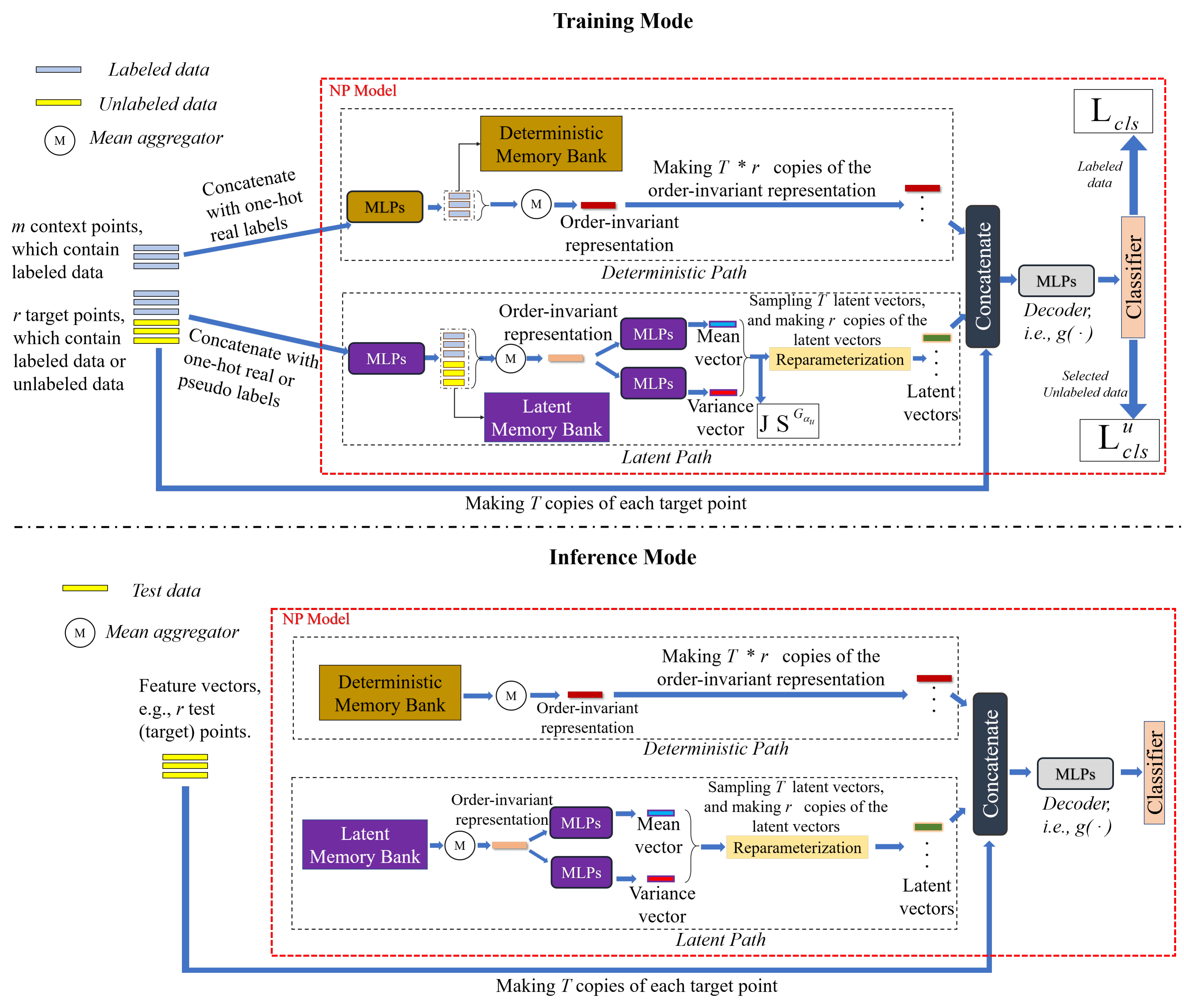}
\vspace{-4ex}
\caption{
An overview of NP-Match. The NP model is shown in the red dotted box, and its input feature vectors come from the global average pooling layer in the convolutional neural network (CNN). For simplicity, the CNN is omitted in the figure.}
\label{fig:np} 
\end{figure*}

In this section, we provide a detailed description of NP-Match. As Figure~\ref{fig:np} shows, NP-Match is mainly composed of two parts: a deep neural network and an NP model. The deep neural network is leveraged for obtaining feature representations of input images, while the NP model is built upon the network to receive the representations for classification. 

\subsection{NP Model for Semi-supervised Image Classification} 
Since we extend the original NPs \cite{garnelo2018neural} to the classification task, $p(y_{1:n}|g(x_{1:n}, z), x_{1:n})$ in Eq.~(\ref{eq:joint2})  should define a categorical distribution rather than a Gaussian distribution. Therefore, we parameterize the categorical distribution by probability vectors from a classifier that contains a weight matrix ($\mathcal{W}$) and a softmax function ($\Phi$):
\begin{equation}
\label{eq:likelihood1}
p(y_{1:n}|g(x_{1:n}, z), x_{1:n}) =  Categorical(\Phi(\mathcal{W}g(x_{1:n}, z))).
\end{equation}
Note that $g(\cdot)$ can be learned via amortised variational inference, and to use this method, two steps need to be done: (1)~parameterize a variational distribution over $z$, and  (2)~find the evidence lower bound (ELBO) as the learning objective. For the first step, we let $q(z|x_{1:n}, y_{1:n})$ be a variational distribution defined on the same measure space, which can be parameterized by a neural network. For the second step, 
given a finite sequence with length $n$, we assume that there are $m$ context points ($x_{1:m}$) and $r$ target points ($x_{m+1:\ m+r}$) in it, i.e., $m+r=n$. Then, the ELBO is given by  (with proof in the Appendix B):
\begin{equation} 
\begin{aligned}
\label{eq:elbo}
&log\ p(y_{1:n}|x_{1:n}) \ge \\ & \mathbb{E}_{q(z|x_{m+1:\ m+r}, y_{m+1:\ m+r})}\Big[\sum^{m+r}_{i=m+1}log\ p(y_i|z, x_i) -  log\ \frac{q(z|x_{m+1:\ m+r}, y_{m+1:\ m+r})}{q(z|x_{1:m}, y_{1:m})}\Big] + const.
\end{aligned}
\end{equation}
To learn the NP model, one can maximize this ELBO. Under the setting of SSL, we consider that only labeled data can be treated as context points, and either labeled or unlabeled data can be treated as target points, since the target points are what the NP model makes predictions for.

\subsection{NP-Match Pipeline} 
 
We now introduce the NP-Match pipeline.
We first focus on the configuration of the NP model, which is shown in the red dotted box in Figure~\ref{fig:np}. 
The NP model is mainly constructed by MLPs, memory banks, and a classifier. Specifically, the classifier is composed of the weight matrix ($\mathcal{W}$) and the softmax function ($\Phi$). Similarly to the original implementation of NPs, we build two paths with the memory banks and MLPs, namely, the latent path and the deterministic path. The decoder $g(\cdot)$ is also implemented with MLPs. 
The workflow of NP-Match at the training stage and the inference stage are different, which are shown in Figure~\ref{fig:np}, and they are introduced separately as follows.

{\bf Training mode.} 
Given a batch of $B$ labeled images  $\mathcal{L} = \{(x_i, y_i): i\,{\in}\, \{1,\ldots,B\}\}$ and a batch of unlabeled images $\mathcal{U} = \{x^u_i: i\,{\in}\, \{1,\ldots, \mu B\}\}$ at each iteration, where $\mu$ determines the relative size of $\mathcal{U}$ to $\mathcal{L}$, we apply weak augmentation (i.e., 
crop-and-flip) on the labeled and unlabeled samples, and strong augmentation (i.e.,  RandAugment \cite{cubuk2020randaugment}) on only the unlabeled samples. 
After the augmentation is applied, the images are passed through the deep neural network, 
and the features are input to the NP model, which finally outputs the predictions and associated uncertainties. The detailed process can be summarized as follows. 
At the start of each iteration, NP-Match is switched to inference mode, and it makes predictions for the weakly-augmented unlabeled data. 
Then, inference mode is turned off, and those predictions are treated as pseudo-labels for unlabeled data.  
After receiving the features, real labels, and pseudo-labels, the NP model first duplicates the labeled samples and treats them as context points, and all the labeled and unlabeled samples in the original batches are then treated as target points, since the NP model needs to make a prediction for them. Thereafter, the target points and context points are separately fed to the latent path and the deterministic path. As for the latent path, target points are concatenated with their corresponding real labels or pseudo labels, and processed by MLPs to get new representations. 
Then, the representations are averaged by a mean aggregator along the batch dimension, leading to an order-invariant representation, which implements the exchangeability  and the consistency condition, and they are simultaneously stored in the latent memory bank, which is updated with a first-in-first-out strategy. 
After the mean aggregator, the order-invariant representation is further processed by other two MLPs in order to get the mean vector and the variance vector, which are used for sampling latent vectors via the reparameterization trick, and the number of latent vectors sampled at each feed-forward pass is denoted $T$. 
As for the deterministic path, context points are input to this path and are processed in the same way as the target points, until an order-invariant representation is procured from the mean aggregator. We also introduce a memory bank to the deterministic path for storing representations.
Subsequently, each target point is concatenated with the $T$ latent vectors and the order-invariant
representations from the deterministic path (note that, practically, the target point and the order-invariant representations from the deterministic path must be copied $T$ times). After the concatenation operation, the $T * r$ feature representations are fed into the decoder $g(\cdot)$ and then the classifier, which outputs $T$ probability distributions over classes for each target point. The final prediction for each target point can be obtained by averaging the $T$ predictions, and the uncertainty is computed as the entropy of the average prediction \cite{kendall2017uncertainties}. 
The ELBO (Eq.~(\ref{eq:elbo})) shows the learning objective.  Specifically, the first term can be achieved by using the cross-entropy loss on the labeled and unlabeled data with their corresponding real labels and pseudo-labels, while the second term is the KL divergence between $q(z|x_{m+1:\ m+r}, y_{m+1:\ m+r})$ and $q(z|x_{1:m}, y_{1:m})$.

{\bf Inference mode.} Concerning a set of test images, they are also passed through the deep neural network at first to obtain their feature representations. Then, they are treated as target points and are fed to the NP model. Since 
the labels of test data are not available, it is impossible to obtain the order-invariant representation from test data. In this case, the stored features in the two memory banks can be directly used. As the bottom diagram of Figure~\ref{fig:np} shows, after the order-invariant representations are obtained from the memory banks, the target points are leveraged in the same way as in the training mode to generate concatenated feature representations for the decoder $g(\cdot)$ and then the~classifier. 

\subsection{Uncertainty-guided Skew-Geometric JS Divergence} 
\label{sec:ugjs}
NP-Match, like many SSL approaches, relies on the use of pseudo-labels for the unlabeled samples. Pseudo-labels, however, are sometimes inaccurate and can lead to the neural network learning poor feature representations. In our pipeline, this can go on to impact the representation procured from the mean-aggregator and hence the model's estimated mean vector, variance vector, and global latent vectors (see ``Latent Path" in \ref{fig:np}).
To remedy this, similarly to how the KL divergence term in the ELBO (Eq.~(\ref{eq:elbo})) is used to learn global latent variables \cite{garnelo2018neural}, we propose a new distribution divergence, called the uncertainty-guided skew-geometric JS divergence ($JS^{G_{\alpha_u}}$). We first formalize the definition of $JS^{G_{\alpha_u}}$:
 
{\bf Definition 4.3.3.1.} \emph{Let $(\Omega, \Sigma)$ be a measurable space, where $\Omega$ denotes the sample space, and $\Sigma$ denotes the $\sigma$-algebra of measurable events. $P$ and $Q$ are two probability measures defined on the measurable space. Concerning a positive measure\footnote{Specifically, the positive measure is usually the Lebesgue measure with the Borel $\sigma$-algebra $\mathcal{B}(\mathbb{R}^d)$ or the counting measure with the power set $\sigma$-algebra $2^\Omega$.}, which is denoted as $\mu$, the uncertainty-guided skew-geometric JS divergence ($JS^{G_{\alpha_u}}$) can be defined as: }
\begin{equation} 
\begin{aligned}
\label{eq:JS_skew}
&JS^{G_{\alpha_u}}(p, q) =  (1 - \alpha_u)  \int p \ log \frac{p}{G(p,q)_{\alpha_u}} d\mu +  \alpha_u \int q \ log \frac{q}{G(p,q)_{\alpha_u}} d\mu,
\end{aligned}
\end{equation} 
\emph{where $p$ and $q$ are the Radon-Nikodym derivatives of $P$ and $Q$ with respect to $\mu$,  the scalar $\alpha_u \in [0, 1]$ is  calculated based on the uncertainty, and $G(p,q)_{\alpha_u} = p^{1-\alpha_u}q^{\alpha_u}$ $/$ $(\int_{\Omega} p^{1-\alpha_u}q^{\alpha_u} d\mu)$. The dual form of $JS^{G_{\alpha_u}}$ is given by: 
}
\begin{equation} 
\begin{aligned}
\label{eq:JS_skew_dual}
&JS_*^{G_{\alpha_u}}(p, q) = (1 - \alpha_u)  \int G(p,q)_{\alpha_u} \ log \frac{G(p,q)_{\alpha_u}}{p} d\mu +   \alpha_u \int G(p,q)_{\alpha_u} \ log \frac{G(p,q)_{\alpha_u}}{q} d\mu.
\end{aligned}
\end{equation} 

The proposed $JS^{G_{\alpha_u}}$ is an extension of the skew-geo\-met\-ric JS diver\-gence first proposed by Nielsen {\it et al.} \cite{nielsen2020generalization}. Specifically, it generalizes the JS divergence with abstract means (quasi-arithmetic means \cite{niculescu2006convex}), in which a scalar $\alpha$ is defined to control the degree of divergence skew.\footnote{The divergence skew means how closely related the intermediate distribution (the abstract mean of $p$ and $q$) is to $p$ or $q$.} By selecting the weighted geometric mean $p^{1-\alpha}q^{\alpha}$,  such generalized JS divergence becomes the skew-geometric JS divergence, which can be easily applied to the Gaussian distribution because of its property that the weighted product of exponential family distributions stays in the exponential family \cite{nielsen2009statistical}.  
Our $JS^{G_{\alpha_u}}$ extends such divergence by incorporating the uncertainty into the scalar $\alpha$ to dynamically adjust the divergence skew. We assume the real variational distribution of the global latent variable under the supervised learning to be $q^*$. If the framework is trained with real labels, the condition $q(z|x_{m+1:\ m+r}, y_{m+1:\ m+r})=q(z|x_{1:m}, y_{1:m})=q^*$ will hold after training, since they are all the marginal distributions of the same stochastic process. 
However, as for SSL, $q(z|x_{m+1:\ m+r}, y_{m+1:\ m+r})$ and $q(z|x_{1:m}, y_{1:m})$ are no longer equal to $q^*$, as some low-quality representations are involved during training, which affect the estimation of $q(z|x_{m+1:\ m+r}, y_{m+1:\ m+r})$ and $q(z|x_{1:m}, y_{1:m})$. Our proposed $JS^{G_{\alpha_u}}$ solves this issue by introducing an intermediate distribution that is  calculated via $G(q(z|x_{1:m},y_{1:m}),$ $q(z|x_{m+1:\ m+r},y_{m+1:\ m+r}))_{\alpha_u}$, where $\alpha_u = u_{c_{avg}} / (u_{c_{avg}} + u_{t_{avg}})$.   Here, $u_{c_{avg}}$  denotes the average value over the uncertainties of the predictions of context points, and $u_{t_{avg}}$ represents the average value over that of target points. With this setting, the intermediate distribution is usually close to  $q^*$. For example, when $u_{c_{avg}}$ is large, and $u_{t_{avg}}$ is small, which means that there are many low-quality feature presentations  involved for calculating $q(z|x_{1:m}, y_{1:m})$, and $q(z|x_{m+1:\ m+r}, y_{m+1:\ m+r})$ is closer to $q^*$, then $G(q(z|x_{1:m}, y_{1:m}), q(z|x_{m+1:\ m+r}, y_{m+1:\ m+r}))_{\alpha_u}$ will be close to $q(z|x_{m+1:\ m+r}, y_{m+1:\ m+r})$, and as a result, the network is optimized to learn the distribution of the global latent variable in the direction to $q^*$, which mitigates the issue to some extent.\footnote{As long as one of $q(z|x_{1:m}, y_{1:m})$ and $q(z|x_{m+1:\ m+r},$ $y_{m+1:\ m+r})$ is close to $q*$, the proposed $JS^{G_{\alpha_u}}$ mitigates the issue, but $JS^{G_{\alpha_u}}$ still has difficulties to solve the problem when both of their calculations involve many low-quality representations.}  Concerning  the variational distribution being  supposed to be a Gaussian distribution, we introduce the following theorem (with proof in the Appendix B) for calculating $JS^{G_{\alpha_u}}$ on Gaussian~distributions:

\smallskip
{\bf Theorem 4.3.3.1.} \emph{Given two multivariate Gaussians $\mathcal{N}_1(\mu_1, \Sigma_1)$ and $\mathcal{N}_2(\mu_2, \Sigma_2)$, the following holds:}  
\begin{equation} 
\begin{aligned}
\label{eq:JS_skew_gaussian}
&JS^{G_{\alpha_u}}(\mathcal{N}_1, \mathcal{N}_2) = \\ 
&\frac{1}{2}(tr(\Sigma^{-1}_{\alpha_u}((1 - \alpha_u)\Sigma_1 + \alpha_u\Sigma_2)) + (1-\alpha_u)(\mu_{\alpha_u} - \mu_1)^T\Sigma^{-1}_{\alpha_u}(\mu_{\alpha_u} - \mu_1) + \\
&\alpha_u(\mu_{\alpha_u} - \mu_2)^T\Sigma^{-1}_{\alpha_u}(\mu_{\alpha_u} - \mu_2) +  log[\frac{det[\Sigma_{\alpha_u}]}{det[\Sigma_1]^{1-\alpha_u} det[\Sigma_2]^{\alpha_u}}] - D)\\
& JS_*^{G_{\alpha_u}}(\mathcal{N}_1, \mathcal{N}_2) = \\
&\frac{1}{2}( log[\frac{det[\Sigma_1]^{1-\alpha_u} det[\Sigma_2]^{\alpha_u}}{det[\Sigma_{\alpha_u}]}] + \alpha_u\mu_2^T\Sigma_2^{-1}\mu_2   -\mu_{\alpha_u}^T\Sigma_{\alpha_u}^{-1}\mu_{\alpha_u} + (1 - \alpha_u)\mu_1^T\Sigma_1^{-1}\mu_1),
\end{aligned}
\end{equation}  
\emph{where $\Sigma_{\alpha_u}=((1-\alpha_u)\Sigma_1^{-1} + \alpha_u\Sigma_2^{-1})^{-1}$ and $\mu_{\alpha_u}=\Sigma_{\alpha_u}((1-\alpha_u)\Sigma_1^{-1}\mu_1 + \alpha_u\Sigma_2^{-1}\mu_2)$,  $D$ denotes the number of dimension, and $det[\cdot]$ represents the determinant.}
\smallskip

With Theorem 1, one can calculate $JS^{G_{\alpha_u}}$ or its dual form $JS_*^{G_{\alpha_u}}$ based on the mean vector and the variance vector, and use $JS^{G_{\alpha_u}}$ or $JS_*^{G_{\alpha_u}}$ to replace the original KL divergence term in the ELBO (Eq.~(\ref{eq:elbo})) for training the whole framework. When the two distributions are diagonal Gaussians, $\Sigma_1$ and $\Sigma_2$ can be implemented by diagonal matrices with the variance vectors for calculating $JS^{G_{\alpha_u}}$ or~$JS_*^{G_{\alpha_u}}$. 

\subsection{Loss Functions}

To calculate loss functions, reliable pseudo-labels are required for unlabeled data. In practice, to select reliable unlabeled samples from $\mathcal{U}$ and  their corresponding pseudo-labels, we preset a confidence threshold ($\tau_c$) and an uncertainty threshold ($\tau_u$). In particular, as for unlabeled data $x_{i}^{u}$, NP-Match gives its prediction $p(y|Aug_w(x_{i}^{u}))$ and associated uncertainty estimate under the inference mode, where $Aug_w(\cdot)$ denotes the weak augmentation. When the highest prediction score $max(p(y|Aug_w(x_{i}^{u})))$ is higher than $\tau_c$, and the uncertainty is smaller than $\tau_u$, the sample will be chosen, and we denote the selected sample as $x_{i}^{u_c}$, since the model is certain about his prediction, and the pseudo-label of $x_{i}^{u_c}$ is $\hat y_i= arg\ max(p(y|Aug_w(x_{i}^{u_c})))$. 
Concerning $\mu B$ unlabeled samples in $\mathcal{U}$, we assume $B_c$ unlabeled samples are selected from them in each feedforward pass. 
According to the ELBO (Eq.~(\ref{eq:elbo})), three loss terms are used for training, namely, $L_{cls}$, $L^u_{cls}$, and $JS^{G_{\alpha_u}}$. 
For each input (labeled or unlabeled), the NP model can give $T$ predictions, and hence~$L_{cls}$ and $L^u_{cls}$ are defined as:
\begin{equation} 
\begin{aligned}
\label{eq:cross-entropy}
&L_{cls} = \frac{1}{B \times T} \sum^{B}_{i=1} \sum^{T}_{j=1} H(y^*_i, p_j(y|Aug_w(x_i))), \\
&L^u_{cls} = \frac{1}{B_c \times T} \sum^{B_c}_{i=1} \sum^{T}_{j=1} H(\hat y_i, p_j(y|Aug_s(x_{i}^{u_{c}}))), 
\end{aligned}
\end{equation} 
where $Aug_s(\cdot)$ denotes the strong augmentation, $y^*_i$ represents the real label for the labeled sample $x_{i}$, and $H(\cdot, \cdot)$ denotes the cross-entropy between two distributions. Thus, the total loss function is given by:
\begin{equation}
\label{eq:overall}
L_{total} = L_{\text{cls}} + \lambda_u L^u_{cls} + \beta JS^{G_{\alpha_u}},
\end{equation}
where $\lambda_u$ and $\beta$ are coefficients. During training, we followed previous work \cite{sohn2020fixmatch,zhang2021flexmatch, rizve2021defense, li2021comatch} to utilize  the exponential moving average (EMA) technique. It is worth noting that, in the real implementation, NP-Match only preserves the averaged representation over all representations in each memory bank after training, which just takes up negligible storage space.

\section{Experiments}
\label{4.4}

We conducted experiments on three different settings, namely standard semi-supervised image classification, imbalanced semi-supervised image classification, and multi-label semi-supervised image classification. Compared to the standard setting, imbalanced semi-supervised image classification is more challenging, since deep models may overfit to frequent classes, leading to inaccurate pseudo-labels. In addition, multi-label classification is also challenging, since it increases the risk of making wrong predictions for unlabeled data, therefore offering a more tough and realistic scenario for evaluating our framework. These two challenging settings are indeed more aligned with real-world scenarios than the standard setting, and evaluating our method under them can provide stronger evidence to verify the effectiveness of our method. To save space, implementation details are given in the Appendix B. 


\subsection{Datasets}

For the standard semi-supervised image classification task, we conducted our experiments on four widely used public SSL benchmarks, including CIFAR-10 \cite{krizhevsky2009learning}, CIFAR-100 \cite{krizhevsky2009learning}, STL-10 \cite{coates2011analysis}, and ImageNet \cite{deng2009imagenet}. CIFAR-10 and CIFAR-100 contain 50,000 images of size $32\times32$ from 10 and 100 classes, respectively. We evaluated NP-match on these two datasets following the evaluation settings used in previous works \cite{sohn2020fixmatch, zhang2021flexmatch, li2021comatch}.
The STL-10 dataset has 5000 labeled samples with size $96\times96$ from 10 classes and 100,000 unlabeled samples, and it is more difficult than CIFAR,  since STL-10  has a number of out-of-distribution images in the unlabeled set. We follow the experimental settings for STL-10 as detailed in \cite{zhang2021flexmatch}. Finally, ImageNet contains around 1.2 million images from 1000 classes. Following the experimental settings in \cite{zhang2021flexmatch}, we used 100K labeled data, namely, 100 labels per class.

For the imbalanced semi-supervised image classification task, we still chose CIFAR-10 \cite{krizhevsky2009learning}, CIFAR-100 \cite{krizhevsky2009learning}, and STL-10 \cite{coates2011analysis}, but with imbalanced class distribution for both labeled and unlabaled data. By following \cite{oh2022daso}, the imbalanced settings were achieved by exponentially decreasing the number of samples within each class. Specifically, the head class size is denoted as $N_1$ ($M_1$) and the imbalance ratio is denoted as $\gamma_l$ ($\gamma_u$) for labeled (unlabeled) data separately, where $\gamma_l$ and $\gamma_u$ are independent from each other.

For the multi-label semi-supervised image classification task, we chose a widely-used medical dataset, named Chest X-Ray14. Chest X-Ray14 is a collection of 112,120 chest X-ray images from 30,805 patients, with 14 labels (each label is a disease) and {\it No Finding} class. Note that each patient can have more than one label, leading to a multi-label classification problem. We employed the official training and testing data split, and we followed \cite{liu2022acpl} to use area under the ROC curve (AUC) as the evaluation metric.

\begin{table*}
\centering 
\resizebox{\textwidth}{!}{
\begin{tabular}{@{}cccccccccc@{}}
 \toprule[1pt]
Dataset  & \multicolumn{3}{c}{CIFAR-10} &\multicolumn{3}{c}{CIFAR-100}  & \multicolumn{3}{c}{STL-10}\\
 \hline  
Label Amount & 40 & 250 & 4000 & 400 & 2500 & 10000 & 40 & 250 & 1000\\
 \hline 
  \quad MixMatch \cite{berthelot2019mixmatch} &  36.19 {\small($\pm$6.48)} &  13.63 {\small($\pm$0.59)} &  6.66 {\small($\pm$0.26)} &  67.59 {\small($\pm$0.66)} & 39.76 {\small($\pm$0.48)}  & 27.78 {\small($\pm$0.29)} &  54.93 {\small($\pm$0.96)} &  34.52 {\small($\pm$0.32)} &  21.70 {\small($\pm$0.68)} \\
 \quad ReMixMatch \cite{berthelot2019remixmatch}  &  9.88 {\small($\pm$1.03)} &  6.30 {\small($\pm$0.05)} &  4.84 {\small($\pm$0.01)} &  42.75 {\small($\pm$1.05)} & {\bf 26.03} {\small($\pm$0.35)}  & {\bf 20.02} {\small($\pm$0.27)} &  32.12 {\small($\pm$6.24)} &  12.49 {\small($\pm$1.28)} &  6.74 {\small($\pm$0.14)} \\
 \quad UDA \cite{xie2019unsupervised}  &  10.62 {\small($\pm$3.75)} &  5.16 {\small($\pm$0.06)} &  4.29 {\small($\pm$0.07)} &  46.39 {\small($\pm$1.59)}  & 27.73 {\small($\pm$0.21)} & 22.49 {\small($\pm$0.23)}  & 37.42 {\small($\pm$8.44)} &  9.72 {\small($\pm$1.15)} &  6.64 {\small($\pm$0.17)} \\
 \quad CoMatch \cite{li2021comatch}  &   6.88 {\small($\pm$0.92)} &  4.90 {\small($\pm$0.35)} &  4.06 {\small($\pm$0.03)} &  40.02 {\small($\pm$1.11)}  & 27.01 {\small($\pm$0.21)} & 21.83 {\small($\pm$0.23)}  & 31.77 {\small($\pm$2.56)} & 11.56 {\small($\pm$1.27)} &   8.66  {\small($\pm$0.41)} \\
 \quad SemCo \cite{nassar2021all}  &   7.87 {\small($\pm$0.22)} &   5.12 {\small($\pm$0.27)} &  {\bf 3.80} {\small($\pm$0.08)} &  44.11 {\small($\pm$1.18)}  & 31.93 {\small($\pm$0.33)} & 24.45 {\small($\pm$0.12)}  & 34.17 {\small($\pm$2.78)} & 12.23 {\small($\pm$1.40)} &   7.49  {\small($\pm$0.29)} \\
 \quad FlexMatch \cite{zhang2021flexmatch} &   4.96 {\small($\pm$0.06)} &  4.98 {\small($\pm$0.09)} &  4.19 {\small($\pm$0.01)} &  39.94 {\small($\pm$1.62)}  & 26.49 {\small($\pm$0.20)} & 21.90 {\small($\pm$0.15)}  & 29.15 {\small($\pm$4.16)} & {\bf 8.23} {\small($\pm$0.39)} &   5.77  {\small($\pm$0.18)} \\
 \quad UPS \cite{rizve2021defense}  &  5.26 {\small($\pm$0.29)} &  5.11 {\small($\pm$0.08)} &  4.25 {\small($\pm$0.05)} &  41.07 {\small($\pm$1.66)}  & 27.14 {\small($\pm$0.24)} & 21.97 {\small($\pm$0.23)}  & 30.82  {\small($\pm$2.16)} & 9.77 {\small($\pm$0.44)} &   6.02  {\small($\pm$0.28)} \\
 \quad FixMatch \cite{sohn2020fixmatch}  &   7.47 {\small($\pm$0.28)} &  {\bf 4.86} {\small($\pm$0.05)} &  4.21 {\small($\pm$0.08)} &  46.42 {\small($\pm$0.82)}  & 28.03 {\small($\pm$0.16)} & 22.20 {\small($\pm$0.12)}  & 35.96 {\small($\pm$4.14)} & 9.81 {\small($\pm$1.04)} &   6.25  {\small($\pm$0.33)} \\
 \quad NP-Match (ours) &  {\bf 4.91} {\small($\pm$0.04)} &  4.96 {\small($\pm$0.06)} &  4.11 {\small($\pm$0.02)} & {\bf 38.91} {\small($\pm$0.99)}  & {\bf 26.03} {\small($\pm$0.26)} & 21.22 {\small($\pm$0.13)}  & {\bf 14.20} {\small($\pm$0.67)} & 9.51 {\small($\pm$0.37)} & {\bf 5.59}  {\small($\pm$0.24)}\\
  \bottomrule[1pt]
  \end{tabular}} 
 \caption{Comparison with SOTA results on CIFAR-10, CIFAR-100, and STL-10. The error rates are reported with standard deviation. } 
 \label{tab:compare_sota} 
 \end{table*}

\subsection{Main Results}
\label{sec:main_results}

In the following, we report the main experimental results in terms of accuracy, the reliability of uncertainty estimates, and time consumption, under different settings.

\subsubsection{Standard Semi-Supervised Image Classification Experimental Results} 

First, in Table~\ref{tab:compare_sota}, we compare NP-Match with SOTA SSL image classification methods on CIFAR-10, CIFAR-100, and STL-10. We see that NP-Match outperforms SOTA results or achieves competitive results under different SSL settings. We highlight two key observations. First, NP-Match outperforms all other methods by a wide margin on all three benchmarks under the most challenging settings,  where the number of labeled samples is smallest.
Second, NP-Match is compared to  UPS\footnote{Note that UPS \cite{rizve2021defense} does not use strong augmentations, thus we re-implemented it with RandAugment \cite{cubuk2020randaugment} for fair comparisons.}, since the UPS framework is the most recent probabilistic model for semi-supervised image classification, and NP-Match completely outperforms them on all three benchmarks with different SSL settings. This suggests that NPs can be a good alternative to MC dropout in probabilistic approaches to semi-supervised learning tasks. We suppose that the superior performance of our method is due to the well-calibrated uncertainty estimates given by the NP model, which offer a reliable standard for selecting unlabeled data with trustworthy pseudo-labels.

 \begin{figure*}[t]
\centering
\includegraphics[width=\linewidth]{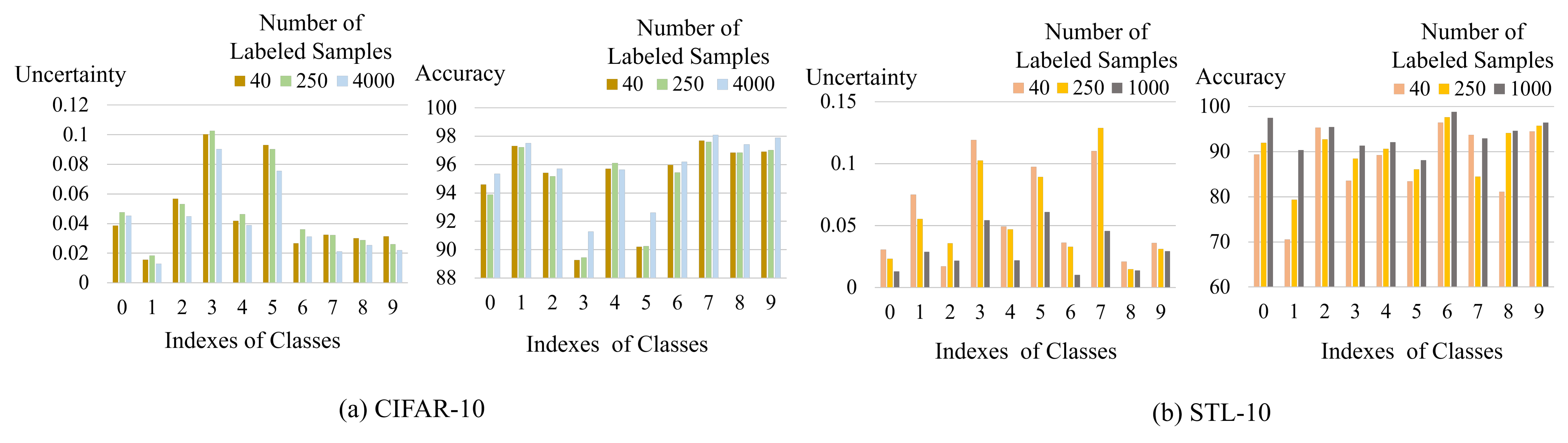}
\vspace{-5ex}
\caption{Analysis of average class-wise uncertainty and accuracy.}
\label{fig:u_vs_p} 
\end{figure*}

\begin{table}[t]
\centering 
\resizebox{0.6\textwidth}{!}{
\begin{tabular}{@{}ccccccc @{}}
 \toprule[1pt]
Dataset  & \multicolumn{3}{c}{CIFAR-10} &\multicolumn{3}{c}{STL-10}   \\
 \hline  
Label Amount & 40 & 250 & 4000 & 40 & 250 & 1000 \\
 \hline 
 UPS (MC Dropout) & 7.96  & 7.02   & {\bf 5.82}  & 17.23  &  9.65  & 5.69 \\
 NP-Match & {\bf 7.23}  & {\bf 6.85}  &  5.89 &  {\bf 12.45}  &  {\bf 8.72}   &  {\bf 5.28} \\
  \bottomrule[1pt]
  \end{tabular}}\vspace*{-1ex}
 \caption{Expected UCEs (\%) of the MC-dropout-based model (i.e., UPS \cite{rizve2021defense}) and of NP-Match on the test sets of CIFAR-10 and STL-10.}  
 \label{tab:ablation_uce} 
 \end{table}

Second, we analyse the relationship between the average class-wise uncertainty and accuracy at test phase on CIFAR-10 and STL10. From Figure~\ref{fig:u_vs_p}, we empirically observe that: (1) when more labeled data are used for training, the average uncertainty of samples' predictions for most classes decreases. This is consistent with the property of NPs and GPs where the model is less uncertain with regard to its prediction when more real and correct labels are leveraged; (2) the classes with higher average uncertainties have lower accuracy, meaning that our method can output reliable uncertainty estimates, which can be used for choosing unlabeled samples. However, in some classes, more labeled data does not lead to higher performance and lower uncertainty. We hypothesize that this phenomenon is because the newly added labeled data in those classes are of low quality, leading to higher aleatoric uncertainty.

Third, the expected uncertainty calibration error (UCE) of our method is also calculated to evaluate the uncertainty estimation. The expected UCE is used to measure the miscalibration of uncertainty \cite{laves2020calibration}, which is an analogue to the expected calibration error (ECE) \cite{guo2017calibration, naeini2015obtaining}. The low expected UCE indicates that the model is certain when making accurate predictions and that the model is uncertain when making inaccurate predictions. More details about the expected UCE can be found in previous works \cite{laves2020calibration, krishnan2020improving}. The results of NP-Match and the MC-dropout-based model (i.e., UPS \cite{rizve2021defense}) are shown in Table~\ref{tab:ablation_uce}; their comparison shows that NP-Match can output more reliable and well-calibrated uncertainty estimates. 

 \begin{figure}[t]
\centering
\includegraphics[width=0.75\linewidth]{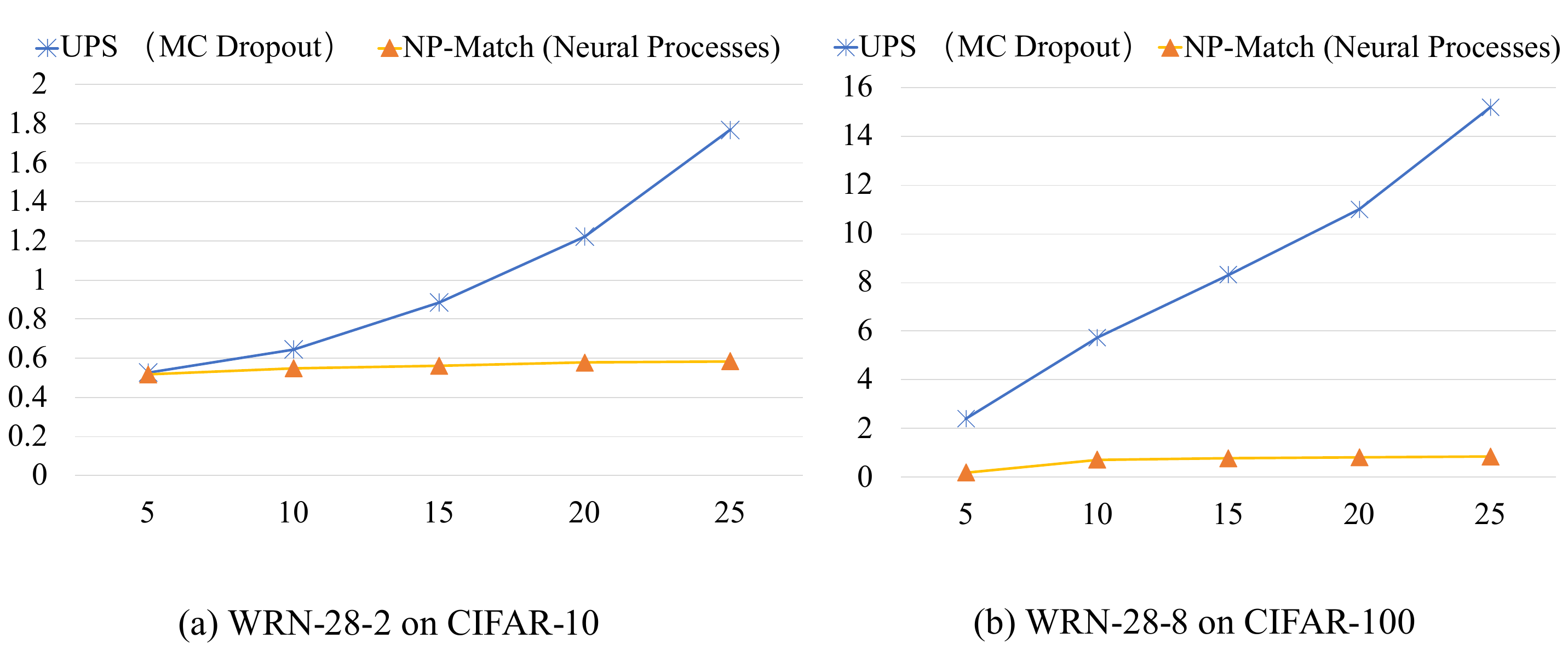}
\vspace{-1ex}
\caption{Time consumption of estimating uncertainty for the MC-dropout-based model (i.e., UPS \cite{rizve2021defense}) and  NP-Match. The horizontal axis refers to the number of predictions used for the uncertainty quantification, and the vertical axis indicates the time consumption (sec).} 
\label{fig:speed}
\end{figure}

\begin{table}
\centering 
\resizebox{0.45\textwidth}{!}{
\begin{tabular}{@{}c|ccc@{}}
 \toprule[1pt]
   & Method & Top-1  & Top-5 \\
 \hline 
  \multirow{3}{*}{\shortstack{Deterministic \\ Methods}} & FixMatch \cite{sohn2020fixmatch} &   43.66  &  21.80 \\ 
 & FlexMatch \cite{zhang2021flexmatch} &  41.85  & 19.48 \\ 
  &CoMatch \cite{li2021comatch}&  42.17 &  19.64 \\
 \hline
\multirow{2}{*}{\shortstack{Probabilistic \\ Methods}} &UPS \cite{rizve2021defense} &  42.69  &  20.23 \\
 & NP-Match  &  {\bf 41.78} &  {\bf 19.33} \\ 
  \bottomrule[1pt]
  \end{tabular}} 
 \caption{Error rates of SOTA methods on ImageNet.} \vspace*{-2ex}
 \label{tab:compare_imagenet} 
 \end{table}

Furthermore, we compare the running time of NP-Match and the MC dropout-based probabilistic model (i.e., UPS  \cite{rizve2021defense}). We use a batch of 16 samples and two network architectures that are widely used in previous works \cite{zagoruyko2016wide, zhang2021flexmatch, sohn2020fixmatch, li2021comatch}, namely, WRN-28-2 on CIFAR-10 (Figure~\ref{fig:speed} (a)) and WRN-28-8 on CIFAR-100 (Figure~\ref{fig:speed} (b)).
In (a), we observe that when the number of predictions ($T$) increases,  the time cost of the UPS framework rises quickly, but the time cost of NP-Match grows slowly. In (b), we observe that the time cost gap between these two methods is even larger when a larger model is tested on a larger dataset. This demonstrates that NP-Match is significantly more computationally efficient than MC dropout-based methods. The reason is that NP-Match does not require multiple feed-forward passes for making predictions and quantifying uncertainty, which makes it a better choice than MC dropout to be incorporated with large backbone networks.

Finally, Table~\ref{tab:compare_imagenet} shows the experiments conducted on ImageNet. Here, NP-Match achieves a SOTA  performance, suggesting that it is effective at handling challenging large-scale datasets. Note that previous works usually evaluate their frameworks under distinct SSL settings, and thus it is hard to compare different methods directly. Therefore, we re-evaluate another two methods proposed recently under the same SSL setting with the same training details, namely, UPS and CoMatch.

It is worth noting that the time consumption experiments and the results on ImageNet demonstrate the scalability of NP-Match on large-scale and complex datasets. This can be observed through two facts. Firstly, ImageNet images are quite different from those in CIFAR, as each has a complex composition, i.e., containing multiple objects or contents unrelated to its label. This increases the difficulty of recognition and uncertainty quantification, but NP-Match can still achieve good performance. Secondly, the low time consumption allows NP-Match to be trained with over millions of images, indicating that one can arbitrarily enlarge the training data to improve performance without considering any additional computational burdens brought by NPs.

\begin{table*}
\centering 

\resizebox{\textwidth}{!}{
\begin{tabular}{@{}ccccccccc@{}}
 \toprule[1pt]
      & \multicolumn{4}{c}{CIFAR-10-LT} &\multicolumn{4}{c}{CIFAR-100-LT} \\
    & \multicolumn{2}{c}{$\gamma=\gamma_l=\gamma_u=100$} &\multicolumn{2}{c}{$\gamma=\gamma_l=\gamma_u=150$}  & \multicolumn{2}{c}{$\gamma=\gamma_l=\gamma_u=10$} &\multicolumn{2}{c}{$\gamma=\gamma_l=\gamma_u=20$} \\
 \cmidrule(r){2-3} \cmidrule(r){4-5} \cmidrule(r){6-7} \cmidrule(r){8-9} 
 \multirow{2}{*}{Methods} & $N_1$=500 & $N_1$=1500 & $N_1$=500 & $N_1$=1500  & $N_1$=50 & $N_1$=150 & $N_1$=50 & $N_1$=150  \\
  & $M_1$=4000 & $M_1$=3000 & $M_1$=4000 & $M_1$=3000 & $M_1$=400 & $M_1$=300 & $M_1$=400 & $M_1$=300 \\
 \cmidrule(r){1-1} \cmidrule(r){2-3} \cmidrule(r){4-5} \cmidrule(r){6-7} \cmidrule(r){8-9} 
  \quad DARP \cite{kim2020distribution} &  74.50 {\small($\pm$0.78)} &  77.80 {\small($\pm$0.63)} &  67.20  {\small($\pm$0.32)} &  73.60 {\small($\pm$0.73)} & 49.40 {\small($\pm$0.20)}  & 58.10 {\small($\pm$0.44)} &  43.40 {\small($\pm$0.87)} & 52.20 {\small($\pm$0.66)}  \\
  \quad CReST \cite{wei2021crest} &  76.30 {\small($\pm$0.86)} &  78.10 {\small($\pm$0.42)} &  67.50 {\small($\pm$0.45)} &  73.70 {\small($\pm$0.34)} & 44.50 {\small($\pm$0.94)}  & 57.40 {\small($\pm$0.18)} &  40.10 {\small($\pm$1.28)} &  52.10 {\small($\pm$0.21)}  \\ 
  \quad DASO \cite{oh2022daso} &  76.00  {\small($\pm$0.37)} &  79.10 {\small($\pm$0.75)} &  70.10 {\small($\pm$1.81)} &  75.10 {\small($\pm$0.77)} & 49.80 {\small($\pm$0.24)}  & 59.20 {\small($\pm$0.35)} &  43.60 {\small($\pm$0.09)} &  52.90  {\small($\pm$0.42)}  \\ 
  \quad  ABC \cite{lee2021abc} + DASO \cite{oh2022daso} &  80.10 {\small($\pm$1.16)} &  83.40 {\small($\pm$0.31)} & 70.60 {\small($\pm$0.80)} &  80.40 {\small($\pm$0.56)} & 50.20 {\small($\pm$0.62)}  & 60.00 {\small($\pm$0.32)} &  44.50 {\small($\pm$0.25)} &  {\bf 55.30}  {\small($\pm$0.53)}  \\ 
   \quad LA \cite{menon2021long} + DARP \cite{kim2020distribution} &  76.60 {\small($\pm$0.92)} &  80.80 {\small($\pm$0.62)} &  68.20 {\small($\pm$0.94)} &  76.70 {\small($\pm$1.13)} & 50.50 {\small($\pm$0.78)}  & 59.90 {\small($\pm$0.32)} &  44.40 {\small($\pm$0.65)} &  53.80 {\small($\pm$0.43)}  \\
  \quad  LA \cite{menon2021long} + CReST \cite{wei2021crest} &  76.70 {\small($\pm$1.13)} &  81.10 {\small($\pm$0.57)} &  70.90  {\small($\pm$1.18)} &  77.90 {\small($\pm$0.71)} & 44.00 {\small($\pm$0.21)}  & 57.10 {\small($\pm$0.55)} &  40.60 {\small($\pm$0.55)} &  52.30  {\small($\pm$0.20)}  \\ 
  \quad  LA \cite{menon2021long} + DASO   \cite{oh2022daso} &  77.90 {\small($\pm$0.88)} &  82.50  {\small($\pm$0.08)} &  70.10  {\small($\pm$1.68)} &  79.00 {\small($\pm$2.23)} & 50.70 {\small($\pm$0.51)}  & {\bf 60.60} {\small($\pm$0.71)} &  44.10 {\small($\pm$0.61)} &  55.10 {\small($\pm$0.72)}  \\ 
  \cmidrule(r){1-1} \cmidrule(r){2-3} \cmidrule(r){4-5} \cmidrule(r){6-7} \cmidrule(r){8-9} 
    \quad DASO w. UPS \cite{rizve2021defense}   &  75.44 {\small($\pm$0.79)}  &  78.11 {\small($\pm$0.43)}  & 69.64 {\small($\pm$1.01)}  & 74.39 {\small($\pm$0.83)}   &  49.16 {\small($\pm$0.33)}  &  57.87 {\small($\pm$0.33)} &  43.02 {\small($\pm$0.38)}   &   52.23 {\small($\pm$0.61)}  \\ 
  \quad LA  + DASO w. UPS \cite{rizve2021defense}   &  78.89 {\small($\pm$0.24)}  & 81.24 {\small($\pm$0.54)}    & 71.39 {\small($\pm$0.78)}  & 77.83 {\small($\pm$0.94)}  &  50.04 {\small($\pm$0.47)}   & 58.92 {\small($\pm$0.49)}  & 43.95 {\small($\pm$0.54)}    &   53.98 {\small($\pm$0.82)}   \\ 
    \quad ABC  + DASO w. UPS \cite{rizve2021defense}   &  79.22 {\small($\pm$0.31)}  & 81.02 {\small($\pm$0.39)}    & 71.67 {\small($\pm$0.65)}  & 78.61 {\small($\pm$0.88)}  &  50.39 {\small($\pm$0.67)}   & 58.55 {\small($\pm$0.69)}  & 44.07 {\small($\pm$0.38)}    &   54.12 {\small($\pm$0.77)}   \\ 
   \cmidrule(r){1-1} \cmidrule(r){2-3} \cmidrule(r){4-5} \cmidrule(r){6-7} \cmidrule(r){8-9} 
 
  \quad DASO w. NPs   &  76.06 {\small($\pm$0.11)}  & 79.23 {\small($\pm$0.42)}   & 70.13 {\small($\pm$1.39)} & 75.17 {\small($\pm$0.81)}  &  49.46 {\small($\pm$0.42)} &  57.66 {\small($\pm$0.55)}  &  43.32 {\small($\pm$0.83)}  &   51.96 {\small($\pm$0.51)}   \\ 
   
  \quad LA  + DASO w. NPs  & {\bf 80.44} {\small($\pm$0.42)}   &   {\bf 84.02} {\small($\pm$0.23)} &  {\bf 73.24} {\small($\pm$0.94)}  & {\bf 81.25} {\small($\pm$0.87)}  &  {\bf 50.97} {\small($\pm$0.55)}  &  58.77  {\small($\pm$0.69)}  &  {\bf 44.65} {\small($\pm$0.66)}  &  53.86 {\small($\pm$0.31)}  \\ 
 
  \bottomrule[1pt]
  \end{tabular}} 
  \caption{Comparison with SOTA results on long-tailed CIFAR-10 (CIFAR-10-LT) and long-tailed CIFAR-100 (CIFAR-100-LT) under $\gamma_l = \gamma_u$  setup. The accuracy is reported with standard deviation. } 
  \label{tab:long-tail}
 \end{table*}

 \begin{table}[t]
\centering 

\resizebox{\textwidth}{!}{
\begin{tabular}{@{}ccccccccc@{}}
 \toprule[1pt]
      & \multicolumn{4}{c}{CIFAR-10-LT ($\gamma_l \neq \gamma_u$)}  &\multicolumn{4}{c}{STL-10-LT ($\gamma_u=N/A$)}  \\
    & \multicolumn{2}{c}{$\gamma_u=1 (uniform)$} &\multicolumn{2}{c}{$\gamma_u=\frac{1}{100} (reversed)$}  & \multicolumn{2}{c}{$\gamma_l=10$} &\multicolumn{2}{c}{$\gamma_l=20$} \\
 \cmidrule(r){2-3} \cmidrule(r){4-5} \cmidrule(r){6-7} \cmidrule(r){8-9} 
 \multirow{2}{*}{Methods} & $N_1$=500 & $N_1$=1500 & $N_1$=500 & $N_1$=1500  & $N_1$=150 & $N_1$=450 & $N_1$=150 & $N_1$=450  \\
  & $M_1$=4000 & $M_1$=3000 & $M_1$=4000 & $M_1$=3000 & $M_1$=100k & $M_1$=100k & $M_1$=100k & $M_1$=100k \\
 \cmidrule(r){1-1} \cmidrule(r){2-3} \cmidrule(r){4-5} \cmidrule(r){6-7} \cmidrule(r){8-9} 
  \quad DARP \cite{kim2020distribution} &  82.50 {\small($\pm$0.75)} &  84.60 {\small($\pm$0.34)} &  70.10  {\small($\pm$0.22)} &  80.00 {\small($\pm$0.93)} & 66.90 {\small($\pm$1.66)}  & 75.60 {\small($\pm$0.45)} &  59.90    {\small($\pm$2.17)} &  72.30 {\small($\pm$0.60)}  \\
  \quad CReST \cite{wei2021crest} &   82.20  {\small($\pm$1.53)} &  86.40  {\small($\pm$0.42)} & 62.90  {\small($\pm$1.39)} &   72.90 {\small($\pm$2.00)} & 61.20 {\small($\pm$1.27)}  &  71.50  {\small($\pm$0.96)} &  56.00 {\small($\pm$3.19)} &   68.50 {\small($\pm$1.88)}  \\ 
  \quad DASO \cite{oh2022daso} & 86.60  {\small($\pm$0.84)} &  {\bf 88.80} {\small($\pm$0.59)} &  71.00 {\small($\pm$0.95)} &  80.30 {\small($\pm$0.65)} & {\bf 70.00} {\small($\pm$1.19)}  & 78.40 {\small($\pm$0.80)} & 65.70 {\small($\pm$1.78)} & {\bf 75.30}  {\small($\pm$0.44)}  \\ 
    \cmidrule(r){1-1} \cmidrule(r){2-3} \cmidrule(r){4-5} \cmidrule(r){6-7} \cmidrule(r){8-9} 
  \quad DASO w. UPS \cite{rizve2021defense}  &   86.32  {\small($\pm$0.41)}   &  87.94 {\small($\pm$0.63)}  & 70.62 {\small($\pm$0.78)}  & 79.59 {\small($\pm$1.02)}   &  69.04 {\small($\pm$0.98)}   & 77.74 {\small($\pm$0.64)}   & 65.10 {\small($\pm$1.22)}   &  74.03 {\small($\pm$0.73)}   \\ 
  \quad DASO w. NPs   & {\bf 87.50}  {\small($\pm$0.65)}   &  88.21  {\small($\pm$0.58)}  & {\bf 73.87}  {\small($\pm$0.87)}  & {\bf 80.42} {\small($\pm$0.96)}  &  68.45 {\small($\pm$1.38)}   & {\bf 78.53} {\small($\pm$0.76)}   & {\bf 66.98} {\small($\pm$1.52)}   &   74.05 {\small($\pm$0.85)}  \\ 
  \bottomrule[1pt]
  \end{tabular}}  
  \caption{Comparison with SOTA methods on long-tailed CIFAR-10 (CIFAR-10-LT) and long-tailed STL-10 (STL-10-LT)  under $\gamma_l \neq \gamma_u$  setup. The accuracy is reported with standard deviation. } 
  \label{tab:long-tail-rev}
 \end{table}

\subsubsection{Imbalanced Semi-Supervised Image Classification Experimental Results}
Concerning the imbalanced semi-supervised image classification task, a recent SOTA method called distribution-aware semantics-oriented (DASO) framework \cite{oh2022daso} is used to validate our method by simply incorporating the NP model into it \footnote{The details about how the NP model is combined with DASO are shown in the Appendix B.}. 
We followed the experimental settings in the previous work \cite{oh2022daso}, and the results are shown in Tables~\ref{tab:long-tail} and~\ref{tab:long-tail-rev}. Here, we denote "DASO w.~NPs" as the framework that combines NP-Match with DASO for imbalanced semi-supervised image classification. 
We summarize some findings according to the accuracy from Tables~\ref{tab:long-tail} and~\ref{tab:long-tail-rev} as follows. First of all, the NP model does not perform well when both labeled and unlabeled data have the same imabalanced distribution, 
since we can observe minor performance drops in some experimental settings in Table~\ref{tab:long-tail} when DASO is equipped with NPs. However,  the logit adjustment strategy \cite{menon2021long} benefits "DASO w.~NPs" more than the original DASO framework, achieving new SOTA results in all imbalanced settings on CIFAR-10 and  competitive results on CIFAR-100. Second, concerning another MC-dropout-based probabilistic method for SSL, namely, UPS \cite{rizve2021defense},  
we combined it with DASO and then found out that it harms the performance in most imbalanced settings.  
Even though we used two different strategies \cite{menon2021long, lee2021abc} to rectify  the bias towards majority classes, the performance of "DASO w.~UPS" is still worse than "DASO w.~NPs" in most cases, which demonstrates the meliority of the NP model over UPS and MC dropout for imbalanced semi-supervised image classification. Third, the NP model can handle the situation when the imbalanced distribution of labeled data and that of unlabeled data are different, and as shown in Table~\ref{tab:long-tail-rev}, it not only improves the accuracy of DASO in most imbalanced settings, but also outperforms UPS \cite{rizve2021defense} by a healthy margin.

\subsubsection{Multi-Label Semi-Supervised Image Classification Experimental Results}
We adopted a recent SOTA method, named anti-curriculum pseudo-labelling (ACPL) \cite{liu2022acpl}, and integrated our NP model into it \footnote{The details about how the NP model is combined with ACPL are shown in the Appendix B.}, which is denoted as "ACPL-NPs" in Table~\ref{tab:x-ray}.  For a fair comparison with another probabilistic approach (a.k.a.~MC dropout), we combined UPS \cite{rizve2021defense} with ACPL, which is denoted as "ACPL-UPS".

\begin{table*}
\centering 
 
\resizebox{\textwidth}{!}{
\begin{tabular}{@{}c|ccc|ccccc@{}}
 \toprule[1pt]
Method Type  & \multicolumn{3}{c|}{Consistency based} &\multicolumn{5}{c}{Pseudo-labelling}  \\
 \hline  
Method & MT \cite{tarvainen2017mean} & SRC-MT \cite{liu2020semi} & $S^2$MT$S^2$ \cite{liu2021self} & GraphXNet \cite{Aviles} & UPS \cite{rizve2021defense}  & ACPL \cite{liu2022acpl} & ACPL-UPS & ACPL-NPs \\
 \hline 
 Atelectasis  &  75.12 &  75.38 & \underline{77.45} & 72.03 &  76.87 & 77.25 &   77.08     & {\bf 77.54} \\ 
 Cardiomegaly &  87.37 &  \underline{87.70} & 86.84 & {\bf 88.21} &  86.01 & 84.68 &   85.26  & 85.30 \\ 
 Effusion &  80.81 &  81.58 & 82.11 & 79.52 & 81.12 & {\bf 83.13} &  82.27  &   \underline{82.95} \\ 
 Infiltration  &  70.67 & 70.40 & 70.32 & {\bf 71.64} &  71.02   & \underline{71.26} &   71.04  & 71.03 \\ 
 Mass &  77.72 &  78.03 & {\bf 82.82} & 80.29 & 81.59 & 81.68 &  81.79   &  \underline{82.39} \\ 
 Nodule & 73.27 &  73.64 & 75.29 & 71.13 &  {\bf 76.89} &  76.00   &  \underline{76.42}   & 75.85 \\ 
 Pneumonia & 69.17 &  69.27 & 72.66 & {\bf 76.28} &  71.44 & \underline{73.66} &  73.11   & 72.78 \\ 
 Pneumothorax  &  85.63 &  86.12 & \underline{86.78} & 84.24 &  86.02 & 86.08 &   86.12  & {\bf 86.80} \\ 
 Consolidation & 72.51 &  73.11 & 74.21 & 73.24 &  \underline{74.38} & {\bf 74.48} &   {\bf 74.48} & 74.34 \\ 
 Edema &  82.72 &  82.94 & \underline{84.23} & 81.77 &  82.88  & \underline{84.23} & 83.91  & {\bf 84.61} \\ 
Emphysema &  88.16 &  88.98 & 91.55 & 84.89 &  90.17 & \underline{92.47} &   91.99  & {\bf 92.69}\\ 
Fibrosis  &  78.24 &  79.22 & 81.29 & 81.25 &  80.54 & {\bf 81.97} &  \underline{81.55}  & 80.94 \\ 
Pleural Thicken  &  74.43 &  75.63  & \underline{77.02} & 76.23 & 76.13 & 76.92 &   76.53 & {\bf 77.05} \\ 
Hernia  &  \underline{87.74} &  87.27 & 85.64 & 86.89 &  84.12  & 84.49 &  85.10  &  {\bf 89.08} \\ 
 \hline 
 Mean &  78.83 &  79.23 &  80.58 & 79.12 &  79.94 &  \underline{80.59}  &   80.48  & {\bf 80.95} \\ 
  \bottomrule[1pt]
  \end{tabular}}  
  \caption{Class-level AUC testing set results comparison on Chest X-Ray14 based on DenseNet-169 \cite{huang2017densely}, under the 20\% of labelled data setting. {\bf Bold} number denotes the best result per class and \underline{underlined} shows second best result in each class.} 
   \label{tab:x-ray}
 \end{table*} 
 
According to Table~\ref{tab:x-ray}, we can summarize the following observations. First, after involving the NP model, ACPL-NPs slightly improves the mean AUC and also outperforms other SOTA methods, indicating that the NP model and our NP-Match pipeline introduces more reliable pseudo-labels for the label ensemble process. 
Besides, compared against ACPL, ACPL-NPs is able to provide uncertainty estimates, based on which clinicians can further double-check the results in a real-world scenario. 
Second, compared to ACPL, ACPL-UPS performs worse due to the introduced MC dropout. Concerning that the pseudo-labels from the model are selected based on both confidence scores and uncertainty, we empirically consider that the performance drop is caused by the inconsistency between the accuracy and the uncertainty, which is supported by the results in Table~\ref{tab:ablation_uce} to some extend. 
Third, it is cruical to notice the performance gap between ACPL-NPs and ACPL-UPS, which demonstrates that NP-Match is a superior probabilistic model for SSL.

 \subsection{Ablation Studies}
Our ablation studies are conducted on CIFAR-10, CIFAR-100, and STL-10 under the standard semi-supervised image classification setting.  We evaluate our uncertainty-guided skew-geometric JS divergence ($JS^{G_{\alpha_u}}$) as well as its dual form ($JS_*^{G_{\alpha_u}}$), and compare them to the original KL divergence in NPs. In Table~\ref{tab:ablation_jsd}, we see that NP-Match with KL divergence consistently underperforms relative to our proposed $JS^{G_{\alpha_u}}$ and $JS_*^{G_{\alpha_u}}$. This suggests that our uncertainty-guided skew-geometric JS divergence can mitigate the problem caused by low-quality feature representations. Between the two, $JS^{G_{\alpha_u}}$ and $JS_*^{G_{\alpha_u}}$ achieve a comparable performance across the three benchmarks, and thus  we select $JS^{G_{\alpha_u}}$ to replace the original KL divergence in the ELBO (Eq.~(\ref{eq:elbo})) for the comparisons to previous SOTA methods in Section~\ref{sec:main_results}.

\begin{table*}
\centering 
\resizebox{\textwidth}{!}{
\begin{tabular}{@{}cccccccccc@{}}
 \toprule[1pt]
Dataset  & \multicolumn{3}{c}{CIFAR-10} &\multicolumn{3}{c}{CIFAR-100}  & \multicolumn{3}{c}{STL-10}\\
 \hline  
Label Amount & 40 & 250 & 4000 & 400 & 2500 & 10000 & 40 & 250 & 1000\\
 \hline 
 KL &  5.32  {\small($\pm$0.06)} &  5.20 {\small($\pm$0.02)} & 4.36 {\small($\pm$0.03)} & 39.15 {\small($\pm$1.53)} &  26.48  {\small($\pm$0.23)} & 21.51 {\small($\pm$0.17)} &   14.67 {\small($\pm$0.38)}   & 9.92 {\small($\pm$0.24)}  & 6.21  {\small($\pm$0.23)} \\
 \quad $JS_*^{G_{\alpha_u}}$  &  4.93  {\small($\pm$0.02)} &  {\bf 4.87} {\small($\pm$0.03)} & 4.19 {\small($\pm$0.04)} & {\bf 38.67} {\small($\pm$1.29)} &  26.24  {\small($\pm$0.17)} & 21.33 {\small($\pm$0.10)} &   14.45 {\small($\pm$0.55)}   & {\bf 9.48} {\small($\pm$0.28)}  & {\bf 5.47}  {\small($\pm$0.19)} \\
 \quad $JS^{G_{\alpha_u}}$  & {\bf 4.91} {\small($\pm$0.04)} &  4.96 {\small($\pm$0.06)} &  {\bf 4.11} {\small($\pm$0.02)} &  38.91 {\small($\pm$0.99)}  & {\bf 26.03} {\small($\pm$0.26)} & {\bf 21.22} {\small($\pm$0.13)}  & {\bf 14.20} {\small($\pm$0.67)} & 9.51 {\small($\pm$0.37)} &  5.59  {\small($\pm$0.24)}\\ 
  \bottomrule[1pt]
  \end{tabular}}\vspace*{-1ex}
 \caption{Ablation studies of the proposed uncertainty-guided skew-geometric JS divergence and its dual form used by NP-Match.} 
 \label{tab:ablation_jsd} 
 \end{table*} 
 
\section{Summary}
\label{4.5}

In this chapter, we proposed the application of neural processes (NPs) to semi-supervised learning (SSL), designing a new framework called NP-Match, and explored its use in semi-supervised large-scale image classification. 
To our knowledge, this is the first such work. To better adapt NP-Match to the SSL task, we proposed a new divergence term, which we call uncertainty-guided skew-geometric JS divergence, to replace the original  KL divergence in NPs. We demonstrated the effectiveness of NP-Match and the proposed divergence term for SSL in extensive experiments, and also showed that NP-Match could be a good alternative~to MC dropout in SSL. 

\section{Limitations}
\label{4.6} 
There are some limitations regarding NP-Match that need further investigation. First, it is valuable to explore NP-Match on multi-label datasets more thoroughly, as the improvements on the Chest X-Ray dataset are marginal. When each sample has more than one pattern, using a single global latent vector shared by different samples may not be representative enough. 
Secondly, the proposed uncertainty-guided skew-geometric JS divergence defines an intermediate distribution based on the overall uncertainty estimates of both target and context data points, respectively. By encouraging the inferred variational distribution to be close to this intermediate distribution, the parameters in the latent path are less affected by poor feature representations, which effectively improves performance. However, our proposed JS divergence is overly specific, and its computation is somewhat complex concerning the calculation of uncertainty estimates and the analytic formula.
Therefore, extending this term to other, more challenging tasks, such as detection or segmentation, is not trivial.

\chapter{NP-SemiSeg: When Neural Processes meet Semi-Supervised Semantic Segmentation}

In this chapter, I extend NPs to the semi-supervised semantic segmentation task. First, I modify the NP model to process 2D feature maps for segmentation, resulting in a new module named NP-SemiSeg. Subsequently, I incorporate an attention aggregator into NP-SemiSeg, enhancing its ability to fit context data. This chapter is organized as follows. 
An introduction is presented in Section~\ref{5.1}. In Section~\ref{sec:related_work3}, related literature is elaborated. Section~\ref{sec:methodology3} presents our NP-SemiSeg, followed by our experimental details and results in Section~\ref{sec:experiments3}. Finally, we give a conclusion and limitations separately in Section~\ref{5.5} and Section~\ref{5.6}.

\section{Introduction}  
\label{5.1}

Semi-supervised image segmentation has many real-world applications, from medical imaging to autonomous driving systems, where the cost and time to annotate  large-scale training datasets with pixel-level labels is prohibitive. In this field, most recent works \cite{alonso2021semi, chen2021semisupervised, chen2021semi, french2020semi, hu2021semi, ouali2020semi, zhong2021pixel, wang2022semi, guan2022unbiased, liu2022perturbed, kwon2022semi, yang2022st++, zhao2022augmentation} also belong to the deterministic approach. 
By contrast, the probabilistic approach is insufficiently investigated, as researchers have barely explored its application to semi-supervised semantic segmentation for computer vision, and as what has been discussed in previous chapters, most related works focus on medical imaging \cite{sedai2019uncertainty, shi2021inconsistency, yu2019uncertainty, li2020self, wang2021tripled, wang2022rethinking, meyer2021uncertainty, xiang2022fussnet}, in which MC dropout is widely leveraged.


To tackle the limitations of MC dropout, in the preceding chapter, NP-Match \cite{wang2022np} is presented, which adapts neural processes (NPs) to SSL.   
Considering the success of NP-Match and insufficient exploration towards the  probabilistic approach for semi-supervised semantic segmentation, in this part, we investigate the application of NPs on  semi-supervised semantic segmentation, and propose a new model, called NP-SemiSeg. 
In particular, we primarily made two modifications when designing NP-SemiSeg. First, a global latent variable is predicted for each input image, rather than producing a global latent vector shared by different images.\footnote{In NP-Match \protect\cite{wang2022np}, NPs generate a global latent vector shared by all images within a given batch, which follows the pipeline of the original NPs \protect\cite{garnelo2018neural}.} This change is inspired by the fact that different images may have different prior label distributions. 
Hence, it is more reasonable to assume that every image has its own specific prior, and NP-SemiSeg should separately predict a global latent vector for every image, shared by all its pixels. Second, attention mechanisms are additionally introduced
to both the deterministic path and the latent path. In the original NPs \cite{garnelo2018neural}, the information of context points or target points is summarized via a mean aggregator in both paths, and NP-Match also follows this practice. However, the mean aggregator introduces the issue that the decoder of NPs cannot capture relevant information for a given target prediction, as the mean aggregator gives the same weight to each point. Inspired by another model named attentive NPs \cite{kim2019attentive}, 
attention mechanisms are also integrated into NP-SemiSeg to solve this issue.

To validate the effectiveness of NP-SemiSeg, we conducted several experiments on two public benchmarks, namely,  PASCAL VOC 2012
and Cityscapes, with diverse SSL settings, and the results show two merits of NP-SemiSeg.  
First, NP-SemiSeg is versatile and flexible, because it can be integrated into different segmentation frameworks, such as CPS \cite{chen2021semi} or U$^2$PL \cite{wang2022semi}. Equipped with NP-SemiSeg, those frameworks are turned into probabilistic models, which are able to make predictions and quantify the uncertainty for  input samples. Second, compared to the widely used MC-dropout-based segmentation models, the segmentation models with NP-SemiSeg are faster in terms of uncertainty quantification and are able to give higher-quality uncertainty estimates with less performance degradation, indicating that NP-SemiSeg can be a good alternative probabilistic method to MC dropout.

It should be noted that the our principal objective is not to introduce a new segmentation approach that surpasses all state-of-the-art methods. Rather, the aim is to present a novel probabilistic model for semi-supervised semantic segmentation, capable of delivering both a good performance and reliable uncertainty estimates. Summarizing, the main contributions are:
\begin{itemize}[leftmargin=*, itemsep=0.5pt]
\item We adjust NPs to semi-supervised semantic segmentation, and propose a new probabilistic model, named NP-SemiSeg, which is flexible and can be combined with different existing segmentation frameworks for making predictions and estimating uncertainty. 

\item We integrate an attention aggregator into NP-SemiSeg, which assigns higher weights to the information that is more relevant to target data, enhancing the performance of NP-SemiSeg.

\item Compared to MC-dropout-based segmentation models, NP-SemiSeg not only performs better in terms of accuracy, but also runs faster regarding uncertainty estimation, showing its potential to be a new probabilistic model for semi-supervised semantic segmentation.
 
\end{itemize}

\section{Related Works}
\label{sec:related_work3}

In this section, we review related works, including SSL for image classification and semi-supervised semantic segmentation.

{\bf SSL for Image Classification.} In the past few years, many methods have been proposed for semi-supervised image classification, which provide insights and research directions for semi-supervised semantic segmentation. The most prevalent method is FixMatch \cite{sohn2020fixmatch}. During training, it produces pseudo-labels for weakly-augmented unlabeled data based on a preset confidence threshold, and the pseudo-labels are used as the ground-truth for their strongly augmented version to train the whole framework. FixMatch \cite{sohn2020fixmatch} thereafter inspired a series of promising methods \cite{li2021comatch, rizve2021defense, zhang2021flexmatch, nassar2021all, pham2021meta, hu2021simple}. 
For example, Le {\it et al.} \cite{li2021comatch} incorporate contrastive learning through additionally designing the projection head that generates low-dimensional embeddings for samples. The low-dimensional embeddings  with similar pseudo-labels are encouraged to be close, which improves the quality of pseudo-labels. Zhang {\it et al.} \cite{zhang2021flexmatch} use dynamic confidence thresholds 
that are adjusted based on the model's learning status of each class, rather than the fixed preset confidence threshold. A more relevant method, named uncertainty-aware pseudo-label selection (UPS) framework, was proposed by Rizve {\it et al.} \cite{rizve2021defense}. This framework can be regarded as a probabilistic approach, as it applies MC dropout to obtain uncertainty estimates, based on which unreliable pseudo-labels are filtered out. Due to the weaknesses of MC dropout mentioned above, Wang {\it et al.} \cite{wang2022np} try to explore a new alternative probabilistic model, i.e., NPs, for semi-supervised image classification, and propose a new method called NP-Match, which not only shows a promising accuracy on several public benchmarks, but also alleviates the problem of MC dropout. These results encourage us to further investigate the application of NPs on semi-supervised semantic segmentation.

{\bf Semi-supervised Semantic Segmentation.} Most methods can be classified into two training paradigms, namely, consistency-training \cite{french2020semi, zhou2021c3, ouali2020semi, zhong2021pixel, liu2022perturbed, ke2019dual} and self-training \cite{alonso2021semi, chen2021semisupervised, hu2021semi, wang2022semi, guan2022unbiased, kwon2022semi, yang2022st++, zou2020pseudoseg, zhao2022augmentation}. 

The consistency-training methods aim to maintain the consistency among the segmentation results of different perturbations of the same unlabeled sample. For example, Ouali {\it et al.} \cite{ouali2020semi} propose a cross-consistency training (CCT) method, and it contains a main decoder and several auxiliary decoders, which share the same encoder. For the unlabeled examples, a consistency between the main decoder’s outputs and the auxiliary outputs is maintained, over different kinds of perturbations leveraged
to the inputs of the auxiliary decoders. Zhong {\it et al.} \cite{zhong2021pixel} design a new framework, named PC$^2$Seg, which takes advantage of both the pixel-contrastive property and the consistency property during training, and their combination further enhances the performance.
Considering the potential inaccurate training signal caused by perturbations, 
Liu {\it et al.} \cite{liu2022perturbed} introduce an additional teacher model, a stricter confidence-weighted cross-entropy loss, and a new type of feature perturbation to improve  consistency learning.  

Self-training methods assign pixel-wise pseudo-labels to unlabeled data, and re-train the segmentation networks. For instance,  PseudoSeg \cite{zou2020pseudoseg} utilizes the predictions of unlabeled data as the labels to re-train the whole framework. To obtain accurate pseudo-labels, a calibrated fusion module is incorporated, which fuses both the outputs of the decoder and the refined class activation map (CAM). The success of self-supervised learning motivates Alonso {\it et al.} \cite{alonso2021semi} to integrate the pixel-level contrastive learning scheme into their framework, which aims at enforcing the feature vector of a target pixel to be similar to the same-class features from the memory bank. Recently, Wang {\it et al.} \cite{wang2022semi} have discovered that some pixels may never be learned in the entire self-training process, due to their low confidence scores. Then, they propose a new framework,  called U$^2$PL, which reconsiders those pixels as negative samples for training. Zhao {\it et al.} \cite{zhao2022augmentation} reconsider the data augmentation techniques used in the self-training process, and they design a new highly random intensity-based augmentation method and an adaptive cutmix-based augmentation method to enhance the performance. 

All above methods do not involve any probabilistic model, and it is only valued in medical imaging \cite{sedai2019uncertainty, shi2021inconsistency, yu2019uncertainty, li2020self, wang2021tripled, wang2022rethinking, meyer2021uncertainty, xiang2022fussnet}, where most methods rely on MC dropout for approximating BNNs and estimating uncertainty. In a nutshell, those methods usually leverage uncertainty maps given by MC dropout to refine pseudo-labels for unlabeled data, thereby boosting the capability of their models.

\section{NP-SemiSeg}
\label{sec:methodology3}
 
\subsection{NPs for semi-supervised semantic segmentation}

Semantic segmentation can be treated as a pixel-wise classification problem, and therefore, $p(y_{1:n}|\phi(x_{1:n}, z), x_{1:n})$ in Eq.~(\ref{eq:joint2}) can be changed to the categorical distribution (denoted as $\mathcal{C}$). 
Specifically, a weight matrix ($\mathcal{W}$) and a softmax function ($\Phi$) can be sequentially applied to the feature presentation of every pixel from the decoder $\phi(\cdot)$, outputting a probability vector that can parameterize  $\mathcal{C}$. 
Furthermore, different images can have distinct prior label distributions, as some objects cannot appear in the same image. For example, if an image captures the main road of a city, fish will not appear, whose prior should be zero. But if the image records the creatures in the sea, the prior of fish is close to one.  
Because of this, employing a global latent variable for different images appears to be unreasonable, which is further corroborated by our experimental results on multi-label semi-supervised image classification presented in Section~\ref{sec:main_results}. 
Therefore, we instead use a latent variable per image. This can be viewed as giving each image its own prior. 
Thus, we rewrite $p(y_{1:n}|\phi(x_{1:n}, z), x_{1:n})$ as follows:
\begin{equation}
\label{eq:likelihood2}
p(y_{1:n}|\phi(x_{1:n}, z_{1:n}), x_{1:n}) =  \mathcal{C}(\Phi(\mathcal{W}\phi(x_{1:n}, z_{1:n}))),
\end{equation}
where the decoder $\phi(\cdot)$ can be learned through amortised variational inference. Specifically, as for a finite sequence with length $n$, we assume  $m$ context data ($x_{1:m}$) and $r$ target data ($x_{m+1:\ m+r}$) in it, i.e., $m+r=n$. We also assume a variational distribution over latent variables, and the ELBO is given by  (with proof in the Appendix C):
\begin{equation} 
\begin{aligned}
\label{eq:elbo1}
&log\ p(y_{1:n}|x_{1:n}) \ge \\ 
& \mathbb{E}_{q(z_{m+1:\ m+r}|x_{m+1:\ m+r}, y_{m+1:\ m+r})}\Big[\sum^{m+r}_{i=m+1}log\ p(y_i|z_i, x_i) - \\
&log\ \frac{q(z_{m+1:\ m+r}|x_{m+1:\ m+r}, y_{m+1:\ m+r})}{q(z_{m+1:\ m+r}|x_{1:m}, y_{1:m})}\Big] + log\ p(y_{1:m}|x_{1:m}).
\end{aligned}
\end{equation}
Then, one can maximize the ELBO to learn the NP model. During training, we follow the setting of NP-Match \cite{wang2022np} which treats only labeled data as context data and treats either labeled or unlabeled data as target data.

\begin{figure*}[t]
\centering
\includegraphics[width=\linewidth]{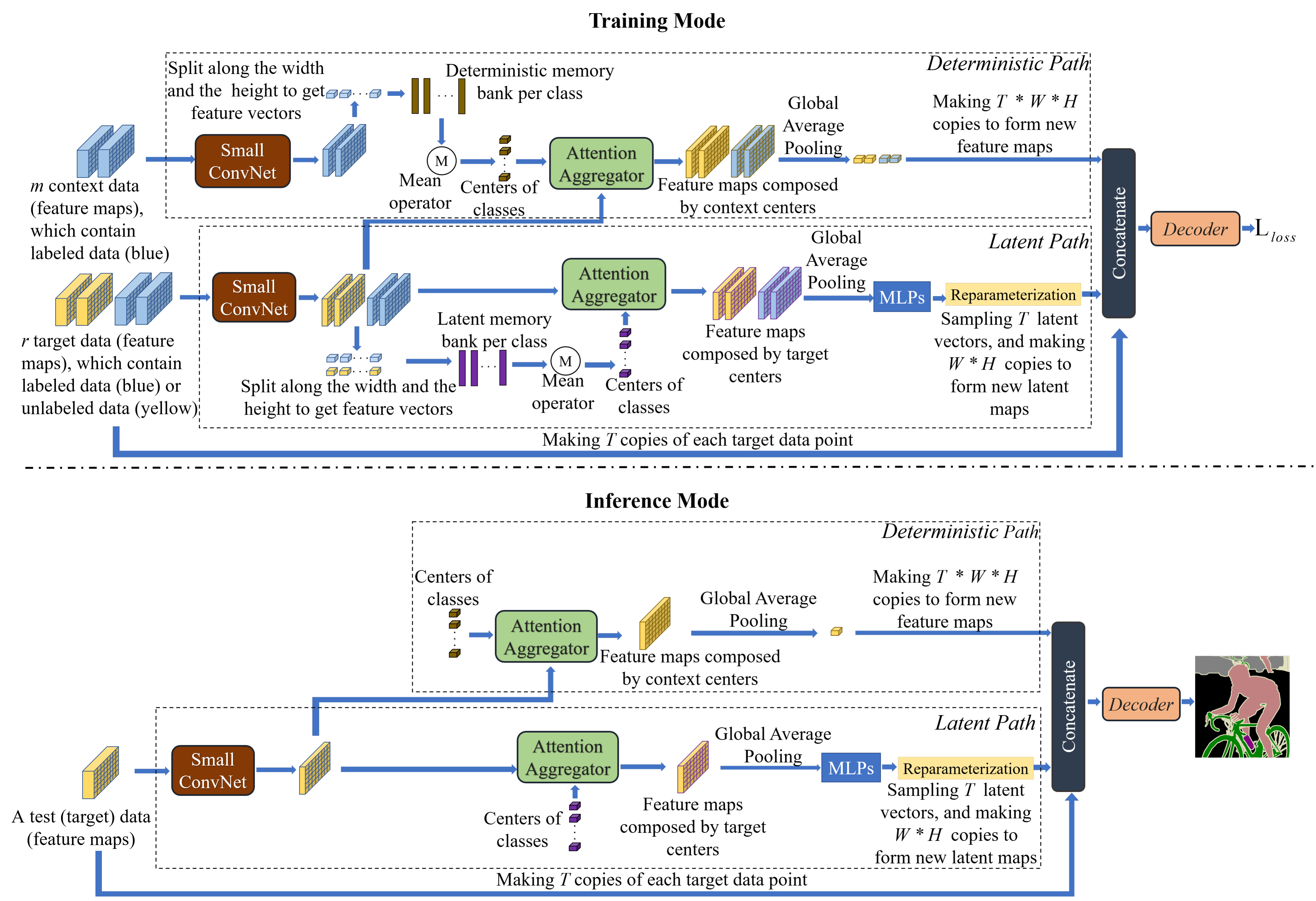}
\vspace{-3.5ex}
\caption{
Overview of NP-SemiSeg. Both the small ConvNet and the attention aggregator are shared by the deterministic path and the latent path. $T$, $W$, and $H$ represent the number of sampled latent vectors, and the width and height of the input feature maps, respectively.}
\vspace{-1ex}
\label{fig:npsemiseg} 
\end{figure*}

\subsection{NP-SemiSeg Pipeline} 
We formulate NP-SemiSeg in a modular fashion, so that it can  directly replace the classification layer of any segmentation pipeline without changing other modules in the pipeline, to output predictions with uncertainty estimates. 
As a result, NP-SemiSeg is flexible and can be used for different segmentation frameworks. To achieve this goal, the input of NP-SemiSeg should be feature maps,\footnote{In general, most segmentation frameworks are based on DeepLab \cite{chen2017deeplab}, where a classifier acts on the final output feature maps from the decoder to predict for every location.} which is consistent with the input of a classifier in other segmentation frameworks. To make explanations clearer, we only focus on NP-SemiSeg itself.

The overall pipeline of NP-SemiSeg is shown in Figure~\ref{fig:npsemiseg}, where we represent the context and target data as generic feature maps, which could be obtained from any semantic segmentation pipeline such as U$^2$PL \cite{wang2022semi} and AugSeg \cite{zhao2022augmentation}. NP-SemiSeg has a training mode and an inference mode. The former aims to calculate loss functions with real labels or pseudo-labels during training, while the latter makes predictions for unlabeled data during training or test data during testing.
In what follows, we describe these two modes:

{\bf Training mode.} 
Given a batch of labeled data and a batch of unlabeled data, NP-SemiSeg is initially switched to inference mode, and it makes predictions for the unlabeled data. Those predictions are regarded as pseudo-labels for unlabeled data by taking the class with the highest probability. Then, NP-SemiSeg  turns to training mode, and it duplicates the labeled samples and treats them as context data. 
Subsequently, the context data are passed through a deterministic path, which aims to obtain order-invariant context representations, and the target data are passed through a latent path,  which aims to produce latent variables. The outputs from both paths are finally concatenated and then passed through a decoder before the loss is computed. Below, we provide details for the latent path and the deterministic path.

As for the latent path, target data are processed by a small ConvNet\footnote{The small ConvNet is mainly composed by $1\times1$ convolutions, and its outputs have the same spatial size as its inputs.} at first for dimensionality reduction, whose outputs are transformed feature maps with a low channel dimension. The transformed feature maps are further split along the width ($W$) and the height ($H$), resulting in feature vectors. Based on the number of classes, a set of latent memory banks have been initialized, each of which is assigned to a category. Those feature vectors are passed to the latent memory banks according to their real or pseudo labels.\footnote{Note that $q(z_*|x_{m+1:m+r}, y_{m+1:m+r})$ is conditioned on both data and labels, and in NP-SemiSeg, the condition on labels is implicitly implemented, i.e., how data are stored in memory banks is  determined by the labels.} Then, a mean operator is used for each memory bank, and we can obtain a center for each class. 
Those centers and the target transformed feature maps are input to an attention aggregator, whose outputs are feature maps composed by target centers. 
Specifically, 
the feature vector of each location in such centers-based feature maps is the weighted summation of the centers, which intends to represent every location by the most relevant features from the memory. 
Thereafter, the global average pooling and MLPs are used to produce a mean vector and a variance vector for each target data point, followed by a reparameterization trick to get $T$ latent vectors whose dimension is $D_t$. Finally, those latent vectors are copied for $W \times H$ times, thereby forming latent maps for each target data point with size $T \times D_t \times W \times H$. 

As for the deterministic path, context data are processed in the same way as the target data, until we obtain the context centers for classes. Then, the context centers as well as the  target transformed feature maps are fed to the attention aggregator, in order to get the feature maps composed by the context centers, which are further processed by global average pooling, leading to an order-invariant context representation with dimension $D_c$ for each target data point. Finally, the order-invariant context representation is copied for $T \times W \times H$ times, thereby forming context maps for each target data point with size $T \times D_c \times W \times H$. 

After the latent maps and the context maps are obtained for each target data point, they are concatenated with the original feature maps of the target data whose size is $T \times D \times W \times H$, and the concatenated feature maps will have the size $T \times (D + D_t + D_c) \times W \times H$, based on which a decoder $\phi(\cdot)$ makes pixel-wise predictions. The final prediction for each target data point can be obtained by averaging the $T$ prediction maps, and the uncertainty map is computed as the entropy of the average prediction \cite{kendall2017uncertainties}. For saving space, only those centers are stored for inference after training, instead of saving those memory banks.

{\bf  Inference mode.} As for a set of test data, they are treated as target data and are first processed by the small ConvNet. Its outputs, the target centers, and the context centers are taken as inputs to the attention aggregator to acquire the feature maps composed by centers. Subsequently, the remaining steps are the same as in the training mode to generate concatenated feature maps where the decoder $\phi(\cdot)$ acts on to make predictions, along with their associated uncertainty estimates.

\subsection{Attention Aggregator}
To predict a target data point, it is beneficial to gather relevant information from memory banks, as the centers close to the target provide similar representations. To achieve this, an attention aggregator is required, whose role is to produce centers-based feature maps based on the distance between query feature maps and a set of centers.   We denote the input feature maps and the input centers as $\mathcal{M}$ and $\mathbf{C}$, respectively.  The output $\mathcal{M}_\mathbf{C}$ is calculated as follows:
\begin{equation} 
\begin{aligned}
\label{eq:atten}
 \mathcal{M}_\mathbf{C}[i, j] = \sum_l \frac{e^{-\Theta(\mathcal{M}[i, j], \mathbf{C}[l]))}}{\sum_k e^{-\Theta(\mathcal{M}[i, j], \mathbf{C}[k]))}} \mathbf{C}[l],
\end{aligned}
\end{equation}
where $i$ and $j$ denote the index of feature maps along width and height. Both $l$ and $k$ denote the index of centers. $\Theta$ is defined as Euclidean distance over two vectors. In summary, the attention aggregator uses $\Theta$ to calculate the distance between the feature vector $\mathcal{M}[i, j]$ at the location ($i$, $j$) and every center, and all distances are further used to calculate weights through the softmax function for centers. Then, the output feature at location [$i$, $j$], namely,  $\mathcal{M}_\mathbf{C}[i, j]$, is the weighted combination of those centers. 
Similarly to ANPs \cite{kim2019attentive}, by using an attention aggregator, only the relevant information from the latent path and the deterministic path is involved for making predictions, thereby improving the model's performance. 

\subsection{Loss Functions} 
The loss function for NP-SemiSeg is derived from the ELBO (Eq.~(\ref{eq:elbo1})). In particular, the first term can be achieved by pixel-wise cross entropy loss $L_c$ for both labeled and unlabeled data, which is widely used in different segmentation frameworks. The second term is the KL divergence between $q(z_{m+1:\ m+r}|x_{m+1:\ m+r}, y_{m+1:\ m+r})$ and $q(z_{m+1:\ m+r}|x_{1:m}, y_{1:m})$. Due to the i.i.d assumption, those $z_*$ are conditionally independent, and thus they can be calculated independently. We assume that the variational distribution follows a multivariate Gaussian with independent components, and for each target sample, the KL divergence term can be analytically written as:
\begin{equation} 
\begin{aligned}
\label{eq:kl}
L_{kl} =& 0.5 \times [\sum_{D_t} log\frac{\sigma_c^2}{\sigma_t^2} + \sum_{D_t} \frac{\sigma_t^2}{\sigma_c^2}  -  D_t +  (m_c-m_t)diag({\sigma_c^{-2}})(m_c-m_t)^T],
\end{aligned}
\end{equation}
where $diag(\cdot)$ receives a vector and converts it into a diagonal matrix. $m_c$ and $m_t$ denote the mean vector of $q(z_*|x_{1:m}, y_{1:m})$ and  $q(z_*| y_{m+1:\ m+r})$, respectively. Similarly, $\sigma_c^2$ and $\sigma_t^2$ denote the variance vector of $q(z_*|x_{1:m}, y_{1:m})$ and  $q(z_*| y_{m+1:\ m+r})$, respectively. 
The third term is a conditional distribution over the context data, but it is ignored in our loss function, as its maximization has been implicitly implemented by the attention aggregator, i.e., matching the transformed feature maps to the centers (classes) according to their distances. 
The overall loss function for NP-SemiSeg can be written as:
\begin{equation} 
\begin{aligned}
\label{eq:all}
L_{loss} = L_c + \lambda_{kl} L_{kl},
\end{aligned}
\end{equation}
where $\lambda_{kl}$ is the coefficient. When NP-SemiSeg is incorporated into different segmentation frameworks, $L_{loss}$ can be naturally incorporated into their loss functions for end-to-end training.

\section{Experiments}
\label{sec:experiments3}

In this section, we present our  experimental results. To save space, the implementation details are given in the Appendix C.

\subsection{Datasets}
We tested our models on two public segmentation benchmarks, namely, Cityscapes \cite{cordts2016cityscapes} and PASCAL VOC 2012 \cite{everingham2010pascal}.  Cityscapes is an urban scene understanding dataset containing 2, 975 training images with fine-annotated masks and 500 validation images. We followed previous works \cite{wang2022semi, zhao2022augmentation, chen2021semi} to use the sliding
evaluation for fair comparisons. PASCAL VOC 2012 is a standard semantic segmentation dataset that has 20 semantic classes and 1 background class. There are 1,464  and 1,449 images in the training set and the validation set, respectively. Following \cite{wang2022semi, zhao2022augmentation, chen2021semi}, we used coarsely-labeled 9,118 images from the Segmentation Boundary dataset (SBD) \cite{hariharan2011semantic}
as additional training data, and  we also evaluated our model on the {\it classic} set and the {\it blender} set. As in previous works \cite{wang2022semi, zhao2022augmentation, chen2021semi}, the center-crops of images were used for evaluation.

\begin{table}
\centering 
\resizebox{0.65\textwidth}{!}{
\begin{tabular}{@{}ccccc@{}}
 \toprule[1pt]
Method & 1/16 (92) &  1/8 (183)   & 1/4 (366)  & 1/2 (732)  \\
 \hline 
  MT   & 48.37  & 58.44  &  65.49  &  68.92 \\ 
  PS-MT & 63.32  & 67.78   &  74.68  &  76.54 \\ 
  U$^2$PL  &  62.13  &  68.11  &  73.22  &  75.60  \\ 
  AugSeg  & 64.22 &  72.17   &  76.17  &   77.40 \\ 
 \hline
 MT   w/ MC dropout   & 47.78  & 57.02   &  64.82  &  67.79 \\ 
 PS-MT w/ MC dropout  & 62.09  & 66.46   &  73.11 &  74.30 \\ 
 U$^2$PL w/ MC dropout & 59.17  & 66.89  & 72.16 &  74.19 \\ 
 AugSeg w/ MC dropout  & 62.78  & 69.87   & 74.76  &  76.13 \\
\hline
 MT   w/ NP-SemiSeg   &  49.02  &  58.91   &  65.27  &  69.34  \\ 
 PS-MT w/ NP-SemiSeg  & 63.76   &   68.17   &  74.93  &   76.33   \\ 
 U$^2$PL w/ NP-SemiSeg  &  59.45  &  68.73   &  74.16   &  75.77 \\ 
 AugSeg w/ NP-SemiSeg  & 65.78  &  72.38  & 75.77 &  77.40   \\ 
  \bottomrule[1pt]
  \end{tabular}}\vspace*{-1ex}
 \caption{The mean IoU of different frameworks using ResNet-50 with either MC dropout or NP-SemiSeg on the  {\it classic} PASCAL VOC 2012 validation set under different partition protocols.}  
 \label{tab:voc_classic}
 \end{table}

\begin{table}[t]
\centering 
\resizebox{0.65\textwidth}{!}{
\begin{tabular}{@{}ccccc@{}}
 \toprule[1pt]
Method & 1/16 (662) & 1/8 (1323)  & 1/4 (2646) & 1/2 (5291)   \\
 \hline 
 MT  & 66.77  & 70.78    &  73.22  & 75.29 \\ 
 PS-MT  & 72.83   & 75.70   &  76.43 &  77.88 \\ 
 U$^2$PL  & 74.74  &  77.44  &  77.51   &  78.62  \\ 
 AugSeg  &  77.28   & 78.27  &   78.24  &  79.02 \\ 
 \hline
 MT   w/ MC dropout   &  65.46 &  69.29 &  72.39  &  74.67 \\ 
 PS-MT w/ MC dropout  & 71.28  &   74.03 &  74.97 &  75.97
 \\\ 
 U$^2$PL w/ MC dropout & 73.79  & 76.23  &  76.56  &  76.41 \\ 
 AugSeg w/ MC dropout  & 76.42  &  76.87    &  77.02   &  77.56   \\
\hline
 MT   w/ NP-SemiSeg   &  66.93   &  71.25  &  73.10  &  75.31 \\ 
 PS-MT w/ NP-SemiSeg  &  73.44  &  76.58  & 76.74  &  76.82 \\ 
 U$^2$PL w/ NP-SemiSeg  & 75.59  &  77.77 & 77.78  & 77.23  \\ 
 AugSeg w/ NP-SemiSeg  & 77.00  & 78.68  &  78.69 &  79.03 \\ 
  \bottomrule[1pt]
  \end{tabular}}\vspace*{-1ex}
 \caption{The mean IoU of different frameworks using ResNet-50 with either MC dropout or NP-SemiSeg on the {\it blender} PASCAL VOC 2012 validation set under different partition protocols.}  
 \label{tab:voc_blender} 
 \end{table}

\subsection{Main Results}
In the following, we report the main experimental results on the mean of Intersection over Union (mIoU),  the Patch Accuracy vs. Patch Uncertainty (PAvPU) metric \cite{mukhoti2018evaluating}, and the running time of NP-SemiSeg over the two benchmarks.

First, because of the flexibility of NP-SemiSeg, we integrated it into different segmentation frameworks to show its performance. We chose four frameworks, namely, MT \cite{tarvainen2017mean}, PS-MT \cite{liu2022perturbed}, U$^2$PL \cite{wang2022semi}, and AugSeg \cite{zhao2022augmentation}. The first two frameworks are classified as the consistency-training method, while the rest belongs to the self-training method. Since MC dropout is the mainstream probabilistic approach in SSL, we also evaluated it by applying it to the four frameworks, and it is inserted after every activation layer in their decoders.
From Tables~\ref{tab:voc_classic}, \ref{tab:voc_blender}, and \ref{tab:city}, we have two findings. First, on PASCAL VOC 2012, NP-SemiSeg can help to further improve the mIoU in most cases. In  contrast, MC dropout leads to a poor performance, and it is outperformed by NP-SemiSeg with a healthy margin. Second, on Cityscapes, though NP-SemiSeg only achieves comparable results, it still performs clearly better than MC dropout. 
Thus, compared to MC dropout, NP-SemiSeg is a more favorable choice for semi-supervised semantic segmentation, as it does not cause a serious performance degradation. 
In the other experiments, we fixed a single framework, i.e., U$^2$PL \cite{wang2022semi},  to further explore NP-SemiSeg.

\begin{table}
\centering 
\resizebox{0.65\textwidth}{!}{
\begin{tabular}{@{}ccccc@{}}
 \toprule[1pt]
Method & 1/16 (186) &  1/8 (372)     & 1/4 (744) &  1/2 (1488)  \\
 \hline 
 MT  & 66.14  & 72.03  &  74.47   & 77.43 \\ 
 PS-MT  & 70.12  &  74.49   &   76.12   &  77.64 \\ 
 U$^2$PL  &  69.03   & 73.02  &   76.31  &   78.64 \\ 
 AugSeg  & 73.73   & 76.49   &  78.76  & 79.33 \\ 
 \hline
 MT   w/ MC dropout   &  65.25 &  71.09 & 72.48  &  74.96 \\ 
 PS-MT w/ MC dropout  &  68.83 &  73.11  & 75.25  &  75.47  \\ 
 U$^2$PL w/ MC dropout & 67.89 &  72.13 & 75.11  &  75.85 \\ 
 AugSeg w/ MC dropout  & 72.28  &  75.84  &  77.69 & 78.04   \\
\hline
 MT   w/ NP-SemiSeg   &  66.20  & 72.14  &   73.89 &  76.29  \\ 
 PS-MT w/ NP-SemiSeg  & 70.27  &  74.67  &  76.14 & 76.93  \\ 
 U$^2$PL w/ NP-SemiSeg  & 69.10  &  73.04   &  75.79  &  75.75  \\ 
 AugSeg w/ NP-SemiSeg  &  73.01 & 77.10  & 78.82   &  78.77 \\ 
  \bottomrule[1pt]
   \end{tabular}}\vspace*{-1ex}
 \caption{The mean IoU of different frameworks using ResNet-50 with either MC dropout or NP-SemiSeg on the Cityscapes validation set under different partition protocols. }  
 \label{tab:city} 
 \end{table}

\begin{table}[t]
\centering 
\resizebox{0.6\textwidth}{!}{
\begin{tabular}{@{}c|c|c|c@{}}
 \toprule[1pt]
Dataset & Label Amount & MC Dropout    & NP-SemiSeg \\
 \hline 
  \multirow{4}{*}{Cityscapes}   & 1/16 (186)  & 82.89   &   84.05  \\  
   & 1/8 (372)  &  82.84   &  83.97    \\ 
   & 1/4 (744)  &   82.78  &  84.55   \\ 
   & 1/2 (1488)  &  82.92 &  84.61   \\ 
  \hline 
  \multirow{4}{*}{VOC (\it classic)}   & 1/16 (92)  & 85.79  & 86.87   \\  
   & 1/8 (183)  & 86.42   &  87.98   \\ 
   & 1/4 (366)  &  87.05  &  88.74  \\ 
   & 1/2 (732)  &  87.64  &  89.69  \\ 
   \hline 
  \multirow{4}{*}{VOC (\it blender)}   & 1/16 (662)  &  88.04 &  89.62 \\  
   & 1/8 (1323)  & 87.96   &  89.87\\ 
   & 1/4 (2646)  & 88.18    &   89.99   \\ 
   & 1/2 (5291)  & 88.42  &  89.34  \\ 
  \bottomrule[1pt]
  \end{tabular}}\vspace*{-1ex}
 \caption{The PAvPU of U$^2$PL \cite{wang2022semi} using ResNet-50 with either MC dropout or NP-SemiSeg on different datasets. }  
 \label{tab:pavpu1} 
 \end{table}

Second, we compare the PAvPU of NP-SemiSeg with that of MC dropout in Table~\ref{tab:pavpu1} for the purpose of evaluating their uncertainty estimation. 
Under the same label amount setting for each dataset, NP-SemiSeg achieves a higher PAvPU metric than MC dropout, showing that the former
can output more reliable uncertainty estimates.
Therefore, it is more suitable than MC dropout for semi-supervised semantic segmentation in terms of uncertainty quantification.

 \begin{figure}[t]
\centering
\includegraphics[width=0.75\linewidth]{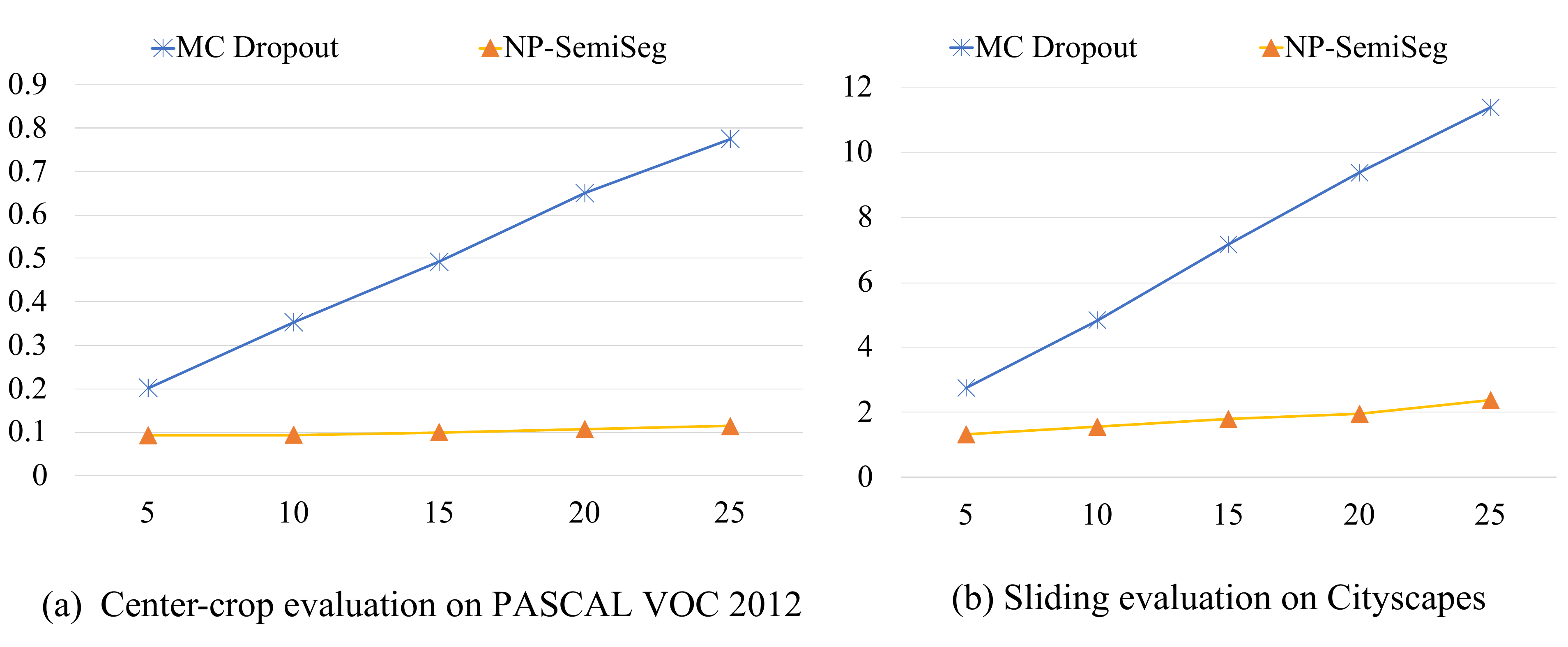}
\vspace{-1ex}
\caption{Time consumption of estimating uncertainty for U$^2$PL \cite{wang2022semi} with MC dropout and NP-SemiSeg. The horizontal axis refers to the number of predictions used for the uncertainty quantification, and the vertical axis indicates the time consumption (sec).} 
\label{fig:speed1}
\end{figure} 

Finally, we compare the running time of NP-SemiSeg and MC dropout for quantifying uncertainty, under two evaluation strategies, namely, the center-crop evaluation on PASCAL VOC 2012 and the sliding evaluation on Cityscapes. Note that the encoder of U$^2$PL in our experiments is a ResNet-50 \cite{he2016deep} pretrained on the ImageNet dataset \cite{deng2009imagenet}, and therefore MC dropout is only inserted into the decoder, and only the decoder performs $T$ times of feedforward passes for saving time. From Figure~\ref{fig:speed1}, we have the following observations. First, when the number of predictions ($T$) increases, the time cost of MC dropout also rises accordingly, and the gap between NP-SemiSeg and MC dropout gradually becomes  significant. Second, if the sliding evaluation is used, the time consumption of MC dropout is hardly acceptable, as MC dropout requires more numbers of feedforward passes than NP-SemiSeg for this strategy. For instance, to evaluate a large image, we need to move the sliding window for $r$ strides in total, and in this case, MC dropout needs $T \times r$ feedforward passes, while NP-SemiSeg only needs $r$ feedforward passes. These observations demonstrate that NP-SemiSeg is computationally more efficient than MC dropout for semi-supervised semantic segmentation.

\begin{table}
\centering 
\resizebox{0.6\textwidth}{!}{
\begin{tabular}{@{}c|c|c|c@{}}
 \toprule[1pt]
Dataset & Label Amount &  w/o Attention   & w/ Attention \\
 \hline 
  \multirow{4}{*}{Cityscapes}   & 1/16 (186)  & 67.86  &   69.10  \\  
   & 1/8 (372)  &  72.44  &  73.04    \\ 
   & 1/4 (744)  &  75.33  &  75.79  \\ 
   & 1/2 (1488)  &   75.45  &  75.75   \\ 
  \hline 
  \multirow{4}{*}{VOC (\it classic)}   & 1/16 (92)  &  58.52   & 59.45   \\  
   & 1/8 (183)  &  68.12  &  68.73    \\ 
   & 1/4 (366)  &   73.72 &   74.16   \\ 
   & 1/2 (732)  &  75.64  &   75.77  \\ 
   \hline 
  \multirow{4}{*}{VOC (\it blender)}   & 1/16 (662)  &   74.80  &   75.59  \\  
   & 1/8 (1323)  & 77.26  &  77.77  \\ 
   & 1/4 (2646)  & 77.38  &   77.78  \\ 
   & 1/2 (5291)  & 76.91  &  77.23   \\ 
  \bottomrule[1pt]
  \end{tabular}}\vspace*{-1ex}
 \caption{Ablation studies of attention aggregation on different datasets. The results are all based on U$^2$PL \cite{wang2022semi} using ResNet-50, and mean IoU is reported.}  
 \label{tab:ablate1} 
 \end{table}

 \begin{table}
\centering 
\resizebox{0.6\textwidth}{!}{
\begin{tabular}{@{}c|c|c|c@{}}
 \toprule[1pt]
Dataset & Label Amount &  w/o Attention   & w/ Attention \\
 \hline 
  \multirow{4}{*}{Cityscapes}   & 1/16 (186)  &  83.46  &  84.05  \\  
   & 1/8 (372)  &  83.61 &  83.97   \\ 
   & 1/4 (744)  &  84.32  &    84.55  \\ 
   & 1/2 (1488)  &  84.60  &   84.61  \\ 
  \hline 
  \multirow{4}{*}{VOC (\it classic)}   & 1/16 (92)  &  86.22  &   86.87   \\  
   & 1/8 (183)  &  87.54  &  87.98   \\ 
   & 1/4 (366)  &  88.57 &   88.74  \\ 
   & 1/2 (732)  &  89.53 &  89.69     \\ 
   \hline 
  \multirow{4}{*}{VOC (\it blender)}  
   & 1/16 (662)  &  89.46 &  89.62 \\  
   & 1/8 (1323)  &  89.53  &  89.87 \\ 
   & 1/4 (2646)  &  89.57  &   89.99   \\ 
   & 1/2 (5291)  &  89.35  &  89.34  \\ 
  \bottomrule[1pt]
  \end{tabular}}\vspace*{-1ex}
 \caption{Ablation studies of attention aggregation on different datasets. The results are all based on U$^2$PL \cite{wang2022semi} using ResNet-50, and PAvPU is reported.}  
 \label{tab:ablate2} 
 \end{table}

\subsection{Ablation Studies}
We conducted ablation studies of the attention aggregator on two public benchmarks, which are shown in Tables~\ref{tab:ablate1} and \ref{tab:ablate2}. For the experiments without using the attention aggregator, we followed the previous work \cite{wang2022np} to use a mean aggregator for assembling the information instead. 

The results show the effectiveness of the attention aggregator. In particular, when it is removed, we can observe that the mIoU decreases in Table~\ref{tab:ablate1}. From the perspective of uncertainty quantification, we  also see the gap regarding PAvPU between NP-SemiSeg with and without attention aggregator.  
Even though the performance gains are marginal, we believe that our attention aggregator contributes to enhancing the performance, and these improvements are not just statistically random occurrences. This belief is supported by the consistent improvements we have observed across all datasets under different label settings. If the attention aggregator were not effective, such consistent enhancements would not be present. 
Our results support the significance of the attention aggregator, which involves relevant information from the memory banks to infer the latent maps and the context maps.

\section{Visualization Results}

We visualize some prediction results and uncertainty maps given by NP-SemiSeg on both PASCAL VOC 2012 ({\it blender}) and Cityscapes. For the uncertainty maps, we calculate pixel-wise predictive entropy, and represent the uncertainty with gray images. 
Each uncertainty map uses pixel values, ranging from black to white, to denote the levels of uncertainty, starting from low to high.

According to the visualization results, NP-SemiSeg can provide a good quality of uncertainty estimates. In general,  it can give a high uncertainty for the pixels that are wrongly predicted. Furthermore, the boundary of an object is more likely to be misclassfied, and therefore,  NP-SemiSeg also gives  high uncertainties to boundaries. Based on this information, one can make decisions or further improve the results in a real-world scenario.

\begin{figure} 
\centering
\includegraphics[width=\linewidth]{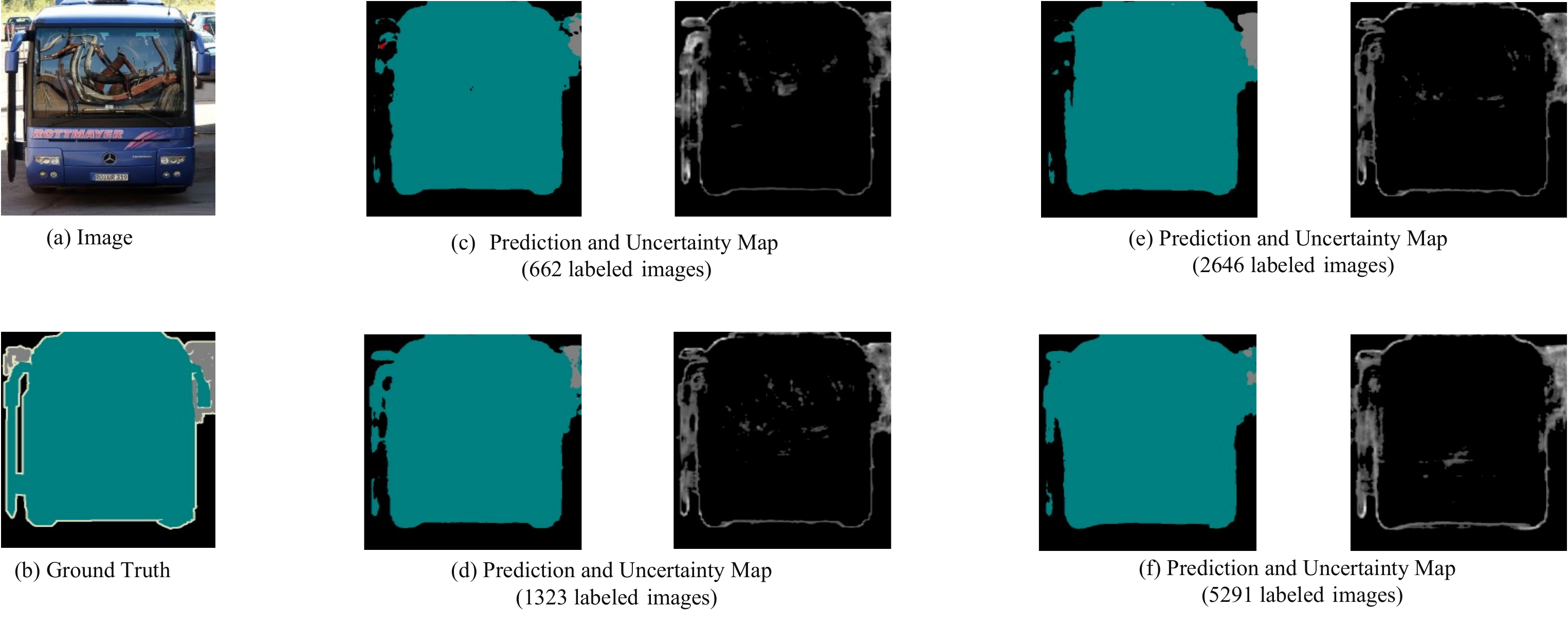}
\vspace{-4ex}
\caption{First set of visualization results on PASCAL VOC 2012 ({\it blender})  under different training protocols. The predictions and their corresponding uncertainty maps are shown. } 
\end{figure}
 
\begin{figure} 
\centering
\includegraphics[width=\linewidth]{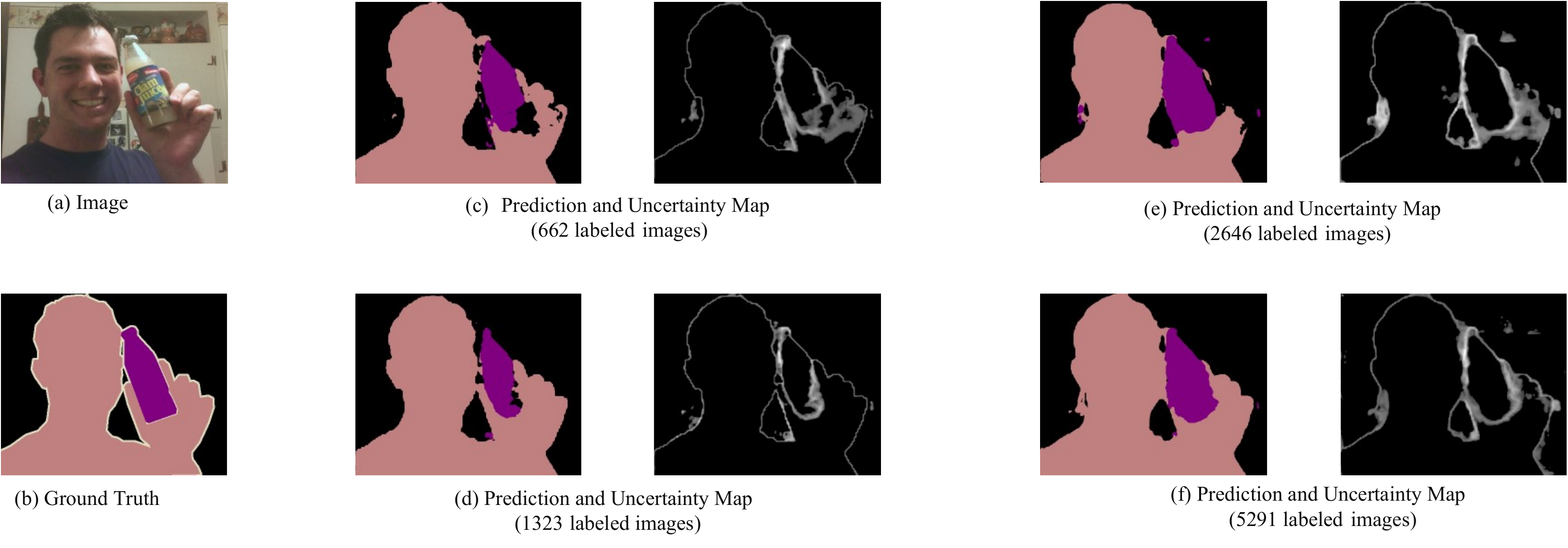}
\vspace{-4ex}
\caption{Second set of visualization results on PASCAL VOC 2012 ({\it blender})  under different training protocols. The predictions and their corresponding uncertainty maps are shown. }  
\end{figure} 
 
\begin{figure} 
\centering
\includegraphics[width=\linewidth]{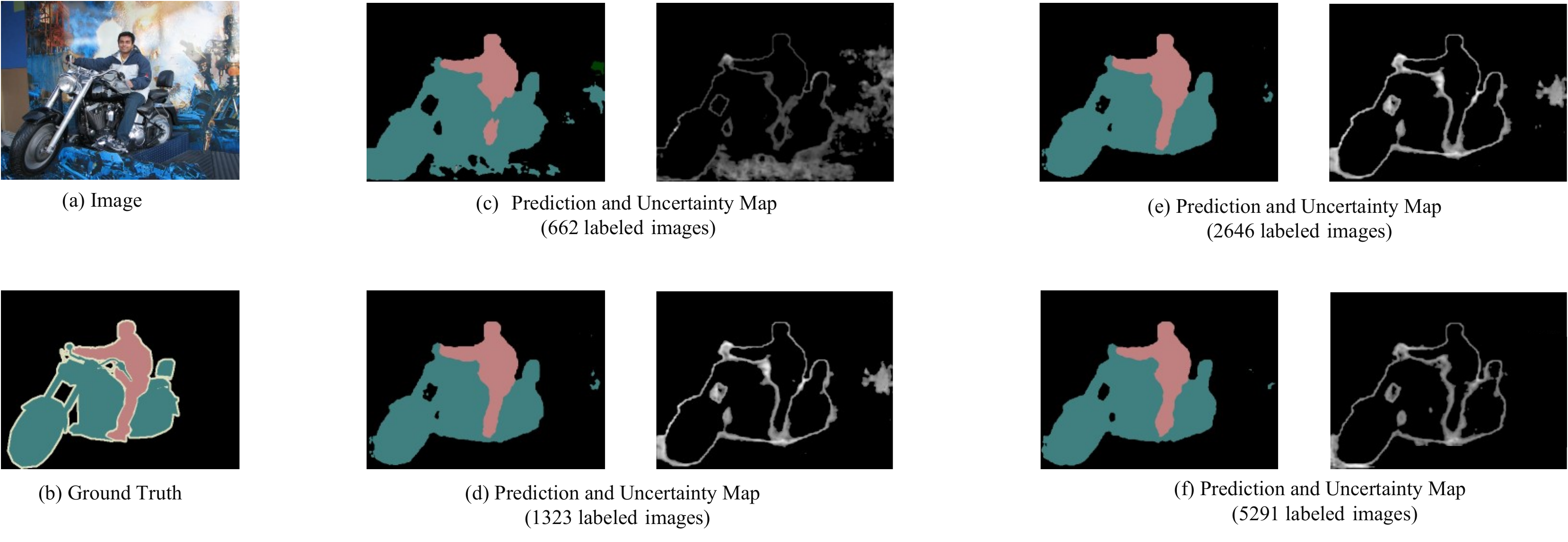}
\vspace{-4ex}
\caption{Third set of visualization results on PASCAL VOC 2012 ({\it blender})  under different training protocols. The predictions and their corresponding uncertainty maps are shown. } 
\end{figure}

\begin{figure} 
\centering
\includegraphics[width=  \linewidth]{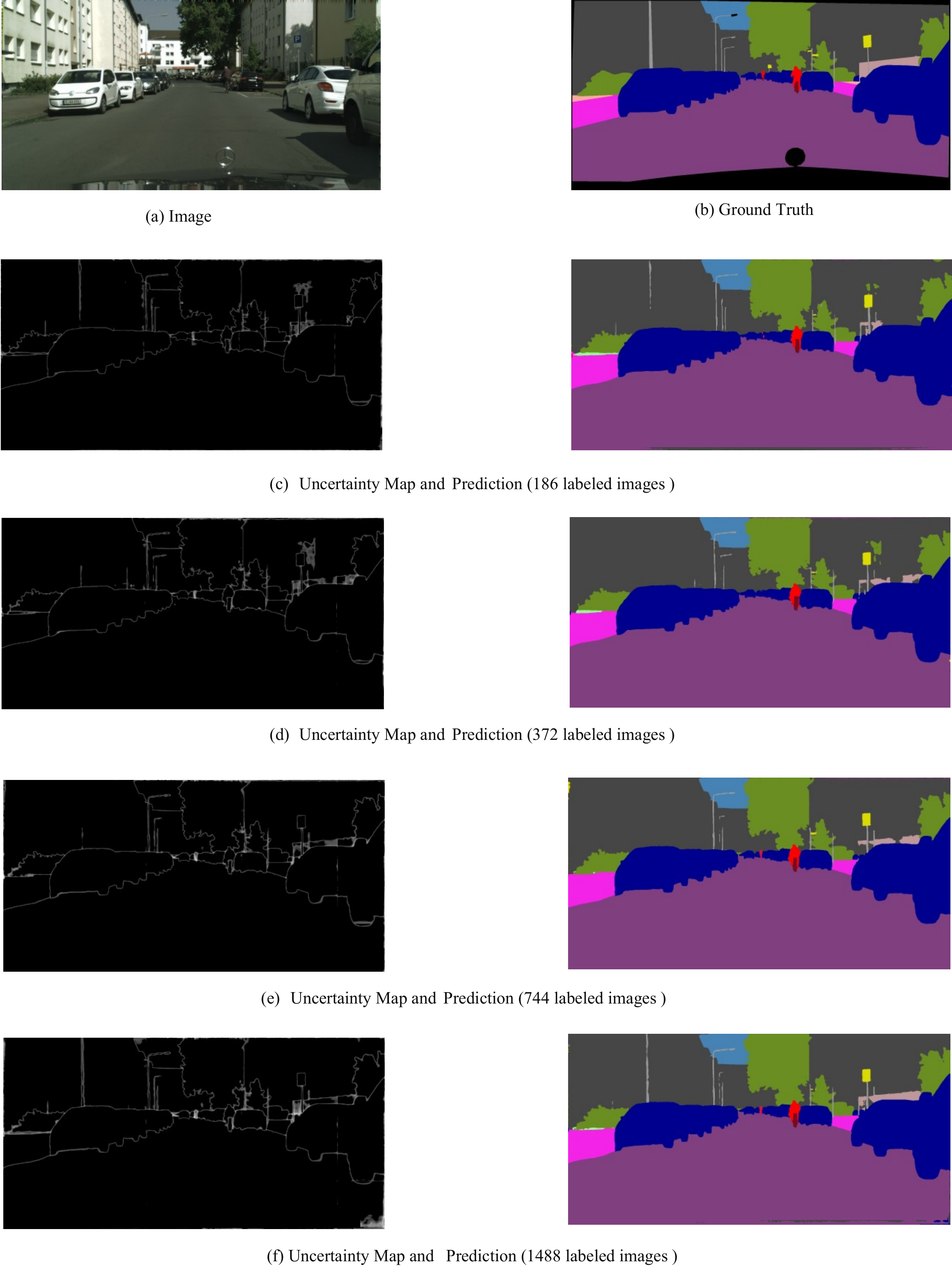}
\vspace{-2ex}
\caption{First set of visualization results on Cityscapes  under different training protocols. The predictions and their corresponding uncertainty maps are shown. } 
\end{figure} 

\begin{figure} 
\centering
\includegraphics[width=  \linewidth]{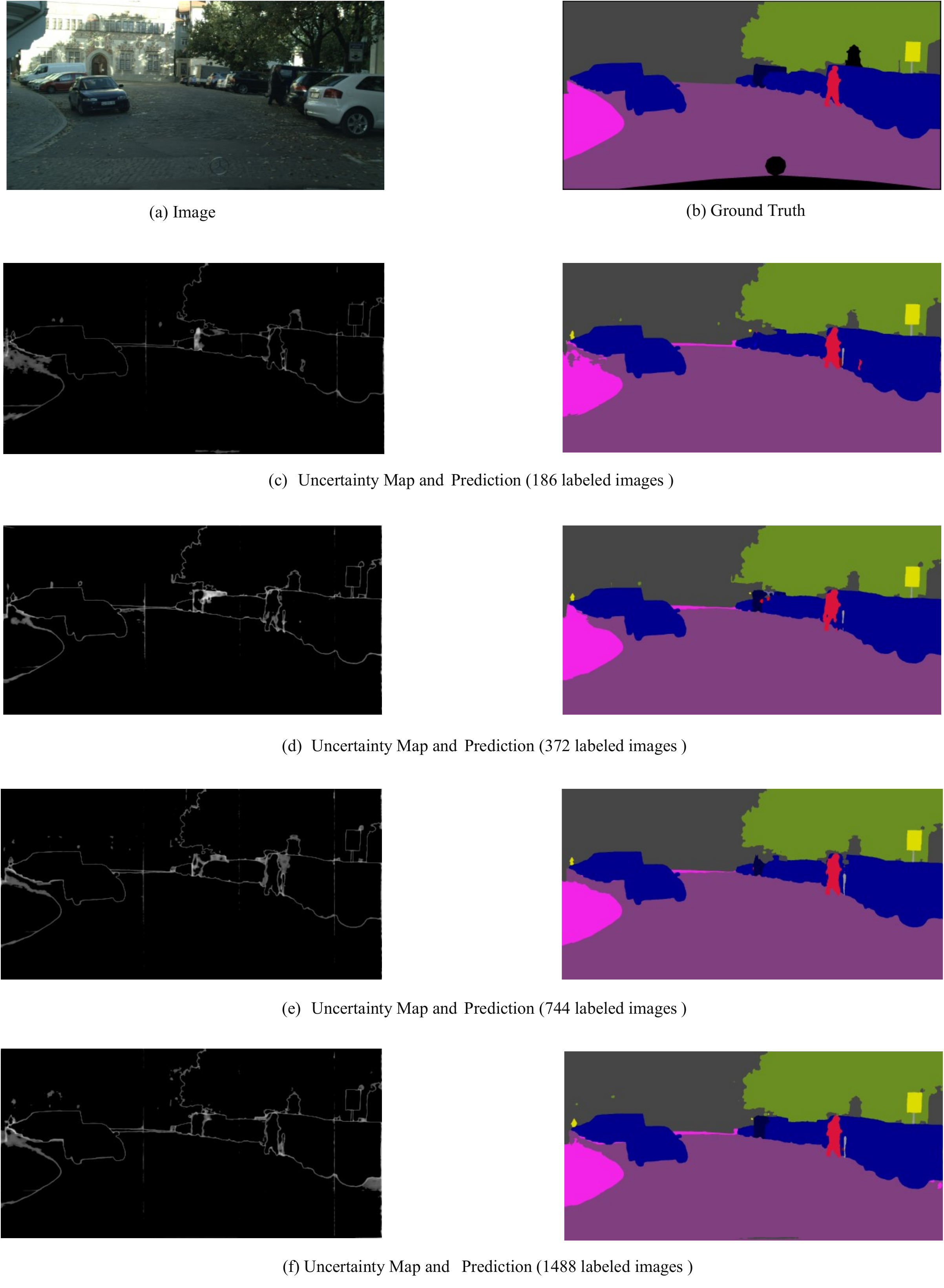}
\vspace{-2ex}
\caption{Second set of visualization results on Cityscapes under different training protocols. The predictions and their corresponding uncertainty maps are shown. }
\end{figure} 

\begin{figure} 
\centering
\includegraphics[width= \linewidth]{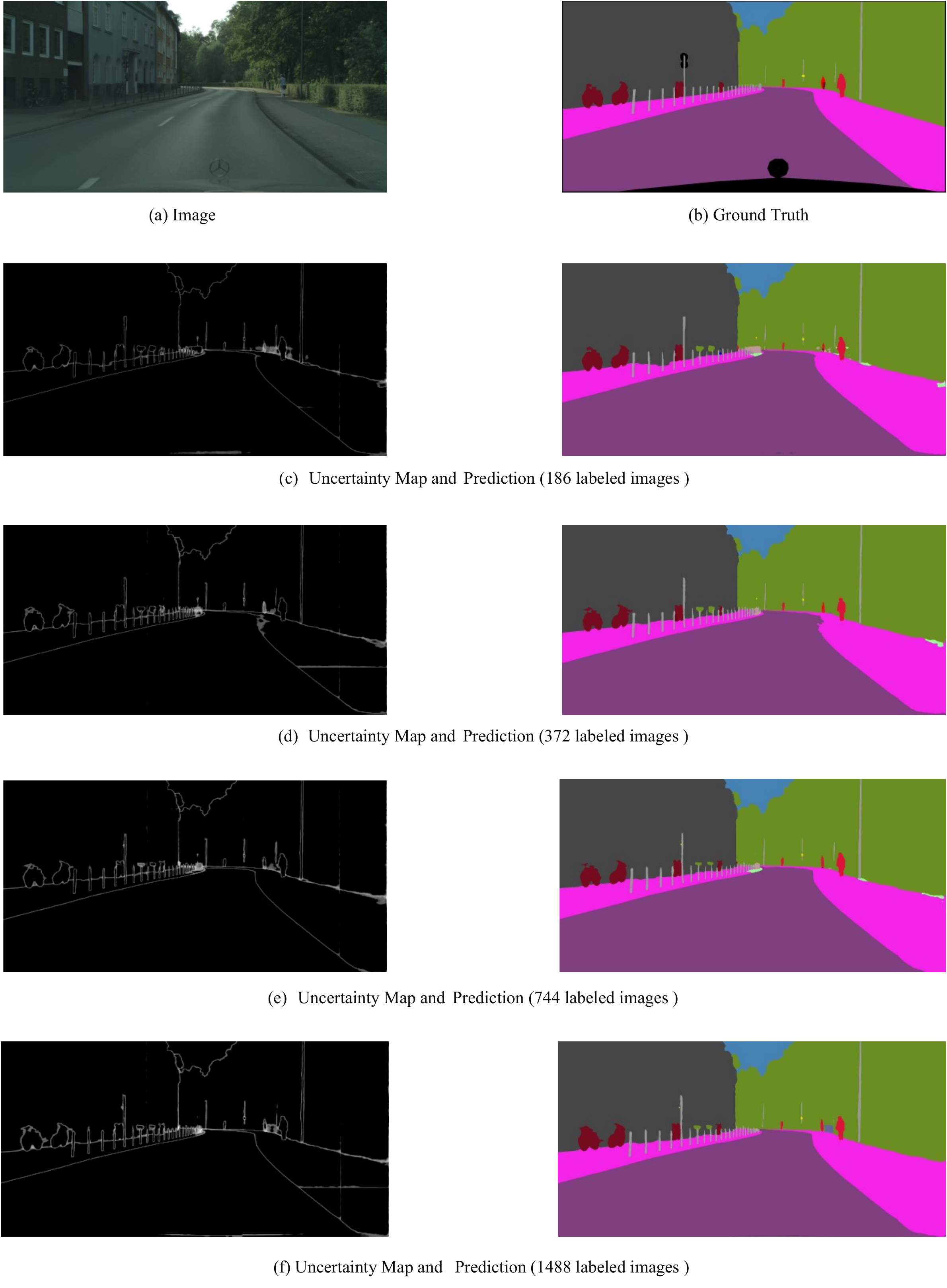}
\vspace{-2ex}
\caption{Third set of visualization results on Cityscapes  under different training protocols. The predictions and their corresponding uncertainty maps are shown. } 
\end{figure} 

\section{Summary}
\label{5.5}
In this chapter, we proposed a new probabilistic model, named NP-SemiSeg, which adjusts neural processes (NPs) to semi-supervised semantic segmentation. To better utilize the information from context data and target data, we integrated an attention aggregator into NP-SemiSeg for assigning higher weights to important information during aggregation, which is not considered in NP-Match. Our experimental results confirm the effectiveness of NP-SemiSeg in both accuracy and uncertainty estimation, thus highlighting its potential to supplant MC dropout as an innovative method for quantifying uncertainty in semi-supervised semantic segmentation.

\section{Limitations}
\label{5.6}

While NP-SemiSeg is superior to MC dropout with respect to uncertainty estimation, it is important to acknowledge its performance deterioration in some SSL settings, particularly with the Cityscapes dataset. This could potentially restrict its practical application. We hypothesize two potential causes for this degradation, both of which warrant further investigation.

Firstly, during the training phase, incorrect pixel-wise pseudo-labels may be assigned to unlabeled data. This could negatively affect NP-SemiSeg's ability to approximate the variational distribution to the true distribution over latent variables, leading to a subpar performance. A similar issue in NP-Match is partially resolved through an uncertainty-guided skew-geometric Jensen-Shannon (JS) divergence. However, it is challenging to directly apply this divergence to the task of segmentation.

Secondly, considering that the performance drop is pronounced in the Cityscapes dataset, it might be attributed to the sliding evaluation strategy, which contradicts NP-SemiSeg's use of global latent variables. NP-SemiSeg operates on the premise that a single latent variable is shared among all pixels in an image. This suggests that the global latent vector is dependent on the entire content (topic) of the target image. If a sliding evaluation strategy is employed, we do not obtain a global latent vector for the entire image, but rather a latent vector for the local region covered by the sliding window. This could negatively impact performance, given the importance of global information in generating a global latent vector for a target image.

\chapter{Conclusions}

SSL, a classic and pervasive learning paradigm in both academia and industry, reduces the need for extensive data annotation, allowing organizations to train foundational AI models more cost-effectively. Most recent SOTA methods rely heavily on the interplay between pseudo-labeling and consistency regularization. However, these methods can introduce errors during training when human annotators are not involved. This may result in AI models yielding subpar performance, which can have serious implications in critical fields such as medical image analysis and autonomous driving. Therefore, to enhance the safety and reliability of these models, it's vital to study their capability to quantify uncertainty, which indicates when AI models could provide incorrect predictions. Yet, these recent SOTA methods are deterministic approaches that don't account for modeling predictive distributions and quantifying uncertainty. 

Thus, in this dissertation, I delve into probabilistic models and their applications to SSL. The main contributions are summarized as follows: 

\begin{itemize}[leftmargin=*, itemsep=0.5pt]
\item 
In Chapter 3, we first focused on exploring probabilistic methods for semi-supervised medical image segmentation. In particular, we revisited the application of MC dropout for this task and identified a lack of theoretical foundations in previous works. To address this gap, we not only proposed a full Bayesian formulation for the task but also introduced a novel GBDL architecture based on this formulation. GBDL outperforms previous state-of-the-art methods on public medical benchmarks, offering both theoretical and practical contributions to semi-supervised medical image segmentation. 
The proposed Bayesian theoretical basis and the generative learning paradigm can provide a solid foundation and a new perspective, respectively, for future research to design semi-supervised medical image segmentation frameworks.

\item In Chapter 4, we investigated a new probabilistic approach for SSL. Typically, to quantify uncertainty for a given input with MC dropout, several rounds of feedforward passes are needed, which is time-consuming for large deep networks. This issue diminishes its practical value, and it is significant to find a new alternative. 
To this end, I adjusted NPs and then proposed a new framework named NP-Match for large-scale semi-supervised image classification. 
To demonstrate the effectiveness of NP-Match, I extensively evaluated it under different settings, including standard semi-supervised image classification, imbalanced semi-supervised image classification, and multi-label semi-supervised image classification.
NP-Match can not only achieve competitive results across different public benchmarks, but also estimate uncertainty much faster than MC-dropout-based probabilistic models. 
Our algorithm has the potential to revolutionize probabilistic models and the approach to uncertainty quantification under SSL, warranting further studies in the future.


\item In Chapter 5,inspired by the success of NP-Match, we further explored the application of NPs on semi-supervised semantic segmentation and proposed a new method named NP-SemiSeg. NP-SemiSeg is remolded from NPs to process 2D maps for segmentation. It also integrates an attention aggregator which assigns higher weights to the information that is more relevant to target data, alleviating the well-known underfitting issue of NPs. Thorough experimental results demonstrate that NP-SemiSeg surpasses MC dropout not just in terms of accuracy but also in speed concerning uncertainty estimation, showing its potential to be a more effective probabilistic model for semi-supervised semantic segmentation. 
The proposed method makes NPs applicable to semi-supervised semantic segmentation, which could have a significant impact on all related safety-critical areas, including autonomous driving systems and medical imaging analysis systems.

\end{itemize}

In summary, this thesis not only possesses practical value but also opens the door to a new research area, namely, developing probabilistic approaches for SSL instead of investigating various new deterministic approaches to surpass all the current SOTA results.

Regarding the future, there are several exciting avenues for research, owing to the promising performance of NPs in SSL:
\begin{itemize}[leftmargin=*, itemsep=0.5pt]
\item 
It would be interesting to apply NPs to other SSL tasks such as object detection and pose estimation. Uncertainty quantification is of great practical value, as it can be a useful reference when AI models make decisions. Given that NPs have achieved promising results on two fundamental computer vision tasks, namely classification and segmentation, we highly recommend studying or evaluating NPs on various vision tasks. The promotion of NPs on different tasks may introduce a new way of quantifying uncertainty in both computer vision and medical vision fields. 

\item
Several successful NP variants have emerged since the original NPs \cite{garnelo2018neural}. Exploring these in SSL could lead to more robust frameworks offering precise predictions and well-calibrated uncertainty estimates. For instance, the predictions of NPs depend on the relative position in input space of context and target points rather than the absolute one; i.e., NPs do not model translation equivariance. Thus, NPs are unable to generalize well when input points are out of range, which means that NPs may fail in open-set SSL tasks. One might investigate and implement translation equivariance when adjusting NPs to open-set SSL tasks, which can be inspired by a  NP variant, namely convolutional conditional NPs \cite{gordon2019convolutional}.

\item Currently, the performance of NPs is still restricted by the setting of limited annotations. Even though we tried to solve this issue by proposing a new loss term called the uncertainty-guided skew-geometric Jensen-Shannon (JS) divergence, it is not easy to transplant to other tasks, and its calculation is also complicated. Hence, to further optimize NPs, one can consider developing a new way to enhance the robustness of NPs against missing labels or noisy labels, which were not studied in the original works \cite{garnelo2018conditional, garnelo2018neural, gordon2019convolutional}.

\item As NP-based frameworks have mainly been evaluated on traditional computer vision datasets containing natural images, it would be fascinating to explore their applications on other data formats, such as medical and satellite images. For example, in the medical domain, uncertainty estimation has been a critical tool to guarantee the safety of AI-based diagnoses. Given that NPs have been an effective and efficient probabilistic model in traditional computer vision tasks, it is valuable to extend NPs to the medical domain.

\end{itemize}

\appendix

\include{appendix_rethink}
\include{appendix_npmatch}
\include{appendix_npsemiseg}

\addcontentsline{toc}{chapter}{Bibliography}
\bibliography{refs}        
\bibliographystyle{plain}  

\end{document}

%% file: declarations.tex
\chapter*{Declarations}

I solemnly affirm that, unless explicitly acknowledged, the contents of this dissertation are original and have not been submitted, in whole or in part, for the pursuit of any other degree or qualification at this University or elsewhere. This dissertation is my own work and contains nothing which is the outcome of work done in collaboration with others, barring those explicitly specified in the text.
 
\vspace{5ex}
{\hfill Jianfeng Wang }

{\hfill May 2023 }

%% file: acknowledgements.tex
\chapter*{Acknowledgements}

Firstly, I extend my profound gratitude to my supervisor, Prof. Thomas Lukasiewicz. His steadfast support, ceaseless encouragement, and inexhaustible patience have been my pillar throughout my DPhil studies. With each progression of my project, he has prioritized my research interests, for which I am immensely grateful. I cherish his invaluable support, which has been present from the inception of my DPhil project when I was still discerning the direction of my research.

Secondly, I wish to express my heartfelt appreciation to my family. Their unwavering encouragement, especially amidst the hurdles of daily life and the challenges of my research, has been my solace. I consider myself immensely fortunate to have such supportive parents who have stood by every decision I've made. To my maternal grandparents, I owe a deep debt of gratitude for their financial and moral support that has fortified my resilience, allowing me to overcome any obstacle. To my paternal grandparents, who watch over me from heaven, I hope that my accomplishments have made you proud.

Thirdly, I wish to acknowledge my research collaborators: Xiaolin Hu, Daniela Massiceti, Vladimir Pavlovic, Jianfei Cai, and Alexandros Neophytou. Their generous provision of computational resources and invaluable feedback on my research have been instrumental in my progress.

I am also deeply indebted to my peers at the Intelligent Systems Lab for creating a warm and enthusiastic environment.

Lastly, I wish to thank my friends at Oxford for the endless moments of joy. A special mention goes to my good friend, Yuedong Chen, whose regular, casual conversations have been my stress reliever.

%% file: abstract.tex
\chapter*{Abstract}

Deep neural networks are increasingly harnessed for computer vision tasks, thanks to their robust performance. However, their training demands large-scale labeled datasets, which are labor-intensive to prepare. Semi-supervised learning (SSL) offers a solution by learning from a mix of labeled and unlabeled data.

While most state-of-the-art SSL methods follow a deterministic approach, the exploration of their probabilistic counterparts remains limited. This research area is important because probabilistic models can provide uncertainty estimates critical for real-world applications. 
For instance, SSL-trained models may fall short of those trained with supervised learning due to potential pseudo-label errors in unlabeled data, and these models are more likely to make wrong predictions in practice. 
Especially in critical sectors like medical image analysis and autonomous driving, decision-makers must understand the model's limitations and when incorrect predictions may occur, insights often provided by uncertainty estimates. Furthermore, uncertainty can also serve as a criterion for filtering out unreliable pseudo-labels when unlabeled samples are used for training, potentially improving deep model performance.

This thesis furthers the exploration of probabilistic models for SSL. Drawing on the widely-used Bayesian approximation tool, Monte Carlo (MC) dropout, I propose a new probabilistic framework, the Generative Bayesian Deep Learning (GBDL) architecture, for semi-supervised medical image segmentation. This approach not only mitigates potential overfitting found in previous methods but also achieves superior results across four evaluation metrics. Unlike its empirically designed predecessors, GBDL is underpinned by a full Bayesian formulation, providing a theoretical probabilistic foundation.

Acknowledging MC dropout's limitations, I introduce NP-Match, a novel probabilistic approach for large-scale semi-supervised image classification. We evaluated NP-Match's generalization capabilities through extensive experiments in different challenging settings such as standard, imbalanced, and multi-label semi-supervised image classification. 
According to the experimental results, NP-Match not only competes favorably with previous state-of-the-art methods but also estimates uncertainty more rapidly than MC-dropout-based models, thus enhancing both training and testing efficiency.

Lastly, I propose NP-SemiSeg, a new probabilistic model for semi-supervised semantic segmentation. This flexible model can be integrated with various existing segmentation frameworks to make predictions and estimate uncertainty. Experiments indicate that NP-SemiSeg surpasses MC dropout in accuracy, uncertainty quantification, and speed.

%% file: publications.tex
\chapter*{Publications}

{\bf Papers included in the thesis.} This thesis is based on the following papers that I have published or finished as the first author during my DPhil:  

\vspace{3ex}

1. Jianfeng Wang and Thomas Lukasiewicz. ``Rethinking Bayesian deep learning methods for semi-supervised volumetric medical image segmentation." In CVPR 2022.

\vspace{.5ex}

2. Jianfeng Wang, Thomas Lukasiewicz,  Daniela Massiceti, Xiaolin Hu, Vladimir Pavlovic, and Alexandros Neophytou. ``NP-Match: When neural processes meet semi-supervised learning." In ICML 2022.

\vspace{.5ex}

3. Jianfeng Wang,  Xiaolin Hu, and Thomas Lukasiewicz. ``NP-Match: Towards a new probabilistic model for semi-supervised learning." In  arXiv:2301.13569.

 \vspace{.5ex}
 
4. Jianfeng Wang, Daniela Massiceti, Xiaolin Hu, Vladimir Pavlovic, and Thomas Lukasiewicz. ``NP-SemiSeg: When neural processes meet semi-supervised semantic segmentation." In ICML 2023.

\vspace{7ex}

{\bf Papers excluded in the thesis.} During my DPhil, I have also published or finished as the first author other two papers that are not included in this thesis:

\vspace{3ex}

1. Jianfeng Wang, Thomas Lukasiewicz, Xiaolin Hu, Jianfei Cai, and Zhenghua Xu. ``RSG: A simple but effective module for learning imbalanced datasets." In CVPR 2021.

\vspace{.5ex}

2. Jianfeng Wang, David Wong, Sunando Sengupta, Alexandros Neophytou, Thomas Lukasiewicz, and Eric Sommerlade. ``Learning generalizable morphable radiance fields for dynamic facial avatars from monocular RGB videos." under review. 

%% file: appendix_rethink.tex
\chapter{Rethinking Bayesian Deep Learning Methods for Semi-Supervised Volumetric Medical Image Segmentation Appendices}

\section{Derivation of ELBO}

\emph{Proof.} As for the joint distribution $P(X,Y)$, we have: 

\begin{equation}
\small
\begin{aligned}
 &logP(X,Y) = log \int_Z P(X, Y, Z) dZ \\
 & = log \int_Z \frac{P(X, Y, Z)}{Q(Z|X)}Q(Z|X) dZ \\
 &  \geq \mathbb{E}_Q[log \frac{P(X, Y, Z)}{Q(Z|X)}] \\
 & =  \mathbb{E}_Q[log \frac{P(Y|X, Z)P(X|Z)P(Z)}{Q(Z|X)}] \\
 & = \mathbb{E}_{Q}[logP(Y| X, Z) + logP(X|Z)] - \mathbb{E}_{Q}[log (\frac{Q(Z|X)} {P(Z)})],
\end{aligned}
\end{equation}
where $Q(Z|X)$ is the variational distribution, and $\mathbb{E}_{Q}$ denotes the expectation over $Q(Z|X)$. \hfill $\square$

\section{Proof of Theorem 3.3.2.1.}

\emph{Proof.} Consider two $D$-dimensional multivariate Gaussian distributions
$\mathcal{N}(\mu_1, \Sigma_{1})$ and $\mathcal{N}(\mu_2, \Sigma_{2})$. 
Then, the product of their PDFs can be written as follows:
\begin{equation}
\tiny		
\label{eqn:proof1}
\begin{aligned}
&(2\pi)^{-D}det[\Sigma_1]^{-\frac{1}{2}}det[\Sigma_2]^{-\frac{1}{2}}e^{-\frac{1}{2}[(x-\mu_1)^T\Sigma_1^{-1}(x-\mu_1)+(x-\mu_2)^T\Sigma_2^{-1}(x-\mu_2)]} \\
&= C_1e^{-\frac{1}{2}(x^T(\Sigma_1^{-1} + \Sigma_2^{-1})x - x^T(\Sigma_1^{-1}\mu_1 + \Sigma_2^{-1}\mu_2)-(\mu_1^T\Sigma_1^{-1}+\mu_2^T\Sigma_2^{-1})x + (\mu_1^T \Sigma_1^{-1} \mu_1 + \mu_2^T \Sigma_2^{-1}\mu_2))} \\
&=  C_1e^{-\frac{1}{2}(x^T(\Sigma_1^{-1} + \Sigma_2^{-1})x - x^T(\Sigma_1^{-1} + \Sigma_2^{-1})(\Sigma_1^{-1} + \Sigma_2^{-1})^{-1}(\Sigma_1^{-1}\mu_1 + \Sigma_2^{-1}\mu_2)-(\mu_1^T\Sigma_1^{-1}+\mu_2^T\Sigma_2^{-1})x + (\mu_1^T \Sigma_1^{-1} \mu_1 + \mu_2^T \Sigma_2^{-1}\mu_2))} \\
&=  C_1e^{-\frac{1}{2}(x^T(\Sigma_1^{-1} + \Sigma_2^{-1})(x - (\Sigma_1^{-1} + \Sigma_2^{-1})^{-1}(\Sigma_1^{-1}\mu_1 + \Sigma_2^{-1}\mu_2))-(\mu_1^T\Sigma_1^{-1}+\mu_2^T\Sigma_2^{-1})x + (\mu_1^T \Sigma_1^{-1} \mu_1 + \mu_2^T \Sigma_2^{-1}\mu_2))} \\
&=  C_1e^{-\frac{1}{2}(x^T(\Sigma_1^{-1} + \Sigma_2^{-1})(x - (\Sigma_1^{-1} + \Sigma_2^{-1})^{-1}(\Sigma_1^{-1}\mu_1 + \Sigma_2^{-1}\mu_2))-(\mu_1^T\Sigma_1^{-1}+\mu_2^T\Sigma_2^{-1})x + (\mu_1^T\Sigma_1^{-1}+\mu_2^T\Sigma_2^{-1})(\Sigma_1^{-1} + \Sigma_2^{-1})^{-1}(\Sigma_1^{-1}\mu_1 + \Sigma_2^{-1}\mu_2) + C_2)} \\
&=  C_1e^{-\frac{1}{2}(x^T(\Sigma_1^{-1} + \Sigma_2^{-1})(x - (\Sigma_1^{-1} + \Sigma_2^{-1})^{-1}(\Sigma_1^{-1}\mu_1 + \Sigma_2^{-1}\mu_2))-(\mu_1^T\Sigma_1^{-1}+\mu_2^T\Sigma_2^{-1})(x - (\Sigma_1^{-1} + \Sigma_2^{-1})^{-1}(\Sigma_1^{-1}\mu_1 + \Sigma_2^{-1}\mu_2)) + C_2)} \\
&=  C_1e^{-\frac{1}{2}(x^T(\Sigma_1^{-1} + \Sigma_2^{-1})(x - (\Sigma_1^{-1} + \Sigma_2^{-1})^{-1}(\Sigma_1^{-1}\mu_1 + \Sigma_2^{-1}\mu_2))-(\mu_1^T\Sigma_1^{-1}+\mu_2^T\Sigma_2^{-1})(\Sigma_1^{-1} + \Sigma_2^{-1})^{-1}(\Sigma_1^{-1} + \Sigma_2^{-1})(x - (\Sigma_1^{-1} + \Sigma_2^{-1})^{-1}(\Sigma_1^{-1}\mu_1 + \Sigma_2^{-1}\mu_2)) + C_2)} \\
&=  C_1e^{-\frac{1}{2}((x^T - (\mu_1^T\Sigma_1^{-1}+\mu_2^T\Sigma_2^{-1})(\Sigma_1^{-1} + \Sigma_2^{-1})^{-1})(\Sigma_1^{-1} + \Sigma_2^{-1})(x - (\Sigma_1^{-1} + \Sigma_2^{-1})^{-1}(\Sigma_1^{-1}\mu_1 + \Sigma_2^{-1}\mu_2))+ C_2)} \\
&= C_1e^{-\frac{1}{2}((x - (\Sigma_1^{-1} + \Sigma_2^{-1})^{-1}(\Sigma_1^{-1}\mu_1 + \Sigma_2^{-1}\mu_2))^T(\Sigma_1^{-1} + \Sigma_2^{-1})(x - (\Sigma_1^{-1} + \Sigma_2^{-1})^{-1}(\Sigma_1^{-1}\mu_1 + \Sigma_2^{-1}\mu_2))+ C_2)} \\
&= C_3e^{-\frac{1}{2}(x - (\Sigma_1^{-1} + \Sigma_2^{-1})^{-1}(\Sigma_1^{-1}\mu_1 + \Sigma_2^{-1}\mu_2))^T(\Sigma_1^{-1} + \Sigma_2^{-1})(x - (\Sigma_1^{-1} + \Sigma_2^{-1})^{-1}(\Sigma_1^{-1}\mu_1 + \Sigma_2^{-1}\mu_2))}, 
\end{aligned}
\end{equation}
where $C_1 = (2\pi)^{-D}det[\Sigma_1]^{-\frac{1}{2}}det[\Sigma_2]^{-\frac{1}{2}}$, $C_2$ is a constant for aborting the terms used for completing the square relative to $x$, and $C_3 = C_1e^{-\frac{1}{2}C_2}$.
The last formula of Eq.~(\ref{eqn:proof1}) is an unnormalized Gaussian curve with mean $\mu_* = (\Sigma_1^{-1} + \Sigma_2^{-1})^{-1}(\Sigma_1^{-1}\mu_1 + \Sigma_2^{-1}\mu_2)$ and covariance $\Sigma_* = (\Sigma_1^{-1} + \Sigma_2^{-1})^{-1}$. Then, we replace the covariance matrix with the precision matrix, and the mean and the precision of the unnormalized Gaussian curve becomes $\mu_* = \Lambda_*^{-1}(\Lambda_1\mu_1 + \Lambda_2\mu_2)$ and $\Lambda_* =  \Lambda_1 + \Lambda_2 $, respectively.
The theorem can be directly extended to the product of more than two multivariate Gaussian PDFs. \hfill $\square$

\section{Proof of Corollary 3.3.2.1.}
\emph{Proof.} Consider two $D$-dimensional multivariate Gaussian distributions
$P_1 = \mathcal{N}(\mu_1, \Sigma_{1})$ and $P_2 = \mathcal{N}(\mu_2, \Sigma_{2})$. Then, $\mathbb{E}_{P_1}[log({P_1}/{P_2})]$ is:

\begin{equation} 
\footnotesize
\label{eqn:proof2}
\begin{aligned}
&\mathbb{E}_{P_1}[log{P_1} - log{P_2}] \\
&= \frac{1}{2} \mathbb{E}_{P_1}[-logdet[\Sigma_1] - (x-\mu_1)^T\Sigma_1^{-1}(x-\mu_1) + logdet[\Sigma_2] + (x-\mu_2)^T\Sigma_2^{-1}(x-\mu_2) ]  \\
&= \frac{1}{2} ( log\frac{det[\Sigma_2]}{det[\Sigma_1]} + \mathbb{E}_{P_1}[ - (x-\mu_1)^T\Sigma_1^{-1}(x-\mu_1) +  (x-\mu_2)^T\Sigma_2^{-1}(x-\mu_2) ] ) \\
&= \frac{1}{2} ( log\frac{det[\Sigma_2]}{det[\Sigma_1]} + \mathbb{E}_{P_1}[ - tr[\Sigma_1^{-1}\Sigma_1] +  tr[\Sigma_2^{-1}(xx^T-2x\mu_2^T+\mu_2\mu_2^T)] ] ) \\
&= \frac{1}{2} log\frac{det[\Sigma_2]}{det[\Sigma_1]} - \frac{D}{2}  + \frac{1}{2} \mathbb{E}_{P_1}[  tr[\Sigma_2^{-1}(xx^T-2x\mu_2^T+\mu_2\mu_2^T)] ]  \\
&= \frac{1}{2} log\frac{det[\Sigma_2]}{det[\Sigma_1]} - \frac{D}{2}  + \frac{1}{2} \mathbb{E}_{P_1}[  tr[\Sigma_2^{-1}((x-\mu_1)(x-\mu_1)^T + 2\mu_1x^T-\mu_1\mu_1^T-2x\mu_2^T+\mu_2\mu_2^T)] ]  \\
&= \frac{1}{2} log\frac{det[\Sigma_2]}{det[\Sigma_1]} - \frac{D}{2}  + \frac{1}{2}  tr[\Sigma_2^{-1}(\Sigma_1 + \mu_1\mu_1^T - 2\mu_2\mu_1^T+\mu_2\mu_2^T)]  \\
&= \frac{1}{2} log\frac{det[\Sigma_2]}{det[\Sigma_1]} - \frac{D}{2}  + \frac{1}{2}  tr[\Sigma_2^{-1}\Sigma_1] + \frac{1}{2} tr[\mu_1^T\Sigma_2^{-1}\mu_1 - 2\mu_1^T\Sigma_2^{-1}\mu_2+\mu_2^T\Sigma_2^{-1}\mu_2)]  \\
&=  \frac{1}{2} log\frac{det[\Sigma_2]}{det[\Sigma_1]} - \frac{D}{2}  + \frac{1}{2}  tr[\Sigma_2^{-1}\Sigma_1] + \frac{1}{2} (\mu_2 - \mu_1)^T\Sigma_2^{-1}(\mu_2 - \mu_1).
\end{aligned}
\end{equation}

Based on our assumption in the paper body (left column, lines 442--445), $P_1$ follows a multivariate Gaussian distribution, and $P_2$ follows a multivariate standard normal distribution. The above equation becomes $- \frac{1}{2} logdet[\Sigma_1] - \frac{D}{2}  + \frac{1}{2}  tr[\Sigma_1] + \frac{1}{2} \mu_1^T\mu_1$. Then, we replace  $\mu_1$ and $\Sigma_1$ with the mean and precision of Theorem 1, obtaining the following equation:
\begin{equation}
\footnotesize
\label{eqn:proof2_1}
\begin{aligned}
\frac{1}{2} (-log\, det [(\sum\nolimits_{i=0}^{n-1}\Lambda_i)^{-1}] +  tr[(\sum\nolimits_{i=0}^{n-1}\Lambda_i)^{-1}] \ +\nonumber  (\sum\nolimits_{i=0}^{n-1}\Lambda_i\mu_i)^T (\sum\nolimits_{i=0}^{n-1}\Lambda_i)^{-2} (\sum\nolimits_{i=0}^{n-1}\Lambda_i\mu_i)
- {D}).
\end{aligned}
\end{equation}
\hfill $\square$

\vspace{-1ex}
\begin{table}[h]
    \centering
    \resizebox{0.85\textwidth}{!}{
    \begin{tabular}{|c|c|}
       \hline
       Type  &  Configuration \\
      \hline 
      3D Conv &  \#In-C: 3, \#F: 64, K: $ 3\times 3\times 3$, S: $ 1\times 1\times 1$, P:  $ 1\times 1\times 1$ \\
      \hline
      3D Conv &  \#In-C: 64, \#F: 64, K: $ 3\times 3\times 3$, S: $ 1\times 1\times 1$, P:  $ 1\times 1\times 1$ \\
      \hline
      3D Max-Pooling & K: $ 2\times 2\times 1$, S: $ 2\times 2\times 1$, P:  0 \\
      \hline
      3D Conv &  \#In-C: 64, \#F: 128, K: $ 3\times 3\times 3$, S: $ 1\times 1\times 1$, P:  $ 1\times 1\times 1$ \\
      \hline
      3D Conv &  \#In-C: 128, \#F: 128, K: $ 3\times 3\times 3$, S: $ 1\times 1\times 1$, P:  $ 1\times 1\times 1$ \\
      \hline
      3D Max-Pooling & K: $ 2\times 2\times 1$, S: $ 2\times 2\times 1$, P:  0 \\
      \hline
      3D Conv &  \#In-C: 128, \#F: 256, K: $ 3\times 3\times 3$, S: $ 1\times 1\times 1$, P:  $ 1\times 1\times 1$ \\
      \hline
      3D Conv &  \#In-C: 256, \#F: 256, K: $ 3\times 3\times 3$, S: $ 1\times 1\times 1$, P:  $ 1\times 1\times 1$ \\
      \hline
      3D Max-Pooling & K: $ 2\times 2\times 1$, S: $ 2\times 2\times 1$, P:  0 \\
      \hline
      3D Conv & \#In-C: 256, \#F: 512, K: $ 3\times 3\times 3$, S: $ 1\times 1\times 1$, P:  $ 1\times 1\times 1$ \\
      \hline
      3D Conv & \#In-C: 512, \#F: 512, K: $ 3\times 3\times 3$, S: $ 1\times 1\times 1$, P:  $ 1\times 1\times 1$ \\
      \hline
      3D Max-Pooling & K: $ 2\times 2\times 1$, S: $ 2\times 2\times 1$, P:  0 \\
      \hline
      3D Conv &  \#In-C: 512, \#F: 512, K: $ 3\times 3\times 3$, S: $ 1\times 1\times 1$, P:  $ 1\times 1\times 1$ \\
      \hline
      3D Conv &  \#In-C: 512, \#F: 512, K: $ 3\times 3\times 3$, S: $ 1\times 1\times 1$, P:  $ 1\times 1\times 1$ \\
      \hline
    \end{tabular}
    }
    \vspace{-0.5ex}
    \caption{Network configuration of the 3D-UNet encoder. Each 3D convolution kernel is followed by an instance normalization layer \cite{ulyanov2017improved} and a ReLU function, which are omitted for simplicity.}
    
    \label{tab:3d_unet_encoder}
\end{table}

\begin{table}[h]
    \centering
    \resizebox{0.85\textwidth}{!}{
    \begin{tabular}{|c|c|}
       \hline
       Type  &  Configuration \\
      \hline 
      3D Up-sampling & Scale factor: $ 2\times 2\times 1$ \\
      \hline
      3D Conv &  \#In-C: 1024, \#F: 512, K: $ 3\times 3\times 3$, S: $ 1\times 1\times 1$, P:  $ 1\times 1\times 1$ \\
      \hline
      3D Conv & \#In-C: 512, \#F: 512, K: $ 3\times 3\times 3$, S: $ 1\times 1\times 1$, P:  $ 1\times 1\times 1$ \\
      \hline
      3D Up-sampling & Scale factor: $ 2\times 2\times 1$ \\
      \hline
      3D Conv &  \#In-C: 512, \#F: 256, K: $ 3\times 3\times 3$, S: $ 1\times 1\times 1$, P:  $ 1\times 1\times 1$ \\
      \hline
      3D Conv &  \#In-C: 256, \#F: 256, K: $ 3\times 3\times 3$, S: $ 1\times 1\times 1$, P:  $ 1\times 1\times 1$ \\
      \hline
      3D Up-sampling & Scale factor: $ 2\times 2\times 1$ \\
      \hline
      3D Conv &  \#In-C: 256, \#F: 128, K: $ 3\times 3\times 3$, S: $ 1\times 1\times 1$, P:  $ 1\times 1\times 1$ \\
      \hline
      3D Conv &  \#In-C: 128, \#F: 128, K: $ 3\times 3\times 3$, S: $ 1\times 1\times 1$, P:  $ 1\times 1\times 1$ \\
      \hline
      3D Up-sampling & Scale factor: $ 2\times 2\times 1$\\
      \hline
      3D Conv &  \#In-C: 128, \#F: 64, K: $ 3\times 3\times 3$, S: $ 1\times 1\times 1$, P:  $ 1\times 1\times 1$ \\
      \hline
      3D Conv &  \#In-C: 64,  \#F: 64, K: $ 3\times 3\times 3$, S: $ 1\times 1\times 1$, P:  $ 1\times 1\times 1$ \\
      \hline
      3D Conv &  \#In-C: 64,  \#F: 3, K: $ 3\times 3\times 3$, S: $ 1\times 1\times 1$, P:  $ 1\times 1\times 1$ \\
      \hline
    \end{tabular}
    }
    \vspace{-0.5ex}
    \caption{Network configuration of the 3D-UNet decoder that is used for reconstruction. Each 3D convolution kernel is followed by an instance normalization layer \cite{ulyanov2017improved} and a ReLU function except for the last one, which are omitted for simplicity. The last 3D convolution kernel is followed by a TanH function.}

    \label{tab:3d_unet_decoder1}
\end{table}

\begin{table}[h]
    \centering
    \resizebox{0.8\textwidth}{!}{
    \begin{tabular}{|c|c|}
       \hline
       Type  &  Configuration \\
      \hline 
      3D Up-sampling & Scale factor: $ 2\times 2\times 1$ \\
      \hline
      3D Conv &  \#In-C: 1536, \#F: 512, K: $ 3\times 3\times 3$, S: $ 1\times 1\times 1$, P:  $ 1\times 1\times 1$ \\
      \hline
      3D Conv & \#In-C: 512, \#F: 512, K: $ 3\times 3\times 3$, S: $ 1\times 1\times 1$, P:  $ 1\times 1\times 1$ \\
      \hline
      3D Up-sampling & Scale factor: $ 2\times 2\times 1$ \\
      \hline
      3D Conv &  \#In-C: 512, \#F: 256, K: $ 3\times 3\times 3$, S: $ 1\times 1\times 1$, P:  $ 1\times 1\times 1$ \\
      \hline
      3D Conv &  \#In-C: 256, \#F: 256, K: $ 3\times 3\times 3$, S: $ 1\times 1\times 1$, P:  $ 1\times 1\times 1$ \\
      \hline
      3D Up-sampling & Scale factor: $ 2\times 2\times 1$ \\
      \hline
      3D Conv &  \#In-C: 256, \#F: 128, K: $ 3\times 3\times 3$, S: $ 1\times 1\times 1$, P:  $ 1\times 1\times 1$ \\
      \hline
      3D Conv &  \#In-C: 128, \#F: 128, K: $ 3\times 3\times 3$, S: $ 1\times 1\times 1$, P:  $ 1\times 1\times 1$ \\
      \hline
      3D Up-sampling & Scale factor: $ 2\times 2\times 1$ \\
      \hline
      3D Conv &  \#In-C: 128, \#F: 64, K: $ 3\times 3\times 3$, S: $ 1\times 1\times 1$, P:  $ 1\times 1\times 1$ \\
      \hline
      3D Conv &  \#In-C: 64,  \#F: 64, K: $ 3\times 3\times 3$, S: $ 1\times 1\times 1$, P:  $ 1\times 1\times 1$ \\
      \hline
      3D Conv &  \#In-C: 64,  \#F: 2, K: $ 3\times 3\times 3$, S: $ 1\times 1\times 1$, P:  $ 1\times 1\times 1$ \\
      \hline
    \end{tabular}
    }
    \vspace{-0.5ex}
    \caption{Network configuration of the 3D-UNet decoder that is used for generating label masks. Each 3D convolution kernel is followed by an instance normalization layer \cite{ulyanov2017improved} and a ReLU function except for the last one, which are omitted for simplicity. The last 3D convolution kernel is followed by a softmax function.}

    \label{tab:3d_unet_decoder2}
\end{table}

\begin{table}[h]
    \centering
    \resizebox{0.85\textwidth}{!}{
    \begin{tabular}{|c|c|}
       \hline
       Type  &  Configuration \\
      \hline 
      3D Conv &  \#In-C: 512, \#F: 512, K: $ 3\times 3\times 1$, S: $ 2\times 2\times 1$, P:  $ 1\times 1\times 0$ \\
      \hline
      Flatten &  Output shape: [$\textit{depth}$, $512 \times \frac{h}{2} \times \frac{w}{2}$]\\
      \hline
      FC &  \#In-D: $512 \times \frac{h}{2} \times \frac{w}{2}$, \#Out-D: 256 \\
      \hline
    \end{tabular}
    }
    \vspace{-0.5ex}
    \caption{Network for generating mean vectors or covariance matrices, which is used after the 3D-UNet encoder. ``h'' and ``w'' denote the height and width of the input feature maps, respectively. ``\#In-D'' and ``\#Out-D'' represent the input dimension and the output dimension of the FC layer, respectively.}

    \label{tab:3d_fc1}
\end{table}

\begin{table}[h]
    \centering
    \resizebox{0.89\textwidth}{!}{
    \begin{tabular}{|c|c|}
       \hline
       Type  &  Configuration \\
      \hline
      FC &  \#In-D: 256, \#Out-D: $512 \times \frac{h}{2} \times \frac{w}{2}$  \\
      \hline
      Reshape &  Output shape: [$512$, $\frac{h}{2}$,  $\frac{w}{2}$, $\textit{depth}$] \\
      \hline 
      3D Transpose Conv &  \#In-C: 512, \#F: 512, K: $ 3\times 3\times 1$, S: $ 2\times 2\times 1$, P:  $ 1\times 1\times 0$ \\
      \hline
    \end{tabular}
    }
    \vspace{-0.5ex}
    \caption{Network for transforming the latent representation of each slice into a feature map, which is used before the 3D UNet-decoder. ``h'' and ``w'' denote the height and width of the final output feature maps, respectively. ``\#In-D'' and ``\#Out-D'' represent the input dimension and the output dimension of the FC layer, respectively.}

   \label{tab:3d_fc2}
\end{table}

\section{Configuration of LRL}
\label{sec:configuration}

LRL contains two 3D-UNet encoders that have the same network configuration (shown in Table~\ref{tab:3d_unet_encoder}), and two 3D-UNet decoders (shown in Tables~\ref{tab:3d_unet_decoder1} and \ref{tab:3d_unet_decoder2}). 
Note that ``\#In-C'', ``\#F'', ``K'', ``S'', and ``P'' in these tables denote the number of channels of input, the number of filters, the kernel size, the stride, and the padding size, respectively. The setting of ``K'', ``S'', and ``P'' are written in the format ``$ \textit{Height} \times \textit{Width} \times \textit{Depth}$''. 
The scale factor denotes the multiplier to the height, the width, and the depth, when the upsampling operation takes place. To obtain the mean vector and the covariance matrix of each slice, another two small networks were used (shown in Table~\ref{tab:3d_fc1}), and they both receive the output of the 3D-UNet encoder. One will generate the mean vector, and the other one will generate a vector, denoted as $v$, which is used to calculate the covariance matrix with $(vv^T)^2+I_c$, where $I_c$ is a diagonal matrix whose diagonal elements are equal and larger than zero \footnote{To compute $L_{KL[Q(Z|X)||P(Z))]}$, the log-determinant term requires the input matrix to be positive definite. But the covariance matrix in multivariate Gaussian distribution only needs to be positive semi-definite, which may lead to infinite $L_{KL[Q(Z|X)||P(Z))]}$. Therefore, we add the $I_c$ to solve this issue.}. To transform the latent representation of each slice into feature maps for decoders, we use another small network that is shown in Table~\ref{tab:3d_fc2}.

The workflow of LRL is summarized as follows. For example, the input are volumes with the shape $128 \times 128 \times 32$. Thus, concerning an input volume $X$, its shape can be written as $3\times 128 \times 128 \times 32$ (we omit the batch dimension, as we only consider one input volume), where 3 is the number of channels of each slice in the volume, and 32 is the number of slices in the volume, i.e., depth. $X$ is fed to a 3D-UNet encoder, resulting in an output volume with the shape $512 \times 8 \times 8 \times 32$. Then, the output volume is fed to the small networks  (Table~\ref{tab:3d_fc1}) that are designed for generating mean vectors and the covariance matrices. After the mean vectors and the covariance matrices are obtained, they are further used to produce latent representations (denoted as $Z$) via the reparameterization trick. 
$Z$ is further processed by the small network (Table~\ref{tab:3d_fc2}) that is designed for transforming the latent representation of each slice into feature maps, resulting in a volume with the shape $512 \times 8 \times 8 \times 32$ (denoted as $Z_{map}$), and thereafter, $Z_{map}$ is used for reconstructing the input volume $X$ via the 3D-UNet decoder shown in Table~\ref{tab:3d_unet_decoder1}. At the same time, the input volume is fed to another 3D-UNet encoder to obtain the feature representation with the shape $512\times 8 \times 8 \times 32$, and then the feature representation is concatenated with  $Z_{map}$ along the channel dimension, which is the input to the 3D-UNet decoder shown in Table~\ref{tab:3d_unet_decoder2} for generating label masks.

\section{Implementation Details}
The proposed method was implemented in PyTorch \cite{paszke2019pytorch}. The configuration of LRL is introduced above, and as for the regular 3D-UNet with MC dropout, we keep the number of parameters similar to VNet in previous works \cite{wang2020double, yu2019uncertainty, luo2021semi} for fair comparison \footnote{Our regular 3D-UNet with MC dropout has 9.60M parameters while VNet has 9.44M parameters. The detailed configuration can be found in our released code.}, and the MC dropout is inserted into the decoder of the regular 3D-UNet. 
The GBDL was trained for 120 epochs on the KiTS19 and the Liver Segmentation dataset, and for 240 epochs on the Atrial Segmentation Challenge dataset. In the first half epochs, only the LRL was trained. Then, the LRL was fixed, and the regular 3D-UNet with MC dropout was optimized in the second half epochs. 
During training, each slice of CT scans was first resized to $256\times256$, and a $160\times160$ center region was cropped from the slice. Then, 
$128 \times 128 \times 32$ volumes were randomly cropped as inputs, where 32 is the depth of the input volumes, i.e., the number of slices. We used the common data augmentation methods, including pixel jittering, random rotation, and random horizontal flipping. The ADAM optimizer was used with an initial learning rate of 0.0001 and a batch size of 4, and the cosine learning rate decay strategy was used.  For the hyperparamters mentioned in the Chapter 3, namely $\lambda_1$, $\lambda_2$, $\lambda_3$, $\lambda_4$, $\beta_1$, $\beta_2$, and $M$, we fixed them to 1.0, 2.0, 1.0, 0.005, 1.0, 2.0, and 5 respectively. 
 
As for the evaluation, we used the sliding window strategy with a stride of $8 \times 8 \times 8$, and the last saved model was used for testing. We took twenty feed-forward passes for each case to get the final result and the corresponding voxel-wise epistemic uncertainty. Each of our experimental results in the paper body was obtained based on five independent runs.

\begin{figure*}[t]
 \centering
 \includegraphics[width=\linewidth]{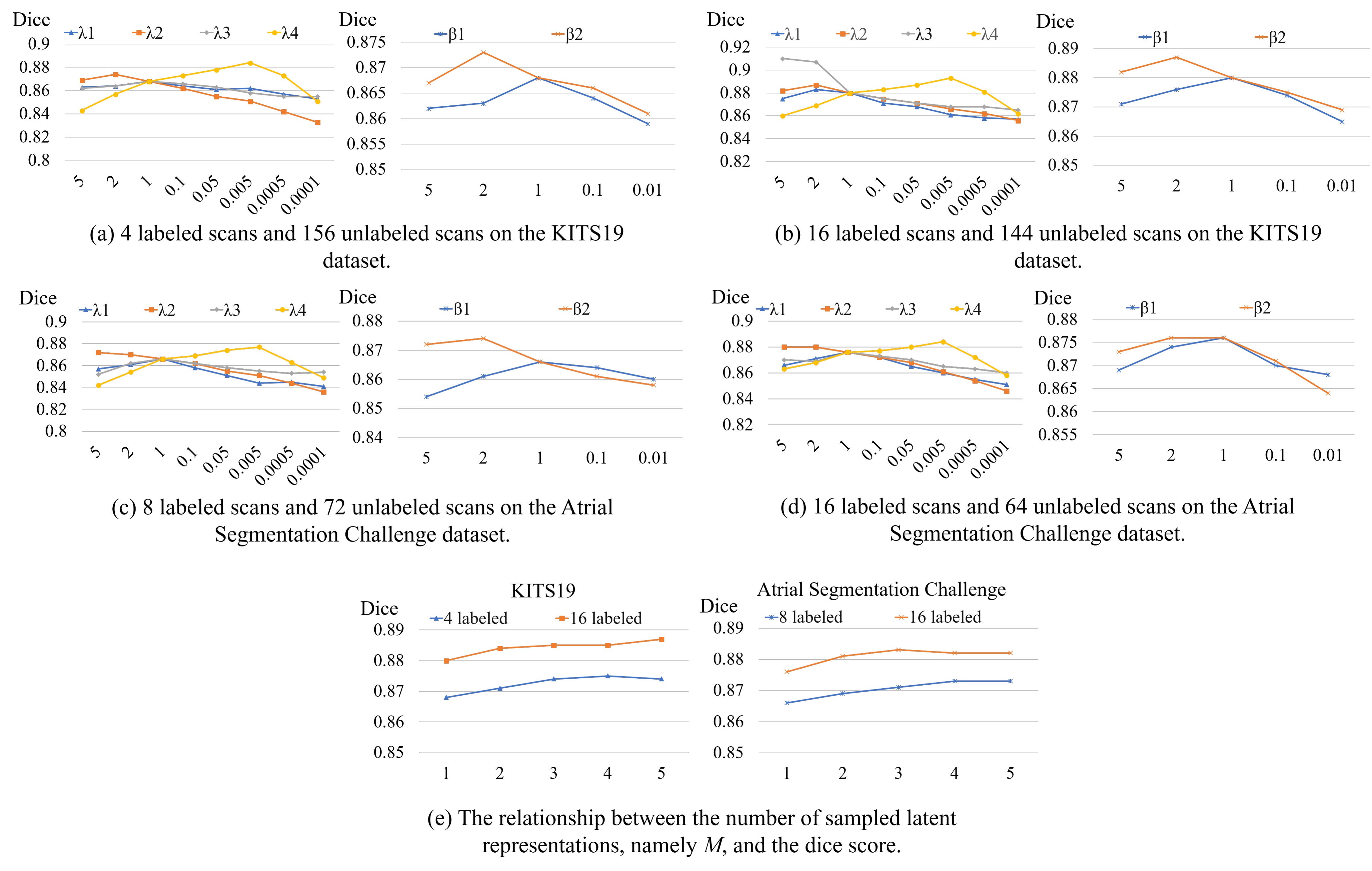}
  \vspace{-2ex}
 \caption{Ablation studies on the coefficient of different loss terms and the number of sampled latent representations. }
 \label{fig:coef} 
\end{figure*} 
 
\section{Hyperparameter Search}
We searched the hyperparameters, including coefficients of different loss terms and the number of sampled latent representations, on the KiTS19 and the Atrial Segmentation Challenge dataset. 
We further took 10 CTs and 5 CTs from the their test sets, respectively, as their validation set, in order to avoid the situation that all of the test data are involved into the search process \footnote{Note that previous works do not hold a validation set for tuning hyper-parameters. Therefore, in this work, we took a very small number of data from the test set for the hyper-parameter searching.}. Since it is hard to find the best combination of the coefficients, we resort to using the variable-controlling approach. 
Initially, all the coefficients are set to 1.0 and only one latent representation is sampled for each feed-forward pass, and then we sequentially changed them one by one. Every time we changed one variable, others were fixed to the initial setting. 
According to the results in Figure~\ref{fig:coef}, we can conclude that when $\lambda_1$, $\lambda_2$, and $\lambda_3$ are decreased, the performance declines to some extent. In  contrast, the performance is not greatly impacted when $\lambda_1$, $\lambda_2$, and $\lambda_3$ are increased, and the performance reaches the peak when $\lambda_1$, $\lambda_2$, and $\lambda_3$ are set to 1.0, 2.0, 1.0, respectively. 
Concerning  $\lambda_4$, the performance can reach the peak when it is set to 0.005. As for the coefficient of the terms in $L_{Seg}$, setting $\beta_1$ and $\beta_2$ to 1.0 and 2.0, respectively, can achieve the best result.  As for the number of sampled latent representations, we sampled $M$ latent representations to get $M$ pseudo-label maps for an unlabeled sample, and the final pseudo-label map is obtained by taking the average over the $M$ 
maps. We notice that increasing $M$  can  improve the performance, but the performance gain becomes moderate when $M$ is larger than 3.  
Then, we fixed these coefficients and set $M$ to 5 for all experiments in the paper body.

\begin{figure}[t] 
\centering
\includegraphics[width=0.7\linewidth]{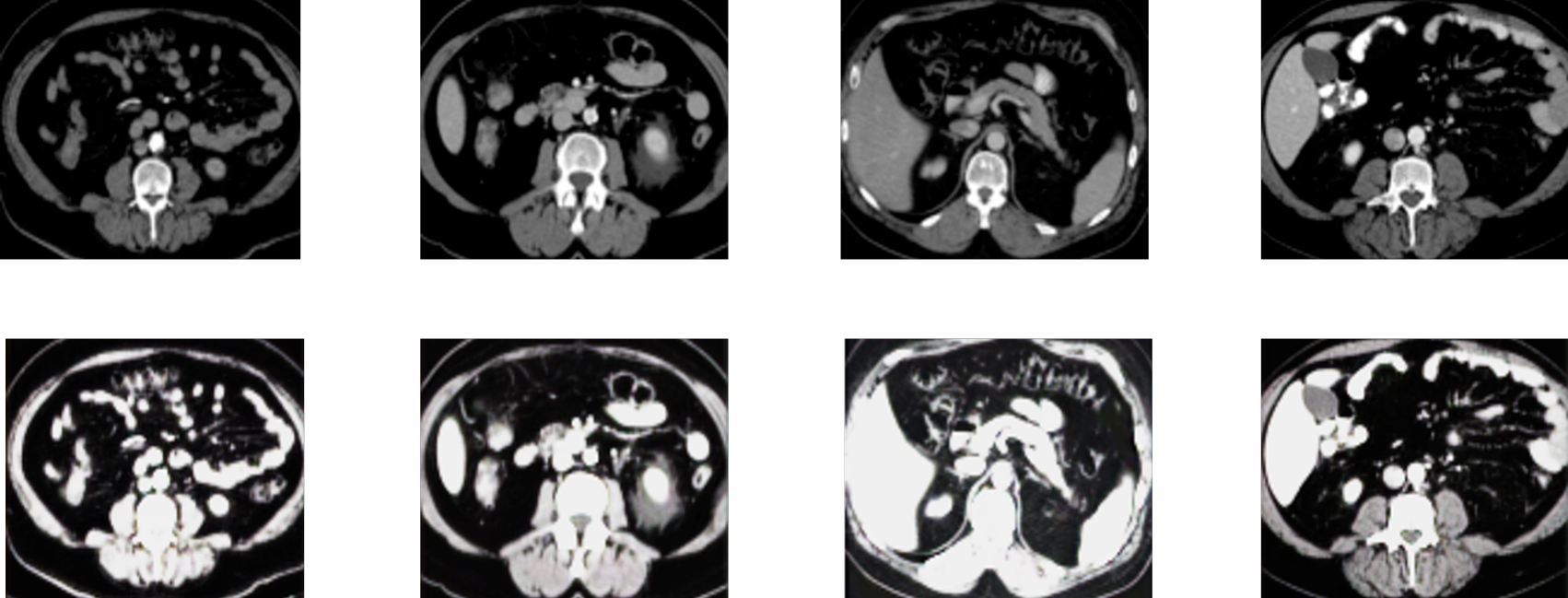} 
\caption{Reconstruction results on the KiTS19 dataset (the first row: inputs, the second row: reconstruction).} 
\label{fig:recon}
\end{figure}

\section{Input Reconstruction Results}
In Figure~\ref{fig:recon}, we randomly pick and show some reconstruction results from the GBDL framework. We can observe that, even though there may be some color differences or variations in the reconstructions, key anatomical structures are well recovered, which play an important role in segmentation tasks. This observation indicates that our GBDL successfully captures the joint distribution of $X$ and $Y$, as well as their intrinsic relationship.

%% file: appendix_npmatch.tex
\chapter{NP-Match: When Neural Processes meet Semi-Supervised Learning}

\section{Derivation of ELBO}

\emph{Proof.} As for the marginal joint distribution $p(y_{1:n} | x_{1:n})$ over $n$ data points in which there are $m$ context points and $r$ target points (i.e., $m+r=n$), we assume a variation distribution $q$, and then: 
\begin{equation}
\footnotesize		
\begin{aligned} 
 &log \ p(y_{1:n} | x_{1:n}) = log \int_z p(z, y_{1:n} | x_{1:n}) \\
 & = log \int_z \frac{p(z, y_{1:n}|x_{1:n})}{q(z|x_{m+1:\ m+r}, y_{m+1:\ m+r})}q(z|x_{m+1:\ m+r}, y_{m+1:\ m+r}) \\
 & \ge \mathbb{E}_{q(z|x_{m+1:\ m+r}, y_{m+1:\ m+r})}[log \ \frac{p(z, y_{1:n}|x_{1:n})}{q(z|x_{m+1:\ m+r}, y_{m+1:\ m+r})}] \\
 & = \mathbb{E}_{q(z|x_{m+1:\ m+r}, y_{m+1:\ m+r})}[log \ \frac{p(y_{1:m})p(z|x_{1:m}, y_{1:m})\prod^{m+r}_{i=m+1}p(y_i|z, x_i)}{q(z|x_{m+1:\ m+r}, y_{m+1:\ m+r})}] \\
 & = \mathbb{E}_{q(z|x_{m+1:\ m+r}, y_{m+1:\ m+r})}[\sum^{m+r}_{i=m+1} log \ p(y_i|z, x_i) + log \ \frac{p(z|x_{1:m}, y_{1:m})}{q(z|x_{m+1:\ m+r}, y_{m+1:\ m+r})} + log \ p(y_{1:m})] \\
 & = \mathbb{E}_{q(z|x_{m+1:\ m+r}, y_{m+1:\ m+r})}[\sum^{m+r}_{i=m+1} log \ p(y_i|z, x_i) - log \ \frac{q(z|x_{m+1:\ m+r}, y_{m+1:\ m+r})}{p(z|x_{1:m}, y_{1:m})}] + const,
\end{aligned}
\end{equation}
where ``$const$'' refers to $\mathbb{E}_{q(z|x_{m+1:\ m+r}, y_{m+1:\ m+r})}[log \ p(y_{1:m})]$, which is a constant term.  Concerning that $p(z|x_{1:m}, y_{1:m})$ is unknown, we replace it with $q(z|x_{1:m}, y_{1:m})$, and then we get:
\begin{equation}
\footnotesize
\begin{aligned} 
&log\ p(y_{1:n}|x_{1:n}) \ge \\
&\mathbb{E}_{q(z|x_{m+1:\ m+r}, y_{m+1:\ m+r})}\Big[\sum^{m+r}_{i=m+1}log\ p(y_i|z, x_i) - log\ \frac{q(z|x_{m+1:\ m+r}, y_{m+1:\ m+r})}{q(z|x_{1:m}, y_{1:m})}\Big] + const.
\end{aligned}
\end{equation}
\hfill $\square$

\section{Proof of Theorem 4.3.3.1}

\emph{Proof.} Let us first show that  $\Sigma_{\alpha_u}=((1-\alpha_u)\Sigma_1^{-1} + \alpha_u\Sigma_2^{-1})^{-1}$ and $\mu_{\alpha_u}=\Sigma_{\alpha_u}((1-\alpha_u)\Sigma_1^{-1}\mu_1 + \alpha_u\Sigma_2^{-1}\mu_2)$. Concerning two Gaussian distributions $\mathcal{N}_1(\mu_1, \Sigma_1)$ and $\mathcal{N}_2(\mu_2, \Sigma_2)$, the weighted geometric mean of them ($\mathcal{N}_1^{1-\alpha_u}\mathcal{N}_2^{\alpha_u}$) is given by:
\begin{equation}
\footnotesize
\begin{aligned}
&(2\pi)^{-\frac{D}{2}}det[\Sigma_1]^{-\frac{1-\alpha_u}{2}}det[\Sigma_2]^{-\frac{\alpha_u}{2}}e^{-\frac{1-\alpha_u}{2}(x-\mu_1)^T\Sigma_1^{-1}(x-\mu_1) - \frac{\alpha_u}{2}(x-\mu_2)^T\Sigma_2^{-1}(x-\mu_2)} \\
&=(2\pi)^{-\frac{D}{2}}det[\Sigma_1]^{-\frac{1-\alpha_u}{2}}det[\Sigma_2]^{-\frac{\alpha_u}{2}}e^{-\frac{1}{2}((x-\mu_1)^T((1-\alpha_u)\Sigma_1^{-1})(x-\mu_1) + (x-\mu_2)^T(\alpha_u\Sigma_2^{-1})(x-\mu_2))}.
\end{aligned}
\end{equation}
Now, we let $\Sigma_{1_u}^{-1} = (1-\alpha_u)\Sigma_1^{-1}$ and $\Sigma_{2_u}^{-1} = \alpha_u\Sigma_2^{-1}$, then:
\begin{equation}
\tiny		
\label{eqn:proof_2}
\begin{aligned}
&(2\pi)^{-\frac{D}{2}}det[\Sigma_1]^{-\frac{1-\alpha_u}{2}}det[\Sigma_2]^{-\frac{\alpha_u}{2}}e^{-\frac{1}{2}((x-\mu_1)^T\Sigma_{1_u}^{-1}(x-\mu_1)+(x-\mu_2)^T\Sigma_{2_u}^{-1}(x-\mu_2))} \\
&= C_1e^{-\frac{1}{2}(x^T(\Sigma_{1_u}^{-1} + \Sigma_{2_u}^{-1})x - x^T(\Sigma_{1_u}^{-1}\mu_1 + \Sigma_{2_u}^{-1}\mu_2)-(\mu_1^T\Sigma_{1_u}^{-1}+\mu_2^T\Sigma_{2_u}^{-1})x + (\mu_1^T \Sigma_{1_u}^{-1} \mu_1 + \mu_2^T \Sigma_{2_u}^{-1}\mu_2))} \\
&=  C_1e^{-\frac{1}{2}(x^T(\Sigma_{1_u}^{-1} + \Sigma_{2_u}^{-1})x - x^T(\Sigma_{1_u}^{-1} + \Sigma_{2_u}^{-1})(\Sigma_{1_u}^{-1} + \Sigma_{2_u}^{-1})^{-1}(\Sigma_{1_u}^{-1}\mu_1 + \Sigma_{2_u}^{-1}\mu_2)-(\mu_1^T\Sigma_{1_u}^{-1}+\mu_2^T\Sigma_{2_u}^{-1})x + (\mu_1^T \Sigma_{1_u}^{-1} \mu_1 + \mu_2^T \Sigma_{2_u}^{-1}\mu_2))} \\
&=  C_1e^{-\frac{1}{2}(x^T(\Sigma_{1_u}^{-1} + \Sigma_{2_u}^{-1})(x - (\Sigma_{1_u}^{-1} + \Sigma_{2_u}^{-1})^{-1}(\Sigma_{1_u}^{-1}\mu_1 + \Sigma_{2_u}^{-1}\mu_2))-(\mu_1^T\Sigma_{1_u}^{-1}+\mu_2^T\Sigma_{2_u}^{-1})x + (\mu_1^T \Sigma_{1_u}^{-1} \mu_1 + \mu_2^T \Sigma_{2_u}^{-1}\mu_2))} \\ 
&=  C_1e^{-\frac{1}{2}(x^T(\Sigma_{1_u}^{-1} + \Sigma_{2_u}^{-1})(x - (\Sigma_{1_u}^{-1} + \Sigma_{2_u}^{-1})^{-1}(\Sigma_{1_u}^{-1}\mu_1 + \Sigma_{2_u}^{-1}\mu_2))-(\mu_1^T\Sigma_{1_u}^{-1}+\mu_2^T\Sigma_{2_u}^{-1})x + (\mu_1^T\Sigma_{1_u}^{-1}+\mu_2^T\Sigma_{2_u}^{-1})(\Sigma_{1_u}^{-1} + \Sigma_{2_u}^{-1})^{-1}(\Sigma_{1_u}^{-1}\mu_1 + \Sigma_{2_u}^{-1}\mu_2) + C_2)} \\
&=  C_1e^{-\frac{1}{2}(x^T(\Sigma_{1_u}^{-1} + \Sigma_{2_u}^{-1})(x - (\Sigma_{1_u}^{-1} + \Sigma_{2_u}^{-1})^{-1}(\Sigma_{1_u}^{-1}\mu_1 + \Sigma_{2_u}^{-1}\mu_2))-(\mu_1^T\Sigma_{1_u}^{-1}+\mu_2^T\Sigma_{2_u}^{-1})(x - (\Sigma_{1_u}^{-1} + \Sigma_{2_u}^{-1})^{-1}(\Sigma_{1_u}^{-1}\mu_1 + \Sigma_{2_u}^{-1}\mu_2)) + C_2)} \\
&=  C_1e^{-\frac{1}{2}(x^T(\Sigma_{1_u}^{-1} + \Sigma_{2_u}^{-1})(x - (\Sigma_{1_u}^{-1} + \Sigma_{2_u}^{-1})^{-1}(\Sigma_{1_u}^{-1}\mu_1 + \Sigma_{2_u}^{-1}\mu_2))-(\mu_1^T\Sigma_{1_u}^{-1}+\mu_2^T\Sigma_{2_u}^{-1})(\Sigma_{1_u}^{-1} + \Sigma_{2_u}^{-1})^{-1}(\Sigma_{1_u}^{-1} + \Sigma_{2_u}^{-1})(x - (\Sigma_{1_u}^{-1} + \Sigma_{2_u}^{-1})^{-1}(\Sigma_{1_u}^{-1}\mu_1 + \Sigma_{2_u}^{-1}\mu_2)) + C_2)} \\
&=  C_1e^{-\frac{1}{2}((x^T - (\mu_1^T\Sigma_{1_u}^{-1}+\mu_2^T\Sigma_{2_u}^{-1})(\Sigma_{1_u}^{-1} + \Sigma_{2_u}^{-1})^{-1})(\Sigma_{1_u}^{-1} + \Sigma_{2_u}^{-1})(x - (\Sigma_{1_u}^{-1} + \Sigma_{2_u}^{-1})^{-1}(\Sigma_{1_u}^{-1}\mu_1 + \Sigma_{2_u}^{-1}\mu_2))+ C_2)} \\
&= C_1e^{-\frac{1}{2}((x - (\Sigma_{1_u}^{-1} + \Sigma_{2_u}^{-1})^{-1}(\Sigma_{1_u}^{-1}\mu_1 + \Sigma_{2_u}^{-1}\mu_2))^T(\Sigma_{1_u}^{-1} + \Sigma_{2_u}^{-1})(x - (\Sigma_{1_u}^{-1} + \Sigma_{2_u}^{-1})^{-1}(\Sigma_{1_u}^{-1}\mu_1 + \Sigma_{2_u}^{-1}\mu_2))+ C_2)} \\
&= C_3e^{-\frac{1}{2}(x - (\Sigma_{1_u}^{-1} + \Sigma_{2_u}^{-1})^{-1}(\Sigma_{1_u}^{-1}\mu_1 + \Sigma_{2_u}^{-1}\mu_2))^T(\Sigma_{1_u}^{-1} + \Sigma_{2_u}^{-1})(x - (\Sigma_{1_u}^{-1} + \Sigma_{2_u}^{-1})^{-1}(\Sigma_{1_u}^{-1}\mu_1 + \Sigma_{2_u}^{-1}\mu_2))}, 
\end{aligned}
\end{equation}
where $C_1 = (2\pi)^{-\frac{D}{2}}det[\Sigma_1]^{-\frac{1-\alpha_u}{2}}det[\Sigma_2]^{-\frac{\alpha_u}{2}}$, $C_2$ is a constant for aborting the terms used for completing the square relative to $x$, and $C_3 = C_1e^{-\frac{1}{2}C_2}$.
The last formula of Eq.~(\ref{eqn:proof_2}) is an unnormalized Gaussian curve with covariance $(\Sigma_{1_u}^{-1} + \Sigma_{2_u}^{-1})^{-1}$ and mean $(\Sigma_{1_u}^{-1} + \Sigma_{2_u}^{-1})^{-1}(\Sigma_{1_u}^{-1}\mu_1 + \Sigma_{2_u}^{-1}\mu_2)$. Therefore, we can get $\Sigma_{\alpha_u}=((1-\alpha_u)\Sigma_1^{-1} + \alpha_u\Sigma_2^{-1})^{-1}$ and $\mu_{\alpha_u}=\Sigma_{\alpha_u}((1-\alpha_u)\Sigma_1^{-1}\mu_1 + \alpha_u\Sigma_2^{-1}\mu_2)$. After the normalization step, we can get a Gaussian distribution $\mathcal{N}_{\alpha_u}(\mu_{\alpha_u}, \Sigma_{\alpha_u})$.

\smallskip
As for $JS^{G_{\alpha_u}}$, we first calculate $\mathbb{E}_{\mathcal{N}_1}[log{\mathcal{N}_1} - log{\mathcal{N}_{\alpha_u}}]$ as follows:
\begin{equation}
\footnotesize	 
\begin{aligned}
&\mathbb{E}_{\mathcal{N}_1}[log{\mathcal{N}_1} - log{\mathcal{N}_{\alpha_u}}] \\
&= \frac{1}{2} \mathbb{E}_{\mathcal{N}_1}[-logdet[\Sigma_1] - (x-\mu_1)^T\Sigma_1^{-1}(x-\mu_1) + logdet[\Sigma_{\alpha_u}] + (x-\mu_{\alpha_u})^T\Sigma_{\alpha_u}^{-1}(x-\mu_{\alpha_u}) ]  \\
&= \frac{1}{2} ( log\frac{det[\Sigma_{\alpha_u}]}{det[\Sigma_1]} + \mathbb{E}_{\mathcal{N}_1}[ - (x-\mu_1)^T\Sigma_1^{-1}(x-\mu_1) +  (x-\mu_{\alpha_u})^T\Sigma_{\alpha_u}^{-1}(x-\mu_{\alpha_u}) ] ) \\
&= \frac{1}{2} ( log\frac{det[\Sigma_{\alpha_u}]}{det[\Sigma_1]} + \mathbb{E}_{\mathcal{N}_1}[ - tr[\Sigma_1^{-1}\Sigma_1] +  tr[\Sigma_{\alpha_u}^{-1}(xx^T-2x\mu_{\alpha_u}^T+\mu_{\alpha_u}\mu_{\alpha_u}^T)] ] ) \\
&= \frac{1}{2} log\frac{det[\Sigma_{\alpha_u}]}{det[\Sigma_1]} - \frac{D}{2}  + \frac{1}{2} \mathbb{E}_{\mathcal{N}_1}[  tr[\Sigma_{\alpha_u}^{-1}(xx^T-2x\mu_{\alpha_u}^T+\mu_{\alpha_u}\mu_{\alpha_u}^T)] ]  \\
&= \frac{1}{2} log\frac{det[\Sigma_{\alpha_u}]}{det[\Sigma_1]} - \frac{D}{2}  + \frac{1}{2} \mathbb{E}_{\mathcal{N}_1}[  tr[\Sigma_{\alpha_u}^{-1}((x-\mu_1)(x-\mu_1)^T + 2\mu_1x^T-\mu_1\mu_1^T-2x\mu_{\alpha_u}^T+\mu_{\alpha_u}\mu_{\alpha_u}^T)] ]  \\
&= \frac{1}{2} log\frac{det[\Sigma_{\alpha_u}]}{det[\Sigma_1]} - \frac{D}{2}  + \frac{1}{2}  tr[\Sigma_{\alpha_u}^{-1}(\Sigma_1 + \mu_1\mu_1^T - 2\mu_{\alpha_u}\mu_1^T+\mu_{\alpha_u}\mu_{\alpha_u}^T)]  \\
&= \frac{1}{2} log\frac{det[\Sigma_{\alpha_u}]}{det[\Sigma_1]} - \frac{D}{2}  + \frac{1}{2}  tr[\Sigma_{\alpha_u}^{-1}\Sigma_1] + \frac{1}{2} tr[\mu_1^T\Sigma_{\alpha_u}^{-1}\mu_1 - 2\mu_1^T\Sigma_{\alpha_u}^{-1}\mu_{\alpha_u}+\mu_{\alpha_u}^T\Sigma_{\alpha_u}^{-1}\mu_{\alpha_u})]  \\
&=  \frac{1}{2} log\frac{det[\Sigma_{\alpha_u}]}{det[\Sigma_1]} - \frac{D}{2}  + \frac{1}{2}  tr[\Sigma_{\alpha_u}^{-1}\Sigma_1] + \frac{1}{2} (\mu_{\alpha_u} - \mu_1)^T\Sigma_{\alpha_u}^{-1}(\mu_{\alpha_u} - \mu_1).
\end{aligned}
\end{equation}

The calculation of $\mathbb{E}_{\mathcal{N}_2}[log{\mathcal{N}_2} - log{\mathcal{N}_{\alpha_u}}]$ is the same, and then, $JS^{G_{\alpha_u}}$ is given by:
\begin{equation}
\footnotesize		 
\begin{aligned}
JS^{G_{\alpha_u}} 
= & \frac{1-{\alpha_u}}{2} log\frac{det[\Sigma_{\alpha_u}]}{det[\Sigma_1]} - \frac{D(1-{\alpha_u})}{2}  + \frac{1-{\alpha_u}}{2}  tr[\Sigma_{\alpha_u}^{-1}\Sigma_1] + \frac{1-\alpha_u}{2} (\mu_{\alpha_u} - \mu_1)^T\Sigma_{\alpha_u}^{-1}(\mu_{\alpha_u} - \mu_1) + \\
&\frac{\alpha_u}{2} log\frac{det[\Sigma_{\alpha_u}]}{det[\Sigma_2]} - \frac{D\alpha_u}{2}  + \frac{\alpha_u}{2}  tr[\Sigma_{\alpha_u}^{-1}\Sigma_2] + \frac{\alpha_u}{2} (\mu_{\alpha_u} - \mu_2)^T\Sigma_{\alpha_u}^{-1}(\mu_{\alpha_u} - \mu_2)   \\
= & \frac{1}{2}(log\frac{det[\Sigma_{\alpha_u}]^{1-\alpha_u}}{det[\Sigma_1]^{1-\alpha_u}} + log\frac{det[\Sigma_{\alpha_u}]^{\alpha_u}}{det[\Sigma_2]^{\alpha_u}}) - \frac{D}{2} + \frac{1}{2}tr(\Sigma^{-1}_{\alpha_u}((1-\alpha_u)\Sigma_1+\alpha_u\Sigma_2)) + \\
& \frac{1-\alpha_u}{2} (\mu_{\alpha_u} - \mu_1)^T\Sigma_{\alpha_u}^{-1}(\mu_{\alpha_u} - \mu_1) +  \frac{\alpha_u}{2} (\mu_{\alpha_u} - \mu_2)^T\Sigma_{\alpha_u}^{-1}(\mu_{\alpha_u} - \mu_2) \\
=& \frac{1}{2}(log[\frac{det[\Sigma_{\alpha_u}]}{det[\Sigma_1]^{1-\alpha_u} det[\Sigma_2]^{\alpha_u}}] - D + tr(\Sigma^{-1}_{\alpha_u}((1-\alpha_u)\Sigma_1+\alpha_u\Sigma_2))+ \\
&(1-\alpha_u) (\mu_{\alpha_u} - \mu_1)^T\Sigma_{\alpha_u}^{-1}(\mu_{\alpha_u} - \mu_1) + 
  \alpha_u (\mu_{\alpha_u} - \mu_2)^T\Sigma_{\alpha_u}^{-1}(\mu_{\alpha_u} - \mu_2)).
\end{aligned}
\end{equation}

As to the dual form $JS_*^{G_{\alpha_u}}$, we calculate $\mathbb{E}_{\mathcal{N}_{\alpha_u}}[log{\mathcal{N}_{\alpha_u}} - log{\mathcal{N}_1}]$, which is given by:
\begin{equation}
\footnotesize
\begin{aligned}
\frac{1}{2} log\frac{det[\Sigma_1]}{det[\Sigma_{\alpha_u}]} - \frac{D}{2}  + \frac{1}{2}  tr[\Sigma_1^{-1}\Sigma_{\alpha_u}] + \frac{1}{2} ( \mu_1-\mu_{\alpha_u})^T\Sigma_1^{-1}(\mu_1-\mu_{\alpha_u}).
\end{aligned}
\end{equation}
Then, the calculation of $\mathbb{E}_{\mathcal{N}_{\alpha_u}}[log{\mathcal{N}_{\alpha_u}} - log{\mathcal{N}_2}]$ is the same, and $JS_*^{G_{\alpha_u}}$ is given by:
\begin{equation}
\footnotesize		 
\begin{aligned}
JS_*^{G_{\alpha_u}} 
=&  \frac{1-\alpha_u}{2} log\frac{det[\Sigma_1]}{det[\Sigma_{\alpha_u}]} - \frac{D(1-\alpha_u)}{2}  + \frac{1-\alpha_u}{2}  tr[\Sigma_1^{-1}\Sigma_{\alpha_u}] + \frac{1-\alpha_u}{2} ( \mu_1-\mu_{\alpha_u})^T\Sigma_1^{-1}(\mu_1-\mu_{\alpha_u}) \\
& + \frac{\alpha_u}{2} log\frac{det[\Sigma_2]}{det[\Sigma_{\alpha_u}]} - \frac{D\alpha_u}{2}  + \frac{\alpha_u}{2}  tr[\Sigma_2^{-1}\Sigma_{\alpha_u}] + \frac{\alpha_u}{2} ( \mu_2-\mu_{\alpha_u})^T\Sigma_2^{-1}(\mu_2-\mu_{\alpha_u}) \\
=&  \frac{1}{2} (log\frac{det[\Sigma_1]^{1-\alpha_u}}{det[\Sigma_{\alpha_u}]^{1-\alpha_u}} + log\frac{det[\Sigma_2]^{\alpha_u}}{det[\Sigma_{\alpha_u}]^{\alpha_u}}) - \frac{D}{2} + \frac{1}{2}tr((1-\alpha_u)\Sigma_1^{-1}\Sigma_{\alpha_u}+\alpha_u\Sigma_2^{-1}\Sigma_{\alpha_u})+\\
& \frac{1-\alpha_u}{2}\mu_1^T\Sigma_1^{-1}\mu_1 - \frac{1-\alpha_u}{2}\mu_1^T\Sigma_1^{-1}\mu_{\alpha_u} - \frac{1-\alpha_u}{2}\mu_{\alpha_u}^T\Sigma_1^{-1}\mu_1 +
\frac{1-\alpha_u}{2}\mu_{\alpha_u}^T\Sigma_1^{-1}\mu_{\alpha_u}
+\frac{\alpha_u}{2}\mu_2^T\Sigma_2^{-1}\mu_2 
  \\
&  - \frac{\alpha_u}{2}\mu_2^T\Sigma_2^{-1}\mu_{\alpha_u} - \frac{\alpha_u}{2}\mu_{\alpha_u}^T\Sigma_2^{-1}\mu_2 + \frac{\alpha_u}{2}\mu_{\alpha_u}^T\Sigma_2^{-1}\mu_{\alpha_u} \\
=&  \frac{1}{2}log\frac{det[\Sigma_1]^{1-\alpha_u}det[\Sigma_2]^{\alpha_u}}{det[\Sigma_{\alpha_u}]} - \frac{D}{2} + \frac{1}{2}tr(\underbrace{((1-\alpha_u)\Sigma_1^{-1}+\alpha_u\Sigma_2^{-1})}_{\Sigma^{-1}_{\alpha_u}}\Sigma_{\alpha_u}) + \frac{1-\alpha_u}{2}\mu_1^T\Sigma_1^{-1}\mu_1 - \\
& (1-\alpha_u)\mu_1^T\Sigma_1^{-1}\mu_{\alpha_u} + \frac{1-\alpha_u}{2}\mu_{\alpha_u}^T\Sigma_1^{-1}\mu_{\alpha_u} + \frac{\alpha_u}{2}\mu_2^T\Sigma_2^{-1}\mu_2 - 
 \alpha_u\mu_2^T\Sigma_2^{-1}\mu_{\alpha_u} + \frac{\alpha_u}{2}\mu_{\alpha_u}^T\Sigma_2^{-1}\mu_{\alpha_u} \\
=&  \frac{1}{2}log\frac{det[\Sigma_1]^{1-\alpha_u}det[\Sigma_2]^{\alpha_u}}{det[\Sigma_{\alpha_u}]} + \frac{1-\alpha_u}{2}\mu_1^T\Sigma_1^{-1}\mu_1 + \frac{\alpha_u}{2}\mu_2^T\Sigma_2^{-1}\mu_2 -  \\
&\underbrace{((1-\alpha_u)\mu_1^T\Sigma_1^{-1} + \alpha_u\mu_2^T\Sigma_2^{-1})}_{\mu^T_{\alpha_u}\Sigma_{\alpha_u}^{-1}}\mu_{\alpha_u} +\frac{1}{2}\mu_{\alpha_u}^T\underbrace{((1-\alpha_u)\Sigma^{-1}_1+\alpha_u\Sigma^{-1}_2)}_{\Sigma^{-1}_{\alpha_u}}\mu_{\alpha_u} \\
=&  \frac{1}{2}(log\frac{det[\Sigma_1]^{1-\alpha_u}det[\Sigma_2]^{\alpha_u}}{det[\Sigma_{\alpha_u}]} + (1-\alpha_u)\mu_1^T\Sigma_1^{-1}\mu_1 + \alpha_u\mu_2^T\Sigma_2^{-1}\mu_2 - {\mu^T_{\alpha_u}\Sigma_{\alpha_u}^{-1}}\mu_{\alpha_u}).
\end{aligned}
\end{equation}

\hfill $\square$

\section{Implementation Details}

\subsection{Standard Semi-Supervised Image Classification}
The deep neural network configuration and training details are summarized in Table~\ref{tab:setting}. 
As for the NP-Match related hyperparameters, we set the lengths of both memory banks ($\mathcal{Q}$) to 2560. The coefficient ($\beta$) is set to 0.01, and we sample $T=10$ latent vectors for each target point. The uncertainty threshold ($\tau_u$) is set to 0.4 for CIFAR-10, CIFAR-100, and  STL-10, and it is set to 1.2 for ImageNet. NP-Match is trained by using stochastic gradient descent (SGD) with a momentum of 0.9.
The initial learning rate is set to 0.03 for CIFAR-10, CIFAR-100, and STL-10, and it is set to 0.05 for ImageNet.
The learning rate is decayed with a cosine decay
schedule \cite{loshchilov2016sgdr}, and NP-Match is trained for $2^{20}$ iterations. The MLPs used in the NP model all have two layers with $\mathcal{M}$ hidden units for each layer. For WRN, $\mathcal{M}$ is a quarter of the channel dimension of the last convolutional layer, and as for ResNet-50, $\mathcal{M}$ is equal to 256. To compete with the most recent SOTA method \cite{zhang2021flexmatch}, we followed it to use the Curriculum Pseudo Labeling (CPL) strategy in our method and UPS \cite{rizve2021defense}. We initialize each memory bank with a random vector.  

\begin{table}[h]
\centering 
\resizebox{0.8\textwidth}{!}{
\begin{tabular}{@{}c|c|c|c|c@{}}
 \toprule[1pt]
Dataset  &  CIFAR-10  & CIFAR-100  & STL-10 & ImageNet \\
 \hline  
 Model &  WRN-28-2   & WRN-28-8  & WRN-37-2  & ResNet-50  \\
\hline
 Weight Decay &  5e-4 & 1e-3  & 5e-4   & 1e-4  \\
\hline
 Batch Size (B) & \multicolumn{3}{c|}{64} & 256 \\
 \hline
 $\mu$ &  \multicolumn{3}{c|}{7} &  1 \\
 \hline
 Confidence Threshold ($\tau_c$) &  \multicolumn{3}{c|}{0.95} &  0.7  \\
  \hline
 EMA Momentum&  \multicolumn{4}{c}{0.999}  \\
  \hline
 $\lambda_u$ &  \multicolumn{4}{c}{1.0}  \\
  \bottomrule[1pt]
  \end{tabular}}
 \caption{Details of the training setting.} 
 \label{tab:setting} 
 \end{table}

We ran each label amount setting for three times using different random seeds to obtain the error bars on CIFAR-10, CIFAR-100, and STL-10, but on ImageNet, we only ran for once. GeForce GTX 1080 Ti GPUs were used for the experiments on CIFAR-10, CIFAR-100, and STL-10, while Tesla V100 SXM2 GPUs were used for the experiments on ImageNet.

\subsection{Imbalanced Semi-Supervised Image Classification}
To incorporate the NP model into the distribution-aware semantics-oriented (DASO) framework \cite{oh2022daso}, we simply replaced the linear classifier in DASO with our NP model. For the original DASO framework \cite{oh2022daso}, the linear classifier is built upon a deep neural network and aims at making predictions for inputs. The pseudo-labels from the linear classifier are chosen based on a confidence threshold ($\tau_c$). After the linear classifier is substituted, $T=10$ predictions for each target point can be obtained at first. Then, the final prediction for each target is obtained by averaging over the $T$ predictions. The pseudo-labels are selected from final predictions based on
the confidence threshold ($\tau_c$)  and uncertainty threshold ($\tau_u$), which is exactly the same as the NP-Match pipeline for generating pseudo-labels, and the thresholds are set to 0.95 and 0.3, respectively. When pseudo-labels are selected, we follow the original DASO framework to combine them with the output from the similarity classifier, and therefore, we refer this modified framework to "DASO w.~NPs", which is indeed the framework combining NP-Match with DASO for imbalanced semi-supervised image classification. We conducted our experiments based on the original DASO codebase \cite{oh2022daso} for fair comparison, and the training settings for "DASO w.~NPs" are the same as the ones in the original paper \cite{oh2022daso}. The $JS^{G_{\alpha_u}}$ term in our $L_{total}$ was added to the overall loss function of DASO with the coefficient $\beta=0.1$, and we set the coefficient of the cross-entropy loss on unlabeled data to 0.3.  As for the NP model configuration, $\mathcal{M}$ is set to one eighth of the channel dimension of the last convolutional layer. When performing logic adjustment strategy \cite{menon2021long}, we first procure $T=10$ logits (the output right before the classifier in the NP model) for an input sample, and then we calculate the final logit by averaging them, so that the logic adjustment strategy \cite{menon2021long} can be applied on the final logit. The standard deviation in our results are obtained by performing three runs with different random seeds.

\subsection{Multi-label Semi-Supervised Image Classification}
To integrate the NP model into the anti-curriculum
pseudo-labelling (ACPL) framework \cite{liu2022acpl}, we also used the NP model to substitute the linear classifier in ACPL. In the original ACPL, the objective of the linear classifier is to assign a confidence score for each sample, and the score is evaluated by an information criterion named cross distribution sample informativeness (CDSI). Then, according to the CDSI criterion, only the most informative subset of unlabeled samples is  selected and the pseudo-labels of samples within the subset are utilized. After the linear classifier is replaced with the NP model, we still leveraged the NP model to assign confidence score for each sample.  In particular, the NP model makes $T=10$ predictions for each sample, and the final confidence score can be obtained by averaging them. We also estimated the uncertainty based on these $T=10$ predictions. Note that we focus on the multi-label semi-supervised image classification task, and therefore we replaced the softmax function in the NP model with the sigmoid function, and the uncertainty is derived by calculating the variance of $T=10$ predictions, instead of entropy. Before evaluating the final score with CDSI, we followed the pipeline of NP-Match to use an uncertainty threshold ($\tau_u$), which is set to 2.4, to filter out the unlabelled data with high uncertainty. We call this modified framework  ”ACPL-NPs”, which is indeed the framework combining NP-Match with ACPL for multi-label semi-supervised image classification. As for the NP model configuration, $\mathcal{M}$ is also set to one eighth of the channel dimension of the last convolutional layer. The $JS^{G_{\alpha_u}}$ term in our $L_{total}$ is added to the overall loss function of ACPL with the coefficent $\beta=0.1$.  
For fair comparisons, all implementations are based on the public code from the original work \cite{liu2022acpl},  whose training settings are directly used in our experiments. 

%% file: appendix_npsemiseg.tex
\chapter{NP-SemiSeg: When Neural Processes meet Semi-Supervised Semantic Segmentation}

\section{Implementation Details}

NP-SemiSeg is a flexible module, and in our experiments, we evaluated it with four different segmentation frameworks, including MT \cite{tarvainen2017mean}, PS-MT \cite{liu2022perturbed}, U$^2$PL \cite{wang2022semi}, and AugSeg \cite{zhao2022augmentation}. When NP-SemiSeg is incorporated into them, we followed their original hyper-parameter settings for fair comparisons, and 
we only made the following changes due to limited computational resources. On the PASCAL VOC 2012 dataset, the training crop size is set to $480 \times 480$, and those frameworks with NP-SemiSeg are trained with 0.001 learning rate and 12 batch size.  On the Cityscapes dataset,  the training crop size is set to $580 \times 580$, and we used 0.005 learning rate and 8 batch size for training. When calculating PAvPU, we use a window size 64, and the uncertainty threshold is set to 0.4.  The encoder is ResNet-50 \cite{he2016deep} that is pre-trained on ImageNet \cite{deng2009imagenet}. 

The hyper-parameters of NP-SemiSeg include the length of each memory bank ($\mathcal{Q}$), 
the coefficient $\lambda_{kl}$,  the number of latent maps $T$. 
We followed NP-Match to set $\mathcal{Q}=2560$ for all memory banks. $T$ was set to 5 at both the training phase and the testing phase. The  coefficient $\lambda_{kl}$ is set to 0.005. The configuration of the small ConvNet and the decoder are separately shown in Tables~\ref{tab:small_convnet} and~\ref{tab:decoder}. The implementation of NP-SemiSeg is modified based on the public official source code of NP-Match \cite{wang2022np}. All experiments are conducted on GeForce RTX 3090 GPUs.

\begin{table}[H]
    \centering
    \resizebox{0.95\textwidth}{!}{
    \begin{tabular}{|c|c|}
       \hline
       Type  &  Configuration \\
      \hline 
      2D Conv &  \# In-C: 512, \# Out-C: 32, Kernel Size: $ 1\times1$, Stride: $ 1\times 1$, Padding:  $ 0 $ \\
      \hline
      InstanceNorm &  \# In-C: 32, \# Out-C: 32 \\
      \hline
      ReLU &  \# In-C: 32, \# Out-C: 32  \\
      \hline
      2D Conv &  \# In-C: 32, \# Out-C: 32, Kernel Size: $ 1\times1$, Stride: $ 1\times 1$, Padding:  $ 0 $ \\
      \hline
      InstanceNorm &  \# In-C: 32, \# Out-C: 32  \\
      \hline
      ReLU &  \# In-C: 32, \# Out-C: 32  \\
      \hline
      2D Conv &  \# In-C: 32, \# Out-C: 32, Kernel Size: $ 1\times1$, Stride: $ 1\times 1$, Padding:  $ 0 $ \\
      \hline
    \end{tabular}
    } 
    \caption{Configuration of the small ConvNet. It is used for dimensional reduction, in order to save GPU memory. ``In-C'' and ``Out-C'' denote the channel dimension of the input feature maps and the output feature maps, respectively.}

    \label{tab:small_convnet}
\end{table}

\begin{table}[H]
    \centering
    \resizebox{0.95\textwidth}{!}{
    \begin{tabular}{|c|c|}
       \hline
       Type  &  Configuration \\
      \hline 
      2D Conv &  \# In-C: 576, \# Out-C: 256, Kernel Size: $ 3\times3$, Stride: $ 1\times 1$, Padding:  $ 1\times 1 $ \\
      \hline
      InstanceNorm &  \# In-C: 256, \# Out-C: 256 \\
      \hline
      ReLU &   \# In-C: 256, \# Out-C: 256 \\
      \hline
      2D Conv &  \# In-C: 256, \# Out-C: 256, Kernel Size: $ 3\times3$, Stride: $ 1\times 1 $, Padding:  $ 1\times 1 $ \\
      \hline
      InstanceNorm &  \# In-C: 256, \# Out-C: 256 \\
      \hline
      ReLU &   \# In-C: 256, \# Out-C: 256 \\
      \hline
      2D Conv &  \# In-C: 256, \# Out-C: n$_{class}$, Kernel Size: $ 1\times1$, Stride: $ 1\times 1$, Padding:  $ 0 $ \\
      \hline
    \end{tabular}
    } 
    \caption{Configuration of the decoder. ``In-C'' and ``Out-C'' denote the channel dimension of the input feature maps and the output feature maps, respectively. ``n$_{class}$'' represents the number of classes.}

    \label{tab:decoder}
\end{table}
 
\clearpage

\section{Derivation of ELBO}

\emph{Proof.} As for the marginal joint distribution $p(y_{1:n} | x_{1:n})$ over $n$ data points in which there are $m$ context data points and $r$ target data points (i.e., $m+r=n$), we assume a variational distribution over latent variables for the target data points, namely, $q(z_{m+1:\ m+r}|x_{m+1:\ m+r}, y_{m+1:\ m+r})$. According to the i.i.d assumption, those $z_*$ are independent from each other, and we denote its integral domain as $D_z$. Then: 
\begin{equation}
\footnotesize		
\begin{aligned} 
 &log \ p(y_{1:n} | x_{1:n}) = log \int\cdots\int_{D_z} p(z_{m+1:\ m+r}, y_{1:n} | x_{1:n}) \\
  =& log \int\cdots\int_{D_z} \frac{p(z_{m+1:\ m+r}, y_{1:n}|x_{1:n})}{q(z_{m+1:\ m+r}|x_{m+1:\ m+r}, y_{m+1:\ m+r})}q(z_{m+1:\ m+r}|x_{m+1:\ m+r}, y_{m+1:\ m+r}) \\
  \ge & \sum_{i=m+1}^{m+r}  \mathbb{E}_{q(z_i|x_{m+1:\ m+r}, y_{m+1:\ m+r})}[log \ \frac{p(z_i, y_{1:n}|x_{1:n})}{q(z_i|x_{m+1:\ m+r}, y_{m+1:\ m+r})}] \\
   = & \mathbb{E}_{q(z_{m+1:\ m+r}|x_{m+1:\ m+r}, y_{m+1:\ m+r})}[log \ \frac{p(y_{1:m}|x_{1:m})p(z_{m+1:\ m+r}|x_{1:m}, y_{1:m})\prod^{m+r}_{i=m+1}p(y_i|z_i, x_i)}{q(z_{m+1:\ m+r}|x_{m+1:\ m+r}, y_{m+1:\ m+r})}] \\
   = & \mathbb{E}_{q(z_{m+1:\ m+r}|x_{m+1:\ m+r}, y_{m+1:\ m+r})}[\sum^{m+r}_{i=m+1} log \ p(y_i|z_i, x_i) + log \ \frac{p(z_{m+1:\ m+r}|x_{1:m}, y_{1:m})}{q(z_{m+1:\ m+r}|x_{m+1:\ m+r}, y_{m+1:\ m+r})} + \\
   & log \ p(y_{1:m}|x_{1:m})] \\
  = & \mathbb{E}_{q(z_{m+1:\ m+r}|x_{m+1:\ m+r}, y_{m+1:\ m+r})}[\sum^{m+r}_{i=m+1} log \ p(y_i|z_i, x_i) - log \ \frac{q(z_{m+1:\ m+r}|x_{m+1:\ m+r}, y_{m+1:\ m+r})}{p(z_{m+1:\ m+r}|x_{1:m}, y_{1:m})}]\\
  & + log \ p(y_{1:m}|x_{1:m}).
\end{aligned}
\end{equation}
Similar to NPs \cite{garnelo2018neural}, $p(z_{m+1:\ m+r}|x_{1:m}, y_{1:m})$ is unknown, we use $q(z_{m+1:\ m+r}|x_{1:m}, y_{1:m})$ to replace it, and then we get:
\begin{equation}
\footnotesize
\begin{aligned} 
&log\ p(y_{1:n}|x_{1:n}) \ge \\
& \mathbb{E}_{q(z_{m+1:\ m+r}|x_{m+1:\ m+r}, y_{m+1:\ m+r})}\Big[\sum^{m+r}_{i=m+1}log\ p(y_i|z_i, x_i) - log\ \frac{q(z_{m+1:\ m+r}|x_{m+1:\ m+r}, y_{m+1:\ m+r})}{q(z_{m+1:\ m+r}|x_{1:m}, y_{1:m})}\Big] \\
& + log \ p(y_{1:m}|x_{1:m}).
\end{aligned}
\end{equation}
\hfill $\square$

%% file: main.bbl
\begin{thebibliography}{100}

\bibitem{alonso2021semi}
Inigo Alonso, Alberto Sabater, David Ferstl, Luis Montesano, and Ana~C Murillo.
\newblock Semi-supervised semantic segmentation with pixel-level contrastive
  learning from a class-wise memory bank.
\newblock In {\em Proceedings of the IEEE/CVF International Conference on
  Computer Vision}, pages 8219--8228, 2021.

\bibitem{Aviles}
Angelica~I. Aviles-Rivero, Nicolas Papadakis, Ruoteng Li, Philip Sellars,
  Qingnan Fan, Robby~T. Tan, and Carola-Bibiane Sch{\"o}nlieb.
\newblock {GraphXNET} --- {C}hest {X}-ray classification under extreme minimal
  supervision.
\newblock In {\em International Conference on Medical Image Computing and
  Computer-Assisted Intervention}. Springer International Publishing, 2019.

\bibitem{bachman2014learning}
Philip Bachman, Ouais Alsharif, and Doina Precup.
\newblock Learning with pseudo-ensembles.
\newblock In {\em Advances in Neural Information Processing Systems}, pages
  3365--3373, 2014.

\bibitem{bai2017semi}
Wenjia Bai, Ozan Oktay, Matthew Sinclair, Hideaki Suzuki, Martin Rajchl,
  Giacomo Tarroni, Ben Glocker, Andrew King, Paul~M Matthews, and Daniel
  Rueckert.
\newblock Semi-supervised learning for network-based cardiac {MR} image
  segmentation.
\newblock In {\em International Conference on Medical Image Computing and
  Computer-Assisted Intervention}, pages 253--260. Springer, 2017.

\bibitem{berthelot2019remixmatch}
David Berthelot, Nicholas Carlini, Ekin~D Cubuk, Alex Kurakin, Kihyuk Sohn, Han
  Zhang, and Colin Raffel.
\newblock {ReMixMatch}: Semi-supervised learning with distribution alignment
  and augmentation anchoring.
\newblock In {\em International Conference on Learning Representations}, 2020.

\bibitem{berthelot2019mixmatch}
David Berthelot, Nicholas Carlini, Ian Goodfellow, Nicolas Papernot, Avital
  Oliver, and Colin Raffel.
\newblock {MixMatch}: A holistic approach to semi-supervised learning.
\newblock In {\em Advances in Neural Information Processing Systems}, 2019.

\bibitem{bortsova2019semi}
Gerda Bortsova, Florian Dubost, Laurens Hogeweg, Ioannis Katramados, and
  Marleen de~Bruijne.
\newblock Semi-supervised medical image segmentation via learning consistency
  under transformations.
\newblock In {\em International Conference on Medical Image Computing and
  Computer-Assisted Intervention}, pages 810--818. Springer, 2019.

\bibitem{bruinsma2021gaussian}
Wessel~P Bruinsma, James Requeima, Andrew~YK Foong, Jonathan Gordon, and
  Richard~E Turner.
\newblock The {G}aussian neural process.
\newblock {\em Advances in Approximate Bayesian Inference}, 2021.

\bibitem{cacciarelli2023active}
Davide Cacciarelli and Murat Kulahci.
\newblock Active learning for data streams: a survey.
\newblock {\em Machine Learning}, pages 1--55, 2023.

\bibitem{cai2021ace}
Jiarui Cai, Yizhou Wang, and Jenq-Neng Hwang.
\newblock {ACE}: Ally complementary experts for solving long-tailed recognition
  in one-shot.
\newblock In {\em Proceedings of the IEEE/CVF International Conference on
  Computer Vision}, pages 112--121, 2021.

\bibitem{cao2019learning}
Kaidi Cao, Colin Wei, Adrien Gaidon, Nikos Arechiga, and Tengyu Ma.
\newblock Learning imbalanced datasets with label-distribution-aware margin
  loss.
\newblock {\em Advances in Neural Information Processing Systems}, 32, 2019.

\bibitem{dong2018tri}
Dong-Dong Chen, Wei Wang, Wei Gao, and Zhi-Hua Zhou.
\newblock Tri-net for semi-supervised deep learning.
\newblock In {\em International Joint Conference on Artificial Intelligence},
  pages 2014--2020, 2018.

\bibitem{chen2021semisupervised}
Huaian Chen, Yi~Jin, Guoqiang Jin, Changan Zhu, and Enhong Chen.
\newblock Semi-supervised semantic segmentation by improving prediction
  confidence.
\newblock {\em IEEE Transactions on Neural Networks and Learning Systems},
  2021.

\bibitem{chen2017deeplab}
Liang-Chieh Chen, George Papandreou, Iasonas Kokkinos, Kevin Murphy, and Alan~L
  Yuille.
\newblock {DeepLab:} {S}emantic image segmentation with deep convolutional
  nets, atrous convolution, and fully connected {CRF}s.
\newblock {\em IEEE Transactions on Pattern Analysis and Machine Intelligence},
  40(4):834--848, 2017.

\bibitem{chen2021semi}
Xiaokang Chen, Yuhui Yuan, Gang Zeng, and Jingdong Wang.
\newblock Semi-supervised semantic segmentation with cross pseudo supervision.
\newblock In {\em Proceedings of the IEEE/CVF Conference on Computer Vision and
  Pattern Recognition}, pages 2613--2622, 2021.

\bibitem{coates2011analysis}
Adam Coates, Andrew Ng, and Honglak Lee.
\newblock An analysis of single-layer networks in unsupervised feature
  learning.
\newblock In {\em International Conference on Artificial Intelligence and
  Statistics}, pages 215--223. JMLR Workshop and Conference Proceedings, 2011.

\bibitem{cordts2016cityscapes}
Marius Cordts, Mohamed Omran, Sebastian Ramos, Timo Rehfeld, Markus Enzweiler,
  Rodrigo Benenson, Uwe Franke, Stefan Roth, and Bernt Schiele.
\newblock The cityscapes dataset for semantic urban scene understanding.
\newblock In {\em Proceedings of the IEEE/CVF Conference on Computer Vision and
  Pattern Recognition}, pages 3213--3223, 2016.

\bibitem{cubuk2020randaugment}
Ekin~D Cubuk, Barret Zoph, Jonathon Shlens, and Quoc~V Le.
\newblock {RandAugment}: Practical automated data augmentation with a reduced
  search space.
\newblock In {\em Proceedings of the IEEE Conference on Computer Vision and
  Pattern Recognition, Workshops}, pages 702--703, 2020.

\bibitem{cui2021parametric}
Jiequan Cui, Zhisheng Zhong, Shu Liu, Bei Yu, and Jiaya Jia.
\newblock Parametric contrastive learning.
\newblock In {\em Proceedings of the IEEE/CVF International Conference on
  Computer Vision}, pages 715--724, 2021.

\bibitem{deng2009imagenet}
Jia Deng, Wei Dong, Richard Socher, Li-Jia Li, Kai Li, and Li~Fei-Fei.
\newblock Image{N}et: {A} large-scale hierarchical image database.
\newblock In {\em Proceedings of the IEEE/CVF Conference on Computer Vision and
  Pattern Recognition}, pages 248--255, 2009.

\bibitem{desai2021learning}
Alakh Desai, Tz-Ying Wu, Subarna Tripathi, and Nuno Vasconcelos.
\newblock Learning of visual relations: The devil is in the tails.
\newblock In {\em Proceedings of the IEEE/CVF International Conference on
  Computer Vision}, pages 15404--15413, 2021.

\bibitem{devlin2018bert}
Jacob Devlin, Ming-Wei Chang, Kenton Lee, and Kristina Toutanova.
\newblock {BERT:} {P}re-training of deep bidirectional transformers for
  language understanding.
\newblock {\em Association for Computational Linguistics}, 2019.

\bibitem{everingham2010pascal}
Mark Everingham, Luc Van~Gool, Christopher~KI Williams, John Winn, and Andrew
  Zisserman.
\newblock The {PASCAL} visual object classes ({VOC}) challenge.
\newblock {\em International Journal of Computer Vision}, 88(2):303--338, 2010.

\bibitem{fang2020dmnet}
Kang Fang and Wu-Jun Li.
\newblock {DMNet}: Difference minimization network for semi-supervised
  segmentation in medical images.
\newblock In {\em International Conference on Medical Image Computing and
  Computer-Assisted Intervention}, pages 532--541. Springer, 2020.

\bibitem{feng2021exploring}
Chengjian Feng, Yujie Zhong, and Weilin Huang.
\newblock Exploring classification equilibrium in long-tailed object detection.
\newblock In {\em Proceedings of the IEEE/CVF International Conference on
  Computer Vision}, pages 3417--3426, 2021.

\bibitem{feng2023latent}
Leo Feng, Hossein Hajimirsadeghi, Yoshua Bengio, and Mohamed~Osama Ahmed.
\newblock Latent bottlenecked attentive neural processes.
\newblock In {\em International Conference on Learning Representations}, 2023.

\bibitem{flaxman2015fast}
Seth Flaxman, Andrew Wilson, Daniel Neill, Hannes Nickisch, and Alex Smola.
\newblock Fast kronecker inference in gaussian processes with non-gaussian
  likelihoods.
\newblock In {\em International Conference on Machine Learning}, pages
  607--616. PMLR, 2015.

\bibitem{french2020semi}
Geoff French, Samuli Laine, Timo Aila, Michal Mackiewicz, and Graham Finlayson.
\newblock Semi-supervised semantic segmentation needs strong, varied
  perturbations.
\newblock In {\em British Machine Vision Conference}, 2020.

\bibitem{gal2016dropout}
Yarin Gal and Zoubin Ghahramani.
\newblock Dropout as a bayesian approximation: Representing model uncertainty
  in deep learning.
\newblock In {\em International Conference on Machine Learning}, pages
  1050--1059. PMLR, 2016.

\bibitem{garnelo2018conditional}
Marta Garnelo, Dan Rosenbaum, Christopher Maddison, Tiago Ramalho, David
  Saxton, Murray Shanahan, Yee~Whye Teh, Danilo Rezende, and SM~Ali Eslami.
\newblock Conditional neural processes.
\newblock In {\em International Conference on Machine Learning}, pages
  1704--1713. PMLR, 2018.

\bibitem{garnelo2018neural}
Marta Garnelo, Jonathan Schwarz, Dan Rosenbaum, Fabio Viola, Danilo~J Rezende,
  SM~Eslami, and Yee~Whye Teh.
\newblock Neural processes.
\newblock {\em arXiv:1807.01622}, 2018.

\bibitem{gordon2019convolutional}
Jonathan Gordon, Wessel~P Bruinsma, Andrew~YK Foong, James Requeima, Yann
  Dubois, and Richard~E Turner.
\newblock Convolutional conditional neural processes.
\newblock {\em International Conference on Learning Representations}, 2020.

\bibitem{guan2022unbiased}
Dayan Guan, Jiaxing Huang, Aoran Xiao, and Shijian Lu.
\newblock Unbiased subclass regularization for semi-supervised semantic
  segmentation.
\newblock In {\em Proceedings of the IEEE/CVF Conference on Computer Vision and
  Pattern Recognition}, pages 9968--9978, 2022.

\bibitem{guan2020multi}
Qingji Guan and Yaping Huang.
\newblock Multi-label chest {X}-ray image classification via category-wise
  residual attention learning.
\newblock {\em Pattern Recognition Letters}, 130:259--266, 2020.

\bibitem{guo2017calibration}
Chuan Guo, Geoff Pleiss, Yu~Sun, and Kilian~Q Weinberger.
\newblock On calibration of modern neural networks.
\newblock In {\em International Conference on Machine Learning}, pages
  1321--1330. PMLR, 2017.

\bibitem{hang2020local}
Wenlong Hang, Wei Feng, Shuang Liang, Lequan Yu, Qiong Wang, Kup-Sze Choi, and
  Jing Qin.
\newblock Local and global structure-aware entropy regularized mean teacher
  model for {3D} left atrium segmentation.
\newblock In {\em International Conference on Medical Image Computing and
  Computer-Assisted Intervention}, pages 562--571. Springer, 2020.

\bibitem{hariharan2011semantic}
Bharath Hariharan, Pablo Arbel{\'a}ez, Lubomir Bourdev, Subhransu Maji, and
  Jitendra Malik.
\newblock Semantic contours from inverse detectors.
\newblock In {\em Proceedings of the IEEE/CVF International Conference on
  Computer Vision}, pages 991--998. IEEE, 2011.

\bibitem{he2021rethinking}
Ju~He, Adam Kortylewski, Shaokang Yang, Shuai Liu, Cheng Yang, Changhu Wang,
  and Alan Yuille.
\newblock Rethinking re-sampling in imbalanced semi-supervised learning.
\newblock {\em arXiv:2106.00209}, 2021.

\bibitem{he2016deep}
Kaiming He, Xiangyu Zhang, Shaoqing Ren, and Jian Sun.
\newblock Deep residual learning for image recognition.
\newblock In {\em Proceedings of the IEEE/CVF Conference on Computer Vision and
  Pattern Recognition}, pages 770--778, 2016.

\bibitem{he2016identity}
Kaiming He, Xiangyu Zhang, Shaoqing Ren, and Jian Sun.
\newblock Identity mappings in deep residual networks.
\newblock In {\em European Conference on Computer Vision}, pages 630--645.
  Springer, 2016.

\bibitem{hensman2015scalable}
James Hensman, Alexander Matthews, and Zoubin Ghahramani.
\newblock Scalable variational gaussian process classification.
\newblock In {\em Artificial Intelligence and Statistics}, pages 351--360.
  PMLR, 2015.

\bibitem{hermoza2020region}
Renato Hermoza, Gabriel Maicas, Jacinto~C Nascimento, and Gustavo Carneiro.
\newblock Region proposals for saliency map refinement for weakly-supervised
  disease localisation and classification.
\newblock In {\em International Conference on Medical Image Computing and
  Computer-Assisted Intervention}, pages 539--549. Springer, 2020.

\bibitem{hong2021disentangling}
Youngkyu Hong, Seungju Han, Kwanghee Choi, Seokjun Seo, Beomsu Kim, and Buru
  Chang.
\newblock Disentangling label distribution for long-tailed visual recognition.
\newblock In {\em Proceedings of the IEEE Conference on Computer Vision and
  Pattern Recognition}, pages 6626--6636, 2021.

\bibitem{hou2022batchformer}
Zhi Hou, Baosheng Yu, and Dacheng Tao.
\newblock {BatchFormer}: Learning to explore sample relationships for robust
  representation learning.
\newblock In {\em Proceedings of the IEEE Conference on Computer Vision and
  Pattern Recognition}, 2022.

\bibitem{hu2021semi}
Hanzhe Hu, Fangyun Wei, Han Hu, Qiwei Ye, Jinshi Cui, and Liwei Wang.
\newblock Semi-supervised semantic segmentation via adaptive equalization
  learning.
\newblock {\em Advances in Neural Information Processing Systems},
  34:22106--22118, 2021.

\bibitem{hu2021simple}
Zijian Hu, Zhengyu Yang, Xuefeng Hu, and Ram Nevatia.
\newblock {SimPLE}: Similar pseudo label exploitation for semi-supervised
  classification.
\newblock In {\em Proceedings of the IEEE/CVF Conference on Computer Vision and
  Pattern Recognition}, pages 15099--15108, 2021.

\bibitem{huang2017densely}
Gao Huang, Zhuang Liu, Laurens Van Der~Maaten, and Kilian~Q Weinberger.
\newblock Densely connected convolutional networks.
\newblock In {\em Proceedings of the IEEE Conference on Computer Vision and
  Pattern Recognition}, pages 4700--4708, 2017.

\bibitem{huang20213d}
Huimin Huang, Nan Zhou, Lanfen Lin, Hongjie Hu, Yutaro Iwamoto, Xian-Hua Han,
  Yen-Wei Chen, and Ruofeng Tong.
\newblock {3D Graph-S$^2$Net}: Shape-aware self-ensembling network for
  semi-supervised segmentation with bilateral graph convolution.
\newblock In {\em International Conference on Medical Image Computing and
  Computer-Assis\-ted Intervention}, pages 416--427. Springer, 2021.

\bibitem{hyun2020class}
Minsung Hyun, Jisoo Jeong, and Nojun Kwak.
\newblock Class-imbalanced semi-supervised learning.
\newblock {\em arXiv:2002.06815}, 2020.

\bibitem{jean2018semi}
Neal Jean, Sang~Michael Xie, and Stefano Ermon.
\newblock Semi-supervised deep kernel learning: Regression with unlabeled data
  by minimizing predictive variance.
\newblock In {\em Advances in Neural Information Processing Systems}, 2018.

\bibitem{jha2022neural}
Saurav Jha, Dong Gong, Xuesong Wang, Richard~E Turner, and Lina Yao.
\newblock The neural process family: Survey, applications and perspectives.
\newblock {\em arXiv:2209.00517}, 2022.

\bibitem{kang2020decoupling}
Bingyi Kang, Saining Xie, Marcus Rohrbach, Zhicheng Yan, Albert Gordo, Jiashi
  Feng, and Yannis Kalantidis.
\newblock Decoupling representation and classifier for long-tailed recognition.
\newblock In {\em International Conference on Learning Representations}, 2020.

\bibitem{ke2019dual}
Zhanghan Ke, Daoye Wang, Qiong Yan, Jimmy Ren, and Rynson~WH Lau.
\newblock Dual student: Breaking the limits of the teacher in semi-supervised
  learning.
\newblock In {\em Proceedings of the IEEE/CVF International Conference on
  Computer Vision}, pages 6728--6736, 2019.

\bibitem{kendall2017uncertainties}
Alex Kendall and Yarin Gal.
\newblock What uncertainties do we need in {B}ayesian deep learning for
  computer vision?
\newblock {\em Advances in Neural Information Processing Systems}, 2017.

\bibitem{kim2019attentive}
Hyunjik Kim, Andriy Mnih, Jonathan Schwarz, Marta Garnelo, Ali Eslami, Dan
  Rosenbaum, Oriol Vinyals, and Yee~Whye Teh.
\newblock Attentive neural processes.
\newblock {\em International Conference on Learning Representations}, 2019.

\bibitem{kim2020distribution}
Jaehyung Kim, Youngbum Hur, Sejun Park, Eunho Yang, Sung~Ju Hwang, and Jinwoo
  Shin.
\newblock Distribution aligning refinery of pseudo-label for imbalanced
  semi-supervised learning.
\newblock In {\em Advances in Neural Information Processing Systems}, pages
  14567--14579, 2020.

\bibitem{kingma2013auto}
Diederik~P Kingma and Max Welling.
\newblock Auto-encoding variational {B}ayes.
\newblock In {\em International Conference on Learning Representations}, 2014.

\bibitem{krahenbuhl2011efficient}
Philipp Kr{\"a}henb{\"u}hl and Vladlen Koltun.
\newblock Efficient inference in fully connected {CRF}s with gaussian edge
  potentials.
\newblock In {\em Advances in Neural Information Processing Systems},
  volume~24, pages 109--117, 2011.

\bibitem{krishnan2020improving}
Ranganath Krishnan and Omesh Tickoo.
\newblock Improving model calibration with accuracy versus uncertainty
  optimization.
\newblock {\em Advances in Neural Information Processing Systems}, 2020.

\bibitem{krizhevsky2009learning}
Alex Krizhevsky.
\newblock Learning multiple layers of features from tiny images.
\newblock 2009.

\bibitem{krizhevsky2012imagenet}
Alex Krizhevsky, Ilya Sutskever, and Geoffrey~E Hinton.
\newblock Image{N}et classification with deep convolutional neural networks.
\newblock In {\em Advances in Neural Information Processing Systems}, pages
  1097--1105, 2012.

\bibitem{kwon2022semi}
Donghyeon Kwon and Suha Kwak.
\newblock Semi-supervised semantic segmentation with error localization
  network.
\newblock In {\em Proceedings of the IEEE/CVF Conference on Computer Vision and
  Pattern Recognition}, pages 9957--9967, 2022.

\bibitem{laine2016temporal}
Samuli Laine and Timo Aila.
\newblock Temporal ensembling for semi-supervised learning.
\newblock In {\em International Conference on Learning Representations}, 2017.

\bibitem{laves2020calibration}
Max-Heinrich Laves, Sontje Ihler, Karl-Philipp Kortmann, and Tobias Ortmaier.
\newblock Calibration of model uncertainty for dropout variational inference.
\newblock {\em arXiv:2006.11584}, 2020.

\bibitem{lee2013pseudo}
Dong-Hyun Lee.
\newblock Pseudo-label: The simple and efficient semi-supervised learning
  method for deep neural networks.
\newblock In {\em International Conference on Learning Representations,
  Workshop on Challenges in Representation Learning}, volume~3, page 896, 2013.

\bibitem{lee2021abc}
Hyuck Lee, Seungjae Shin, and Heeyoung Kim.
\newblock {ABC}: Auxiliary balanced classifier for class-imbalanced
  semi-supervised learning.
\newblock In {\em Advances in Neural Information Processing Systems},
  volume~34, pages 7082--7094, 2021.

\bibitem{lee2020bootstrapping}
Juho Lee, Yoonho Lee, Jungtaek Kim, Eunho Yang, Sung~Ju Hwang, and Yee~Whye
  Teh.
\newblock Bootstrapping neural processes.
\newblock {\em Advances in Neural Information Processing Systems}, pages
  6606--6615, 2020.

\bibitem{li2022nested}
Jun Li, Zichang Tan, Jun Wan, Zhen Lei, and Guodong Guo.
\newblock Nested collaborative learning for long-tailed visual recognition.
\newblock In {\em Proceedings of the IEEE Conference on Computer Vision and
  Pattern Recognition}, pages 6949--6958, 2022.

\bibitem{li2021comatch}
Junnan Li, Caiming Xiong, and Steven~CH Hoi.
\newblock {CoMatch}: Semi-supervised learning with contrastive graph
  regularization.
\newblock In {\em Proceedings of the IEEE/CVF International Conference on
  Computer Vision}, pages 9475--9484, 2021.

\bibitem{li2020shape}
Shuailin Li, Chuyu Zhang, and Xuming He.
\newblock Shape-aware semi-supervised {3D} semantic segmentation for medical
  images.
\newblock In {\em International Conference on Medical Image Computing and
  Computer-Assisted Intervention}, pages 552--561. Springer, 2020.

\bibitem{li2021self}
Tianhao Li, Limin Wang, and Gangshan Wu.
\newblock Self-supervision to distillation for long-tailed visual recognition.
\newblock In {\em Proceedings of the IEEE/CVF International Conference on
  Computer Vision}, pages 630--639, 2021.

\bibitem{li2020self}
Yuexiang Li, Jiawei Chen, Xinpeng Xie, Kai Ma, and Yefeng Zheng.
\newblock Self-loop uncertainty: A novel pseudo-label for semi-supervised
  medical image segmentation.
\newblock In {\em International Conference on Medical Image Computing and
  Computer-Assisted Intervention}, pages 614--623. Springer, 2020.

\bibitem{li2018thoracic}
Zhe Li, Chong Wang, Mei Han, Yuan Xue, Wei Wei, Li-Jia Li, and Li~Fei-Fei.
\newblock Thoracic disease identification and localization with limited
  supervision.
\newblock In {\em Proceedings of the IEEE Conference on Computer Vision and
  Pattern Recognition}, pages 8290--8299, 2018.

\bibitem{liu2020semi}
Chenghao Liu, Fengqian Pang, Yanlin Liu, Kongming Liang, Xiuli Li, Xiangzhu
  Zeng, and Chuyang Ye.
\newblock Semi-supervised brain lesion segmentation using training images with
  and without lesions.
\newblock In {\em 2020 IEEE 17th International Symposium on Biomedical Imaging
  (ISBI)}, pages 279--282. IEEE, 2020.

\bibitem{liu2022acpl}
Fengbei Liu, Yu~Tian, Yuanhong Chen, Yuyuan Liu, Vasileios Belagiannis, and
  Gustavo Carneiro.
\newblock {ACPL: A}nti-curriculum pseudo-labelling for semi-supervised medical
  image classification.
\newblock In {\em Proceedings of the IEEE Conference on Computer Vision and
  Pattern Recognition}, pages 20697--20706, 2022.

\bibitem{liu2021self}
Fengbei Liu, Yu~Tian, Filipe~R Cordeiro, Vasileios Belagiannis, Ian Reid, and
  Gustavo Carneiro.
\newblock Self-supervised mean teacher for semi-supervised chest {X}-ray
  classification.
\newblock In {\em International Conference on Medical Image Computing and
  Computer-Assisted Intervention, Workshops}, pages 426--436. Springer, 2021.

\bibitem{liu2019align}
Jingyu Liu, Gangming Zhao, Yu~Fei, Ming Zhang, Yizhou Wang, and Yizhou Yu.
\newblock Align, attend and locate: Chest {X}-ray diagnosis via contrast
  induced attention network with limited supervision.
\newblock In {\em Proceedings of the IEEE/CVF International Conference on
  Computer Vision}, pages 10632--10641, 2019.

\bibitem{liu2022perturbed}
Yuyuan Liu, Yu~Tian, Yuanhong Chen, Fengbei Liu, Vasileios Belagiannis, and
  Gustavo Carneiro.
\newblock Perturbed and strict mean teachers for semi-supervised semantic
  segmentation.
\newblock In {\em Proceedings of the IEEE/CVF Conference on Computer Vision and
  Pattern Recognition}, pages 4258--4267, 2022.

\bibitem{liu2020uncertainty}
Zhao-Yang Liu, Shao-Yuan Li, Songcan Chen, Yao Hu, and Sheng-Jun Huang.
\newblock Uncertainty aware graph {G}aussian process for semi-supervised
  learning.
\newblock In {\em Proceedings of the AAAI Conference on Artificial
  Intelligence}, pages 4957--4964, 2020.

\bibitem{liu2019large}
Ziwei Liu, Zhongqi Miao, Xiaohang Zhan, Jiayun Wang, Boqing Gong, and Stella~X
  Yu.
\newblock Large-scale long-tailed recognition in an open world.
\newblock In {\em Proceedings of the IEEE Conference on Computer Vision and
  Pattern Recognition}, pages 2537--2546, 2019.

\bibitem{loshchilov2016sgdr}
Ilya Loshchilov and Frank Hutter.
\newblock {SGDR: S}tochastic gradient descent with warm restarts.
\newblock {\em arXiv:1608.03983}, 2016.

\bibitem{louizos2019functional}
Christos Louizos, Xiahan Shi, Klamer Schutte, and Max Welling.
\newblock The functional neural process.
\newblock {\em Advances in Neural Information Processing Systems}, 2019.

\bibitem{luo2021semi}
Xiangde Luo, Jieneng Chen, Tao Song, and Guotai Wang.
\newblock Semi-supervised medical image segmentation through dual-task
  consistency.
\newblock In {\em Proceedings of the AAAI Conference on Artificial
  Intelligence}, pages 8801--8809, 2021.

\bibitem{luo2021efficient}
Xiangde Luo, Wenjun Liao, Jieneng Chen, Tao Song, Yinan Chen, Shichuan Zhang,
  Nianyong Chen, Guotai Wang, and Shaoting Zhang.
\newblock Efficient semi-supervised gross target volume of nasopharyngeal
  carcinoma segmentation via uncertainty rectified pyramid consistency.
\newblock In {\em International Conference on Medical Image Computing and
  Computer-Assisted Intervention}, pages 318--329. Springer, 2021.

\bibitem{ma2019multi}
Congbo Ma, Hu~Wang, and Steven~CH Hoi.
\newblock Multi-label thoracic disease image classification with
  cross-attention networks.
\newblock In {\em International Conference on Medical Image Computing and
  Computer-Assisted Intervention}, pages 730--738. Springer, 2019.

\bibitem{menon2021long}
Aditya~Krishna Menon, Sadeep Jayasumana, Ankit~Singh Rawat, Himanshu Jain,
  Andreas Veit, and Sanjiv Kumar.
\newblock Long-tail learning via logit adjustment.
\newblock In {\em International Conference on Learning Representations}, 2021.

\bibitem{meyer2021uncertainty}
Anneke Meyer, Suhita Ghosh, Daniel Schindele, Martin Schostak, Sebastian
  Stober, Christian Hansen, and Marko Rak.
\newblock Uncertainty-aware temporal self-learning ({UATS}): Semi-supervised
  learning for segmentation of prostate zones and beyond.
\newblock {\em Artificial Intelligence in Medicine}, 116:102073, 2021.

\bibitem{milletari2016v}
Fausto Milletari, Nassir Navab, and Seyed-Ahmad Ahmadi.
\newblock {V-Net}: Fully convolutional neural networks for volumetric medical
  image segmentation.
\newblock In {\em International Conference on 3D Vision}, pages 565--571. IEEE,
  2016.

\bibitem{mukhoti2018evaluating}
Jishnu Mukhoti and Yarin Gal.
\newblock Evaluating {B}ayesian deep learning methods for semantic
  segmentation.
\newblock {\em arXiv:1811.12709}, 2018.

\bibitem{naeini2015obtaining}
Mahdi~Pakdaman Naeini, Gregory Cooper, and Milos Hauskrecht.
\newblock Obtaining well calibrated probabilities using {B}ayesian binning.
\newblock In {\em Proceedings of the AAAI Conference on Artificial
  Intelligence}, 2015.

\bibitem{nassar2021all}
Islam Nassar, Samitha Herath, Ehsan Abbasnejad, Wray Buntine, and Gholamreza
  Haffari.
\newblock All labels are not created equal: Enhancing semi-supervision via
  label grouping and co-training.
\newblock In {\em Proceedings of the IEEE/CVF Conference on Computer Vision and
  Pattern Recognition}, pages 7241--7250, 2021.

\bibitem{ng2018bayesian}
Yin~Cheng Ng, Nicol{\`o} Colombo, and Ricardo Silva.
\newblock Bayesian semi-supervised learning with graph {G}aussian processes.
\newblock In {\em Advances in Neural Information Processing Systems}, 2018.

\bibitem{nguyen2022transformer}
Tung Nguyen and Aditya Grover.
\newblock {Transformer neural processes:} {U}ncertainty-aware meta learning via
  sequence modeling.
\newblock In {\em International Conference on Machine Learning}, 2022.

\bibitem{niculescu2006convex}
Constantin Niculescu and Lars-Erik Persson.
\newblock {\em Convex functions and their applications}.
\newblock Springer, 2006.

\bibitem{nielsen2020generalization}
Frank Nielsen.
\newblock On a generalization of the {Jensen-Shannon divergence and the
  Jensen-Shannon} centroid.
\newblock {\em Entropy}, 22(2):221, 2020.

\bibitem{nielsen2009statistical}
Frank Nielsen and Vincent Garcia.
\newblock Statistical exponential families: A digest with flash cards.
\newblock {\em arXiv:0911.4863}, 2009.

\bibitem{oh2022daso}
Youngtaek Oh, Dong-Jin Kim, and In~So Kweon.
\newblock {DASO}: Distribution-aware semantics-oriented pseudo-label for
  imbalanced semi-supervised learning.
\newblock In {\em Proceedings of the IEEE Conference on Computer Vision and
  Pattern Recognition}, pages 9786--9796, 2022.

\bibitem{oksendal2003stochastic}
Bernt {\O}ksendal.
\newblock Stochastic differential equations.
\newblock In {\em Stochastic Differential Equations}, pages 65--84. Springer,
  2003.

\bibitem{oliver2018realistic}
Avital Oliver, Augustus Odena, Colin~A Raffel, Ekin~Dogus Cubuk, and Ian
  Goodfellow.
\newblock Realistic evaluation of deep semi-supervised learning algorithms.
\newblock {\em Advances in Neural Information Processing Systems}, 31, 2018.

\bibitem{ouali2020semi}
Yassine Ouali, C{\'e}line Hudelot, and Myriam Tami.
\newblock Semi-supervised semantic segmentation with cross-consistency
  training.
\newblock In {\em Proceedings of the IEEE/CVF Conference on Computer Vision and
  Pattern Recognition}, pages 12674--12684, 2020.

\bibitem{paszke2019pytorch}
Adam Paszke, Sam Gross, Francisco Massa, Adam Lerer, James Bradbury, Gregory
  Chanan, Trevor Killeen, Zeming Lin, Natalia Gimelshein, Luca Antiga, et~al.
\newblock {PyTorch:} {A}n imperative style, high-performance deep learning
  library.
\newblock In {\em Advances in Neural Information Processing Systems}, pages
  8026--8037, 2019.

\bibitem{pham2021meta}
Hieu Pham, Zihang Dai, Qizhe Xie, and Quoc~V Le.
\newblock Meta pseudo labels.
\newblock In {\em Proceedings of the IEEE/CVF Conference on Computer Vision and
  Pattern Recognition}, pages 11557--11568, 2021.

\bibitem{qiao2018deep}
Siyuan Qiao, Wei Shen, Zhishuai Zhang, Bo~Wang, and Alan Yuille.
\newblock Deep co-training for semi-supervised image recognition.
\newblock In {\em European Conference on Computer Vision}, pages 135--152,
  2018.

\bibitem{rajpurkar2017chexnet}
Pranav Rajpurkar, Jeremy Irvin, Kaylie Zhu, Brandon Yang, Hershel Mehta, Tony
  Duan, Daisy Ding, Aarti Bagul, Curtis Langlotz, Katie Shpanskaya, et~al.
\newblock {CheXNet:} {R}adiologist-level pneumonia detection on chest {X}-rays
  with deep learning.
\newblock {\em arXiv:1711.05225}, 2017.

\bibitem{gpml}
Carl~Edward Rasmussen and Christopher K.~I. Williams.
\newblock {\em Gaussian processes for machine learning}.
\newblock MIT Press, 2006.

\bibitem{rizve2021defense}
Mamshad~Nayeem Rizve, Kevin Duarte, Yogesh~S Rawat, and Mubarak Shah.
\newblock In defense of pseudo-labeling: An uncertainty-aware pseudo-label
  selection framework for semi-supervised learning.
\newblock {\em International Conference on Learning Representations}, 2021.

\bibitem{ronneberger2015u}
Olaf Ronneberger, Philipp Fischer, and Thomas Brox.
\newblock {U-Net}: Convolutional networks for biomedical image segmentation.
\newblock In {\em International Conference on Medical Image Computing and
  Computer-Assisted Intervention}, pages 234--241. Springer, 2015.

\bibitem{sajjadi2016regularization}
Mehdi Sajjadi, Mehran Javanmardi, and Tolga Tasdizen.
\newblock Regularization with stochastic transformations and perturbations for
  deep semi-supervised learning.
\newblock In {\em Advances in Neural Information Processing Systems}, pages
  1163--1171, 2016.

\bibitem{samuel2021distributional}
Dvir Samuel and Gal Chechik.
\newblock Distributional robustness loss for long-tail learning.
\newblock In {\em Proceedings of the IEEE/CVF International Conference on
  Computer Vision}, pages 9495--9504, 2021.

\bibitem{sedai2019uncertainty}
Suman Sedai, Bhavna Antony, Ravneet Rai, Katie Jones, Hiroshi Ishikawa, Joel
  Schuman, Wollstein Gadi, and Rahil Garnavi.
\newblock Uncertainty guided semi-supervised segmentation of retinal layers in
  {OCT} images.
\newblock In {\em International Conference on Medical Image Computing and
  Computer-Assisted Intervention}, pages 282--290. Springer, 2019.

\bibitem{shi2021inconsistency}
Yinghuan Shi, Jian Zhang, Tong Ling, Jiwen Lu, Yefeng Zheng, Qian Yu, Lei Qi,
  and Yang Gao.
\newblock Inconsistency-aware uncertainty estimation for semi-supervised
  medical image segmentation.
\newblock {\em IEEE Transactions on Medical Imaging}, 2021.

\bibitem{simonyan2014very}
Karen Simonyan and Andrew Zisserman.
\newblock Very deep convolutional networks for large-scale image recognition.
\newblock In {\em International Conference on Learning Representations}, 2014.

\bibitem{sindhwani2007semi}
Vikas Sindhwani, Wei Chu, and S~Sathiya Keerthi.
\newblock Semi-supervised {G}aussian process classifiers.
\newblock In {\em International Joint Conference on Artificial Intelligence},
  pages 1059--1064, 2007.

\bibitem{snelson2005sparse}
Edward Snelson and Zoubin Ghahramani.
\newblock Sparse gaussian processes using pseudo-inputs.
\newblock {\em Advances in Neural Information Processing Systems}, 18, 2005.

\bibitem{sohn2020fixmatch}
Kihyuk Sohn, David Berthelot, Chun-Liang Li, Zizhao Zhang, Nicholas Carlini,
  Ekin~D Cubuk, Alex Kurakin, Han Zhang, and Colin Raffel.
\newblock {FixMatch}: Simplifying semi-supervised learning with consistency and
  confidence.
\newblock {\em Advances in Neural Information Processing Systems}, 2020.

\bibitem{sohn2015learning}
Kihyuk Sohn, Honglak Lee, and Xinchen Yan.
\newblock Learning structured output representation using deep conditional
  generative models.
\newblock In {\em Advances in Neural Information Processing Systems},
  volume~28, pages 3483--3491, 2015.

\bibitem{srivastava2014dropout}
Nitish Srivastava, Geoffrey Hinton, Alex Krizhevsky, Ilya Sutskever, and Ruslan
  Salakhutdinov.
\newblock Dropout: A simple way to prevent neural networks from overfitting.
\newblock {\em The Journal of Machine Learning Research}, 15(1):1929--1958,
  2014.

\bibitem{szegedy2015going}
Christian Szegedy, Wei Liu, Yangqing Jia, Pierre Sermanet, Scott Reed, Dragomir
  Anguelov, Dumitru Erhan, Vincent Vanhoucke, and Andrew Rabinovich.
\newblock Going deeper with convolutions.
\newblock In {\em Proceedings of the IEEE/CVF Conference on Computer Vision and
  Pattern Recognition}, pages 1--9, 2015.

\bibitem{taghanaki2019infomask}
Saeid~Asgari Taghanaki, Mohammad Havaei, Tess Berthier, Francis Dutil, Lisa
  Di~Jorio, Ghassan Hamarneh, and Yoshua Bengio.
\newblock Infomask: Masked variational latent representation to localize chest
  disease.
\newblock In {\em International Conference on Medical Image Computing and
  Computer-Assisted Intervention}, pages 739--747. Springer, 2019.

\bibitem{tarvainen2017mean}
Antti Tarvainen and Harri Valpola.
\newblock Mean teachers are better role models: Weight-averaged consistency
  targets improve semi-supervised deep learning results.
\newblock {\em Advances in Neural Information Processing Systems}, 30, 2017.

\bibitem{teye2018bayesian}
Mattias Teye, Hossein Azizpour, and Kevin Smith.
\newblock Bayesian uncertainty estimation for batch normalized deep networks.
\newblock In {\em International Conference on Machine Learning}, pages
  4907--4916. PMLR, 2018.

\bibitem{ulyanov2017improved}
Dmitry Ulyanov, Andrea Vedaldi, and Victor Lempitsky.
\newblock Improved texture networks: Maximizing quality and diversity in
  feed-forward stylization and texture synthesis.
\newblock In {\em Proceedings of the IEEE Conference on Computer Vision and
  Pattern Recognition}, pages 6924--6932, 2017.

\bibitem{van2020survey}
Jesper~E Van~Engelen and Holger~H Hoos.
\newblock A survey on semi-supervised learning.
\newblock 109(2):373--440, 2020.

\bibitem{walker2019graph}
Ian Walker and Ben Glocker.
\newblock Graph convolutional {G}aussian processes.
\newblock In {\em International Conference on Machine Learning}, pages
  6495--6504. PMLR, 2019.

\bibitem{wan2013regularization}
Li~Wan, Matthew Zeiler, Sixin Zhang, Yann Le~Cun, and Rob Fergus.
\newblock Regularization of neural networks using dropconnect.
\newblock In {\em International Conference on Machine Learning}, pages
  1058--1066. PMLR, 2013.

\bibitem{wang2023np1}
Jianfeng Wang, Xiaolin Hu, and Thomas Lukasiewicz.
\newblock {NP-Match: T}owards a new probabilistic model for semi-supervised
  learning.
\newblock {\em arXiv:2301.13569}, 2023.

\bibitem{wang2022rethinking}
Jianfeng Wang and Thomas Lukasiewicz.
\newblock Rethinking bayesian deep learning methods for semi-supervised
  volumetric medical image segmentation.
\newblock In {\em Proceedings of the IEEE/CVF Conference on Computer Vision and
  Pattern Recognition}, pages 182--190, 2022.

\bibitem{wang2021rsg}
Jianfeng Wang, Thomas Lukasiewicz, Xiaolin Hu, Jianfei Cai, and Zhenghua Xu.
\newblock {RSG}: A simple but effective module for learning imbalanced
  datasets.
\newblock In {\em Proceedings of the IEEE Conference on Computer Vision and
  Pattern Recognition}, pages 3784--3793, 2021.

\bibitem{wang2022np}
Jianfeng Wang, Thomas Lukasiewicz, Daniela Massiceti, Xiaolin Hu, Vladimir
  Pavlovic, and Alexandros Neophytou.
\newblock {NP-Match}: When neural processes meet semi-supervised learning.
\newblock In {\em International Conference on Machine Learning}, pages
  22919--22934. PMLR, 2022.

\bibitem{wang2023np}
Jianfeng Wang, Daniela Massiceti, Xiaolin Hu, Vladimir Pavlovic, and Thomas
  Lukasiewicz.
\newblock {NP-SemiSeg: W}hen neural processes meet semi-supervised semantic
  segmentation.
\newblock In {\em International Conference on Machine Learning}. PMLR, 2023.

\bibitem{wang2021tripled}
Kaiping Wang, Bo~Zhan, Chen Zu, Xi~Wu, Jiliu Zhou, Luping Zhou, and Yan Wang.
\newblock Tripled-uncertainty guided mean teacher model for semi-supervised
  medical image segmentation.
\newblock In {\em International Conference on Medical Image Computing and
  Computer-Assisted Intervention}, pages 450--460. Springer, 2021.

\bibitem{wang2020enaet}
Xiao Wang, Daisuke Kihara, Jiebo Luo, and Guo-Jun Qi.
\newblock {EnAET: A} self-trained framework for semi-supervised and supervised
  learning with ensemble transformations.
\newblock {\em IEEE Transactions on Image Processing}, 30:1639--1647, 2020.

\bibitem{wang2017chestx}
Xiaosong Wang, Yifan Peng, Le~Lu, Zhiyong Lu, Mohammadhadi Bagheri, and
  Ronald~M Summers.
\newblock Chest{X}-ray8: Hospital-scale chest {X}-ray database and benchmarks
  on weakly-supervised classification and localization of common thorax
  diseases.
\newblock In {\em Proceedings of the IEEE Conference on Computer Vision and
  Pattern Recognition}, pages 2097--2106, 2017.

\bibitem{wang2020double}
Yixin Wang, Yao Zhang, Jiang Tian, Cheng Zhong, Zhongchao Shi, Yang Zhang, and
  Zhiqiang He.
\newblock Double-un\-cer\-tainty weighted method for semi-supervised learning.
\newblock In {\em International Conference on Medical Image Computing and
  Computer-Assisted Intervention}, pages 542--551. Springer, 2020.

\bibitem{wang2022semi}
Yuchao Wang, Haochen Wang, Yujun Shen, Jingjing Fei, Wei Li, Guoqiang Jin,
  Liwei Wu, Rui Zhao, and Xinyi Le.
\newblock Semi-supervised semantic segmentation using unreliable pseudo-labels.
\newblock In {\em Proceedings of the IEEE/CVF Conference on Computer Vision and
  Pattern Recognition}, pages 4248--4257, 2022.

\bibitem{wei2021crest}
Chen Wei, Kihyuk Sohn, Clayton Mellina, Alan Yuille, and Fan Yang.
\newblock {CReST: A} class-rebalancing self-training framework for imbalanced
  semi-supervised learning.
\newblock In {\em Proceedings of the IEEE Conference on Computer Vision and
  Pattern Recognition}, pages 10857--10866, 2021.

\bibitem{wilson2016deep}
Andrew~Gordon Wilson, Zhiting Hu, Ruslan Salakhutdinov, and Eric~P Xing.
\newblock Deep kernel learning.
\newblock In {\em International Conference on Artificial Intelligence and
  Statistics}, pages 370--378. PMLR, 2016.

\bibitem{xiang2022fussnet}
Jinyi Xiang, Peng Qiu, and Yang Yang.
\newblock {FUSSNet}: Fusing two sources of uncertainty for semi-supervised
  medical image segmentation.
\newblock In {\em International Conference on Medical Image Computing and
  Computer-Assisted Intervention}, pages 481--491. Springer, 2022.

\bibitem{xiao2023delving}
Junfei Xiao, Yutong Bai, Alan Yuille, and Zongwei Zhou.
\newblock Delving into masked autoencoders for multi-label thorax disease
  classification.
\newblock In {\em Proceedings of the IEEE/CVF Winter Conference on Applications
  of Computer Vision}, pages 3588--3600, 2023.

\bibitem{xie2019unsupervised}
Qizhe Xie, Zihang Dai, Eduard Hovy, Minh-Thang Luong, and Quoc~V Le.
\newblock Unsupervised data augmentation for consistency training.
\newblock In {\em Advances in Neural Information Processing Systems}, 2020.

\bibitem{xie2020pairwise}
Yutong Xie, Jianpeng Zhang, Zhibin Liao, Johan Verjans, Chunhua Shen, and Yong
  Xia.
\newblock Pairwise relation learning for semi-supervised gland segmentation.
\newblock In {\em International Con\-ference on Medical Image Computing and
  Computer-Assis\-ted Intervention}, pages 417--427. Springer, 2020.

\bibitem{xiong2020global}
Zhaohan Xiong, Qing Xia, Zhiqiang Hu, Ning Huang, Cheng Bian, Yefeng Zheng,
  Sulaiman Vesal, Nishant Ravikumar, Andreas Maier, Xin Yang, et~al.
\newblock A global benchmark of algorithms for segmenting the left atrium from
  late gadolinium-enhanced cardiac magnetic resonance imaging.
\newblock {\em Medical Image Analysis}, 67, 2020.

\bibitem{yang2022st++}
Lihe Yang, Wei Zhuo, Lei Qi, Yinghuan Shi, and Yang Gao.
\newblock {ST++}: Make self-training work better for semi-supervised semantic
  segmentation.
\newblock In {\em Proceedings of the IEEE/CVF Conference on Computer Vision and
  Pattern Recognition}, pages 4268--4277, 2022.

\bibitem{yang2020rethinking}
Yuzhe Yang and Zhi Xu.
\newblock Rethinking the value of labels for improving class-imbalanced
  learning.
\newblock In {\em Advances in Neural Information Processing Systems}, pages
  19290--19301, 2020.

\bibitem{yasarla2020syn2real}
Rajeev Yasarla, Vishwanath~A Sindagi, and Vishal~M Patel.
\newblock {Syn2Real} transfer learning for image deraining using {G}aussian
  processes.
\newblock In {\em Proceedings of the IEEE Conference on Computer Vision and
  Pattern Recognition}, pages 2726--2736, 2020.

\bibitem{yu2019uncertainty}
Lequan Yu, Shujun Wang, Xiaomeng Li, Chi-Wing Fu, and Pheng-Ann Heng.
\newblock Uncertainty-aware self-ensembling model for semi-supervised {3D} left
  atrium segmentation.
\newblock In {\em International Conference on Medical Image Computing and
  Computer-Assisted Intervention}, pages 605--613. Springer, 2019.

\bibitem{zagoruyko2016wide}
Sergey Zagoruyko and Nikos Komodakis.
\newblock Wide residual networks.
\newblock {\em British Machine Vision Conference}, 2016.

\bibitem{zaheer2017deep}
Manzil Zaheer, Satwik Kottur, Siamak Ravanbakhsh, Barnabas Poczos, Russ~R
  Salakhutdinov, and Alexander~J Smola.
\newblock Deep sets.
\newblock {\em Advances in Neural Information Processing Systems}, 30, 2017.

\bibitem{zeng2021reciprocal}
Xiangyun Zeng, Rian Huang, Yuming Zhong, Dong Sun, Chu Han, Di~Lin, Dong Ni,
  and Yi~Wang.
\newblock Reciprocal learning for semi-supervised segmentation.
\newblock In {\em International Conference on Medical Image Computing and
  Computer-Assisted Intervention}, pages 352--361. Springer, 2021.

\bibitem{zhai2019s4l}
Xiaohua Zhai, Avital Oliver, Alexander Kolesnikov, and Lucas Beyer.
\newblock {S4L: S}elf-supervised semi-supervised learning.
\newblock In {\em Proceedings of the IEEE/CVF International Conference on
  Computer Vision}, pages 1476--1485, 2019.

\bibitem{zhang2021flexmatch}
Bowen Zhang, Yidong Wang, Wenxin Hou, Hao Wu, Jindong Wang, Manabu Okumura, and
  Takahiro Shinozaki.
\newblock {FlexMatch}: Boosting semi-supervised learning with curriculum pseudo
  labeling.
\newblock {\em Advances in Neural Information Processing Systems}, 34, 2021.

\bibitem{zhang2018survey}
Qingchen Zhang, Laurence~T Yang, Zhikui Chen, and Peng Li.
\newblock A survey on deep learning for big data.
\newblock {\em Information Fusion}, 42:146--157, 2018.

\bibitem{zhao2022augmentation}
Zhen Zhao, Lihe Yang, Sifan Long, Jimin Pi, Luping Zhou, and Jingdong Wang.
\newblock Augmentation matters: A simple-yet-effective approach to
  semi-supervised semantic segmentation.
\newblock In {\em Proceedings of the IEEE/CVF Conference on Computer Vision and
  Pattern Recognition}, 2023.

\bibitem{zhong2021pixel}
Yuanyi Zhong, Bodi Yuan, Hong Wu, Zhiqiang Yuan, Jian Peng, and Yu-Xiong Wang.
\newblock Pixel contrastive-consistent semi-supervised semantic segmentation.
\newblock In {\em Proceedings of the IEEE/CVF International Conference on
  Computer Vision}, pages 7273--7282, 2021.

\bibitem{zhong2021improving}
Zhisheng Zhong, Jiequan Cui, Shu Liu, and Jiaya Jia.
\newblock Improving calibration for long-tailed recognition.
\newblock In {\em Proceedings of the IEEE Conference on Computer Vision and
  Pattern Recognition}, pages 16489--16498, 2021.

\bibitem{zhou2021c3}
Yanning Zhou, Hang Xu, Wei Zhang, Bin Gao, and Pheng-Ann Heng.
\newblock {C3-SemiSeg}: Contrastive semi-supervised segmentation via cross-set
  learning and dynamic class-balancing.
\newblock In {\em Proceedings of the IEEE/CVF International Conference on
  Computer Vision}, pages 7036--7045, 2021.

\bibitem{zhu2020grasping}
Haiyue Zhu, Yiting Li, Fengjun Bai, Wenjie Chen, Xiaocong Li, Jun Ma, Chek~Sing
  Teo, Pey~Yuen Tao, and Wei Lin.
\newblock Grasping detection network with uncertainty estimation for
  confidence-driven semi-supervised domain adaptation.
\newblock In {\em IEEE/RSJ International Conference on Intelligent Robots and
  Systems}, pages 9608--9613. IEEE, 2020.

\bibitem{zou2020pseudoseg}
Yuliang Zou, Zizhao Zhang, Han Zhang, Chun-Liang Li, Xiao Bian, Jia-Bin Huang,
  and Tomas Pfister.
\newblock {PseudoSeg}: Designing pseudo labels for semantic segmentation.
\newblock {\em International Conference on Learning Representations}, 2021.

\end{thebibliography}
